\pdfoutput=1
\documentclass[11pt]{article}
\usepackage[preprint]{acl}
\usepackage{times}
\usepackage{latexsym}
\usepackage[T1]{fontenc}
\usepackage[utf8]{inputenc}
\usepackage{microtype}
\usepackage{graphicx}
\usepackage{booktabs}
\usepackage{amsmath}
\usepackage{amssymb}
\usepackage{amsthm}
\usepackage{xurl}

\newcommand{\NewBatchFlips}{11\nobreakdash--22}
\newcommand{\ExcessRange}{6.5\nobreakdash--13.3}
\newcommand{\MembershipRange}{0.22\nobreakdash--0.44}
\newcommand{\StarsExcessRange}{1.4\nobreakdash--2.0}
\newcommand{\StarsNewFlips}{2.9\nobreakdash--3.9}
\newcommand{\OtherVerdicts}{7\nobreakdash--23}
\newcommand{\OtherTexts}{9\nobreakdash--26}

\newcommand{\OwnReproduced}{97.8\nobreakdash--99.5}
\newcommand{\LevelChangeRange}{16\nobreakdash--39}
\newcommand{\StarsLevelChange}{16\nobreakdash--21}

\newcommand{\NumItems}{2{,}000}
\newcommand{\BatchedBaselineRange}{4.2\nobreakdash--8.4}
\newcommand{\NumericalRange}{0.5\nobreakdash--2.1}
\newcommand{\DownRange}{8\nobreakdash--54}
\newcommand{\AloneVerdictRange}{2\nobreakdash--12}
\newcommand{\FillerRange}{12\nobreakdash--41}
\newcommand{\FirstRange}{6\nobreakdash--22}
\newcommand{\ShuffledRange}{5\nobreakdash--19}
\newcommand{\BothRange}{9\nobreakdash--22}
\newcommand{\UnsureFlips}{14\nobreakdash--38}
\newcommand{\SureFlips}{6\nobreakdash--21}
\newcommand{\StarsUnsureShare}{1.4\nobreakdash--4.7}
\newcommand{\StarsSureFlips}{2.2\nobreakdash--3.5}
\newcommand{\KFiveRange}{9\nobreakdash--16}
\newcommand{\KFiftyRange}{12\nobreakdash--23}
\newcommand{\MaxSpearmanDrop}{0.10}
\newcommand{\AurocRange}{0.72\nobreakdash--0.83}
\newcommand{\StarsAurocRange}{0.96}
\newcommand{\PopBatchedRange}{13\nobreakdash--21}
\newcommand{\PopRenamedRange}{0.9\nobreakdash--3.2}

\newcommand{\GemmaRaceAlone}{58}
\newcommand{\GemmaRaceBatched}{35}
\newcommand{\UnsureShareRange}{2.4\nobreakdash--47.8}

\newcommand{\StarsExcessLower}{0.6\nobreakdash--1.3}
\newcommand{\NewBatchMean}{14.8}
\newcommand{\AurocDeltaRange}{$-0.033$ to $+0.010$}
\newcommand{\AurocDeltaSig}{six}

\newcommand{\KeptMembershipRange}{0.16\nobreakdash--0.48}
\newcommand{\StarsDRange}{0.005\nobreakdash--0.008}
\newcommand{\HGExcessLower}{5.3\nobreakdash--11.1}
\newcommand{\OwnMisses}{50}
\newcommand{\OwnClearMisses}{10}
\newcommand{\OwnClearMax}{4}
\newcommand{\OwnClearModels}{\mbox{Qwen3-4B} and \mbox{Qwen3-8B}}
\newcommand{\StarsNewFlipsMean}{3\nobreakdash--4}
\newcommand{\BothBelowEither}{seven}
\newcommand{\GOneGrading}{five}
\newcommand{\GOneStars}{none}
\newcommand{\PopGroupFewer}{652 on race for Granite}
\newcommand{\MissingMax}{13}
\newcommand{\ValidMin}{1{,}971}
\newcommand{\LeftOutMax}{5}
\newcommand{\LeftOutShift}{0.25}
\newcommand{\DNewRange}{0.06\nobreakdash--0.24}
\newcommand{\OrderShareRange}{0.42\nobreakdash--0.69}
\newcommand{\SlotMainRange}{0.09\nobreakdash--0.32}
\newcommand{\SlotMainSig}{nine}
\newcommand{\MembMainRange}{0.01\nobreakdash--0.25}
\newcommand{\MembMainMean}{0.15}
\newcommand{\MembMainSig}{six}
\newcommand{\SlotMembRange}{$-$0.07 to 0.18}
\newcommand{\SlotMembMean}{0.07}
\newcommand{\SlotMembSig}{none}
\newcommand{\SigBorder}{; for Granite on grading, the lower ends for the slot and for membership lie within 0.01 of zero}
\newcommand{\ComponentsNegative}{two}

\newcommand{\NegativeSmallest}{$-$0.07}

\newcommand{\KeepFirst}{4.0}
\newcommand{\KeepOthers}{9.9\nobreakdash--11.8}
\newcommand{\ChanceFlips}{11.0}
\newcommand{\ChanceSE}{0.7}

\newcommand{\ChanceN}{2{,}470}
\newcommand{\ChancePlacements}{494}
\newcommand{\ChanceDistinct}{467}
\newcommand{\KeepWeightedFlips}{10.8}
\newcommand{\KeepRawFlips}{9.6}

\newcommand{\MembFixedRange}{0.10\nobreakdash--0.30}
\newcommand{\MembFixedMean}{0.22}
\newcommand{\MembFixedSig}{four}
\newcommand{\MembFixedModels}{one}
\newcommand{\GTwoWeighted}{not be met}

\newcommand{\DecisionOrderRange}{0.62\nobreakdash--0.80}
\newcommand{\DecisionMembRange}{0.00\nobreakdash--0.14}
\newcommand{\MembMainDecMean}{0.08}
\newcommand{\SlotMembDecMean}{0.07}
\newcommand{\SlotMembGeMain}{five}
\newcommand{\CheckSplitObs}{24.5\nobreakdash--49.9}
\newcommand{\CheckSplitErr}{0.4}

\newcommand{\CheckMajErr}{0.7}

\newcommand{\CheckRepErr}{0.3}

\newcommand{\CheckRepObs}{2.5\nobreakdash--5.2}
\newcommand{\CheckRepLow}{ten}
\newcommand{\CheckRepAltMax}{0.42}
\newcommand{\CheckRepAltLow}{ten}

\newcommand{\VoteFiveNull}{54\nobreakdash--58}
\newcommand{\VoteNineNull}{42\nobreakdash--45}
\newcommand{\VoteFiveAvg}{46\nobreakdash--49}
\newcommand{\VoteNineAvg}{34\nobreakdash--37}

\newcommand{\BiasLossRange}{11\nobreakdash--20}

\newcommand{\BiasModelRange}{8\nobreakdash--19}

\newcommand{\BiasAllThreeRange}{3.5\nobreakdash--10.2}
\newcommand{\BiasBoundRange}{0.35\nobreakdash--7.06}
\newcommand{\BiasBoundSig}{eight}

\newcommand{\BiasStarsModel}{3.8\nobreakdash--5.2}
\newcommand{\BiasStarsBound}{0.98\nobreakdash--2.04}
\newcommand{\BiasStarsBoundSig}{five}
\newcommand{\ErrorRange}{20\nobreakdash--38}
\newcommand{\ErrorBoundRange}{41\nobreakdash--74}
\newcommand{\DisjointUpper}{1.22}

\newcommand{\ChanRatioRange}{0.7\nobreakdash--2.5}
\newcommand{\ChanCorrRange}{0.26\nobreakdash--0.63}
\newcommand{\ChanNearTwo}{nine}
\newcommand{\ChanCrossMargin}{0.2}
\newcommand{\ChanCrossedRange}{0.0\nobreakdash--13.8}
\newcommand{\ChanCrossedAll}{8.7}
\newcommand{\ChanCrossedTargets}{349}
\newcommand{\ChanTargets}{3{,}991}
\newcommand{\ChanCrossedJ}{40}
\newcommand{\ChanCrossedUndecided}{95}

\newcommand{\ChanDecCrossed}{1.8}

\newcommand{\ChanSpurious}{53}
\newcommand{\ChanSecondLost}{7}
\newcommand{\ChanFirstCrossed}{348}
\newcommand{\ChanNoiseTotal}{60}
\newcommand{\ChanNoiseTotalUpper}{115}
\newcommand{\ChanNoiseHugeCount}{nine}
\newcommand{\ChanNoiseHugeUndecided}{1.2}
\newcommand{\ChanNoiseHugeDecided}{1}
\newcommand{\ChanNoiseDecidedN}{3{,}330}
\newcommand{\ChanSecondCrossed}{342}
\newcommand{\ThreshHarmRatio}{1.4\nobreakdash--2.8}
\newcommand{\ThreshGradingRatio}{1.0\nobreakdash--2.2}
\newcommand{\ThreshGradingFlat}{four}
\newcommand{\ThreshStarsRatio}{3.2\nobreakdash--5.2}
\newcommand{\ThreshStarsPrimaryMin}{five}
\newcommand{\StarsDMax}{0.043}
\newcommand{\OtherDMin}{0.057}
\newcommand{\ThreshHarmGrows}{four}
\newcommand{\LawOne}{4.5}
\newcommand{\LawTen}{14.4}
\newcommand{\CheckSplitUnder}{nine}
\newcommand{\CheckSplitUnderMean}{0.3}

\newcommand{\CheckMajOver}{seven}
\newcommand{\CheckMajOverMean}{0.6}
\newcommand{\CheckMajSig}{three}

\newcommand{\CheckSplitSE}{0.3\nobreakdash--0.4}
\newcommand{\CheckMajSE}{0.6\nobreakdash--0.8}
\newcommand{\CheckRepSE}{0.3\nobreakdash--0.4}
\newcommand{\HolmMembMain}{five}
\newcommand{\HolmSlotMain}{nine}
\newcommand{\HolmBias}{six}
\newcommand{\HolmMembNarrow}{Gemma-4 on harm}
\newcommand{\MembMainPooledRange}{0.04\nobreakdash--0.22}
\newcommand{\MembMainPooledMean}{0.14}

\newcommand{\BiasParaShift}{4.6}
\newcommand{\MembPoolRange}{0.12\nobreakdash--0.35}
\newcommand{\MembPoolSig}{seven}
\newcommand{\MembPoolModels}{three}

\newcommand{\MembPoolOneMax}{two}

\newcommand{\MembMainSigCl}{five}

\newcommand{\ClusterRatioMembMain}{0.90\nobreakdash--1.19}
\newcommand{\MembMainClNote}{(no longer for Granite on grading, and for Gemma-4 on harm only by less than 0.01)}

\newcommand{\BiasStarsAbove}{four}
\newcommand{\BiasStarsFlips}{3.1\nobreakdash--4.0}
\newcommand{\BiasStarsExceptions}{Qwen3-8B (3.8\% against 4.0\%)}
\newcommand{\SynRuns}{12}
\newcommand{\SynTrueD}{0.30}

\newcommand{\SynTrueOrder}{0.11}
\newcommand{\SynOrder}{0.13 (0.02)}
\newcommand{\SynCoverOrder}{12}
\newcommand{\SynTrueSlot}{0.43}
\newcommand{\SynSlot}{0.41 (0.03)}
\newcommand{\SynCoverSlot}{11}
\newcommand{\SynTrueMembership}{0.24}
\newcommand{\SynMembership}{0.24 (0.03)}
\newcommand{\SynCoverMembership}{11}
\newcommand{\SynTrueSlotMembership}{0.22}
\newcommand{\SynSlotMembership}{0.22 (0.05)}
\newcommand{\SynCoverSlotMembership}{12}
\newcommand{\SynVoteThreeSim}{20.2}
\newcommand{\SynVoteThreePred}{19.6}
\newcommand{\SynVoteThreeExact}{19.8}
\newcommand{\SynVoteThreeModel}{20.0}
\newcommand{\SynVoteFiveSim}{18.4}
\newcommand{\SynVoteFivePred}{17.7}
\newcommand{\SynVoteFiveExact}{17.9}
\newcommand{\SynVoteFiveModel}{18.1}
\newcommand{\SynVoteNineSim}{16.8}
\newcommand{\SynVoteNinePred}{16.0}
\newcommand{\SynVoteNineExact}{16.2}
\newcommand{\SynVoteNineModel}{16.4}
\newcommand{\SynVoteGap}{0.16\nobreakdash--0.17}
\newcommand{\SynVoteSE}{0.29}
\newcommand{\SynModelSE}{0.01}
\newcommand{\SynVoteResidZ}{1.9}
\newcommand{\SynModelItems}{16{,}000{,}000}
\newcommand{\SynBiasTrue}{11.5}
\newcommand{\SynBiasModel}{11.7}
\newcommand{\SynBiasBound}{$-$1.2}
\newcommand{\SynBiasNine}{14.5}

\newcommand{\FigOneModel}{Qwen3-4B}
\newcommand{\PopRenamedQwenSmall}{24.2}
\newcommand{\PopRenamedQwenMid}{3.2}
\newcommand{\PopRenamedQwenLarge}{1.1}
\newcommand{\PopBatchedQwenSmall}{24.0}
\newcommand{\PopBatchedQwenMid}{12.9}
\newcommand{\PopBatchedQwenLarge}{12.8}
\newcommand{\PopTinyParsed}{674}
\newcommand{\PopTinyBatched}{6{,}000}
\newcommand{\PopOverlap}{258}
\newcommand{\StarsAdjacentRange}{76\nobreakdash--83}
\newcommand{\LevelRatioRange}{1.6\nobreakdash--2.9}

\newcommand{\SmokePlacements}{24}

\newcommand{\SmokeTieMaxDiff}{0.10}
\newcommand{\SmokeClearCount}{one}
\newcommand{\SmokeClearModel}{Qwen3-4B}
\newcommand{\SmokeClearDiff}{0.47}

\newcommand{\ScaleNewRange}{5.9\nobreakdash--14.6}
\newcommand{\ScaleNewRangeInt}{6\nobreakdash--15}
\newcommand{\ScaleHarmNewRange}{8.4\nobreakdash--14.6}
\newcommand{\ScaleNewMean}{10.1}

\newcommand{\ScaleExcessRange}{4.2\nobreakdash--9.1}

\newcommand{\ScaleAloneRange}{5.8\nobreakdash--17.9}
\newcommand{\ScaleAurocHarmRange}{0.011\nobreakdash--0.057}
\newcommand{\ScaleAurocHarmSig}{three}
\newcommand{\ScaleAurocGradingRange}{$-$0.021 to +0.035}
\newcommand{\ScaleExpAurocRange}{$-$0.023 to +0.012}
\newcommand{\ScaleLevelThreeAlone}{36\nobreakdash--52}
\newcommand{\ScaleLevelThreeBatched}{21\nobreakdash--39}
\newcommand{\ScaleAurocHarmBorder}{\mbox{Qwen3-14B}}
\newcommand{\ScaleFlagDropCount}{three}
\newcommand{\ScaleFlagDropRange}{9\nobreakdash--17}
\newcommand{\ScaleWordingRange}{$-$1.6 to +0.2}
\newcommand{\ScaleRewordedRange}{6.1\nobreakdash--13.0}
\newcommand{\ScaleBoundSig}{six}
\newcommand{\ScaleBiasModelRange}{4.1\nobreakdash--17.1}

\newcommand{\ScaleBoundCalPairs}{\mbox{Qwen3-14B} and \mbox{Gemma-4-31B} on harm}
\newcommand{\ScaleBoundCalBorder}{\mbox{Qwen3-14B} on grading}
\newcommand{\ScaleBoundCalHalf}{five}
\newcommand{\MainBoundCalHalf}{four}
\newcommand{\ScaleGTwoCount}{five}
\newcommand{\ScaleGTwoBoth}{\mbox{Gemma-4-31B} and \mbox{GLM-4-32B}}
\newcommand{\ScaleGTwoWeighted}{two}
\newcommand{\ScaleGTwoWeightedPairs}{\mbox{Gemma-4-31B} and \mbox{GLM-4-32B} on harm}
\newcommand{\ScaleVsNNeg}{four}
\newcommand{\ScaleVsNRange}{$-$0.16 to $-$0.05}
\newcommand{\ScaleQwenRenamed}{2.1}
\newcommand{\ScaleQwenFourteenAlone}{12.0}
\newcommand{\ScaleQwenThirtyTwoAlone}{17.0}
\newcommand{\GemmaLargeDNew}{0.018\nobreakdash--0.030}
\newcommand{\GemmaLargeVsN}{$-$0.37 [$-$0.77, $-$0.06]}
\newcommand{\VoteMeasThreeRange}{71\nobreakdash--76}
\newcommand{\VoteMeasFiveRange}{60\nobreakdash--69}
\newcommand{\VoteMeasNineRange}{52\nobreakdash--60}
\newcommand{\VoteMeasNinePooled}{56\% [53, 58]}

\newcommand{\VoteInSampleGap}{0.2\nobreakdash--0.3}

\newcommand{\VoteInSampleRatioMax}{6.1}
\newcommand{\VoteHolm}{two, two, one and none}
\newcommand{\VoteLimitRange}{4.5\nobreakdash--8.9}
\newcommand{\VoteNMin}{1{,}977}
\newcommand{\VoteTwelveMean}{0.2}
\newcommand{\VoteTwelveMiss}{$-$2.0 to +2.2}
\newcommand{\VoteTwelveRatioMiss}{$-$10 to +11}
\newcommand{\VoteTwelveInside}{eight}
\newcommand{\VoteVNDecRange}{0.38\nobreakdash--0.81}
\newcommand{\VoteVNLatentRange}{0.11\nobreakdash--0.22}
\newcommand{\VoteVNLatentPooled}{0.17}
\newcommand{\VoteSoftNineRange}{49\nobreakdash--60}
\newcommand{\VoteAloneOneRange}{11.2\nobreakdash--19.7}
\newcommand{\VoteAloneNineRange}{9.1\nobreakdash--18.6}
\newcommand{\VoteWordingRange}{$-$1.9 to +0.1}
\newcommand{\VoteRewordedRange}{9.9\nobreakdash--21.0}
\newcommand{\HardwareDifferRange}{0.4\nobreakdash--2.8}
\newcommand{\HardwareVsNumericRange}{$-$0.3 to +0.4}
\newcommand{\HardwarePartOneMax}{2.8}
\newcommand{\HardwareFlipRange}{$-$0.1 to +1.1}
\newcommand{\HardwareFlipSig}{one}
\newcommand{\WordingNegative}{16}
\newcommand{\WordingPairs}{18}
\newcommand{\RewordedAllRange}{6.1\nobreakdash--21.0}
\newcommand{\BvoteNewOnlyMax}{1.6}

\newcommand{\BvoteNineMAE}{0.5}

\newcommand{\CalSingleThirteen}{10.4}
\newcommand{\CalHeldThirteen}{11.0}
\newcommand{\CalLessMax}{3.4}
\newcommand{\CalLessPair}{\mbox{Granite} on grading}
\newcommand{\CalMoreMax}{4.3}
\newcommand{\CalMorePair}{\mbox{Qwen3-8B} on grading}
\newcommand{\BvoteParaShiftRange}{$-$4.5 to +1.7}
\newcommand{\BvoteParaPooled}{12.5}
\newcommand{\BvoteAlonePooledBoth}{12.6}
\newcommand{\BvoteOneRange}{10.9\nobreakdash--19.8}
\newcommand{\BvoteFifteenRange}{8.4\nobreakdash--17.8}
\newcommand{\BvoteOnePooled}{16.0}
\newcommand{\BvoteFifteenPooled}{12.6\% [12.1, 13.2]}

\newcommand{\BvoteNineToFifteen}{0.7}
\newcommand{\BvoteModelMax}{1.5}
\newcommand{\BvoteModelMAE}{0.6}
\newcommand{\BvoteInSampleMax}{0.9}
\newcommand{\BvoteDFivePooled}{$-$0.2 [$-$0.3, $-$0.1]}
\newcommand{\BvoteCalRange}{7.9\nobreakdash--13.5}
\newcommand{\BvoteCalPooled}{10.4\% [9.9, 10.9]}
\newcommand{\BvoteCalShare}{82}
\newcommand{\BvoteGThree}{$-$0.4 [$-$0.5, $-$0.3]}
\newcommand{\BvoteUFifteen}{+1.8 [+1.4, +2.2]}
\newcommand{\BvoteUFive}{+0.2}
\newcommand{\BvoteUNine}{+0.9}
\newcommand{\BvoteHolmD}{two}
\newcommand{\BvoteHolmG}{five}
\newcommand{\BvoteHolmU}{four}
\newcommand{\BvoteHardwareRange}{0.5\nobreakdash--1.8}
\newcommand{\BvoteNRange}{1{,}975\nobreakdash--2{,}000}
\newcommand{\BvoteLabelCount}{four}
\newcommand{\BvoteLabelMax}{83}

\newcommand{\level}[1]{\mbox{\textsc{#1}}}
\hypersetup{pdftitle={Judged in Company: In-Context Batching Makes LLM Verdicts Depend on Other Items},
  pdfauthor={Peng Xie, Amr Alanwar}}
\newcommand{\E}{\mathbb{E}}
\newcommand{\Var}{\operatorname{Var}}
\newcommand{\Cov}{\operatorname{Cov}}
\newcommand{\Prob}{\mathbb{P}}
\newtheorem{proposition}{Proposition}
\newtheorem{lemma}{Lemma}
\theoremstyle{definition}
\newtheorem{definition}{Definition}
\newtheorem{assumption}{Assumption}

\title{Judged in Company: In-Context Batching Makes LLM Verdicts\\ Depend on Other Items}

\author{Peng Xie \quad Amr Alanwar \\
  TUM School of Computation, Information and Technology, Department of Computer Engineering \\
  Technical University of Munich, 74076 Heilbronn, Germany \\
  \texttt{p.xie@tum.de} \quad \texttt{alanwar@tum.de}}

\begin{document}
\maketitle

\begin{abstract}
To save cost, practitioners often ask a language model to grade many independent items in one prompt. This assumes that verdicts are \emph{separable}: that the verdict on an item does not depend on the other items in the prompt or on its position among them. Across three judgement tasks (harm in social-media comments, short-answer grading and review star ratings) and five open-weight models with 4\nobreakdash--9B parameters, assigning the same \NumItems{} items to different random batches of 25 changes \NewBatchFlips\% of decisions on harm and grading, \ExcessRange{} points more than renaming the item keys or rewording the instruction, while AUROC against human labels differs from single-item grading only by \AurocDeltaRange; for four models with 14\nobreakdash--32B parameters, \ScaleNewRangeInt\% of decisions change, \ScaleExcessRange{} points more. We show that flip rates decompose exactly into components of order, slot and membership, and that under a threshold model they grow with the square root of the variance that the context contributes, so that effects that are small for scores are large for decisions. Most of the latent dissimilarity between partitions comes from the order of the neighbours and the slot. The main effect of which items share the prompt, the part that voting over permutations of a batch cannot remove, accounts for \VoteVNLatentRange{} of it, and majority votes over nine permutations still leave \VoteMeasNineRange\% of the flips. Batched verdicts also differ systematically from single-item verdicts: even the majority of fifteen random batches disagrees with them on \BvoteFifteenRange\% of decisions. Editing a single generation shows that both the model's own earlier verdicts and the neighbours' texts carry the effect. On the easy task, star ratings, only \StarsNewFlips\% of decisions change. We recommend grading ambiguous items one per prompt, recovering most of the cost with a cached prefix, and reporting agreement with single-item grading whenever items are batched.
\end{abstract}

\section{Introduction}

Language models are now routinely used to label, grade and moderate text at scale. Because every request repeats the same instructions, a common way to cut cost is \emph{in-context batching}: many items are placed in one prompt and the model returns one verdict per item, for example as a JSON object \citep{cheng2023batch}. Vendors recommend it for moderation that need not happen in real time \citep{anthropic2026moderation}, and a recent study of social-science annotation finds batches of 25 to 100 items, with up to ten coded variables per prompt, safe for most current models: they save more than 80\% of token costs and lose less than 2 points of accuracy \citep{pipal2026waste}.

Batching rests on an assumption that is rarely stated: that the verdict for an item depends only on that item and not on its \emph{company} (the other items in the prompt) or on its position among them. Nothing in a decoder-only model enforces this. When the model writes the verdict for the twelfth item, it has read the other 24 items and the eleven verdicts it has already written. Accuracy is also a poor test of the assumption, because shifts that push some items up and others down can leave aggregate scores almost unchanged.

\begin{figure}[t]
\centering
\small
\fbox{\begin{minipage}{0.94\columnwidth}
\textbf{Question:} What is the relationship between an organism's range of tolerance and its optimum condition?\\
\textbf{Reference answer:} The optimum condition is within the range of tolerance.\\
\textbf{Student's answer:} \emph{His optimum condition is in his range of tolerance.} \hfill (human grade: \level{correct})\\[4pt]
\begin{tabular}{@{}l@{\hspace{6pt}}l@{}}
\toprule
\FigOneModel & verdict (probability) \\
\midrule
graded alone & \level{correct} (0.97) \\
\quad alone, keys renamed & \level{correct} (1.00) \\
in random batch A, item 14 of 25 & \level{partial} (1.00) \\
\quad batch A, keys renamed & \level{partial} (1.00) \\
\quad batch A, instruction reworded & \level{partial} (1.00) \\
in random batch B, item 15 of 25 & \level{partial} (0.85) \\
in random batch C, item 8 of 25 & \level{correct} (1.00) \\
\bottomrule
\end{tabular}
\end{minipage}}
\caption{One short answer from SciEntsBank graded by the same model. The verdict is stable under trivial changes of the prompt (renamed keys, a reworded instruction) but changes with the batch; batches A, B and C differ in which 24 other answers share the prompt, in their order and in the slot of this answer. Each verdict is given with high confidence.}
\label{fig:example}
\end{figure}

Figure~\ref{fig:example} shows what can happen. A correct student answer is graded \level{correct} when it is alone, \level{partial} in two random batches and \level{correct} in a third, each time with high confidence. Renaming the item keys or rewording the instruction does not change the verdict within a batch; changing the batch does.

We measure how often this happens, where it comes from and when it matters. We state the assumption formally as \emph{separability}, show that the flip rate between random partitions decomposes exactly into components of order, slot and membership, and link flips to the variance of latent scores (\S\ref{sec:setup}). We grade the same 2{,}000 items per task alone, in three independent random partitions, and in conditions that each change a known subset of three aspects of the batch: the order of an item's neighbours, the item's slot, and which items are its neighbours. We compare every change against baselines measured in the same regime: renamed keys, a reworded instruction, and the same prompts generated one at a time. We then intervene inside single generations, editing either the model's earlier verdicts or the neighbours' texts, to see through which channel the batch acts. Three pass/fail criteria (\emph{gates}), on generality, membership and mechanism, were fixed before the runs. Follow-up runs, not covered by the gates, repeat the core design on harm and grading with four 14\nobreakdash--32B models and measure, for the five main models, what voting over permutations of a batch and over independent batches leaves (Appendix~\ref{app:followup}). Our findings:

\begin{itemize}\itemsep 1pt
\item On harm and short-answer grading, a new random partition changes \NewBatchFlips\% of decisions for all five models, \ExcessRange{} points above the largest baseline, while AUROC differs from single-item grading only by \AurocDeltaRange{} (\S\ref{sec:exists}, \S\ref{sec:accuracy}). For four models with 14\nobreakdash--32B parameters, a new partition changes \ScaleNewRangeInt\% of decisions, \ScaleExcessRange{} points above the largest baseline (Appendix~\ref{app:followup}).
\item The order of the neighbours and the slot account for most of the latent dissimilarity between partitions. Which items share the batch accounts for \MembFixedRange{} of it when the item keeps its slot (\MembershipRange{} by the measure we fixed before the runs, which overweights the first slot). The main effect of membership, which voting over permutations \citep{lin2024batchprompt} cannot remove, accounts for \VoteVNLatentRange{} of the dissimilarity in runs that grade every batch in nine orders; there, majority votes over five and nine permutations leave \VoteMeasFiveRange\% and \VoteMeasNineRange\% of the flips, and a threshold model fitted to these runs predicts the flip rates within one point (\S\ref{sec:where}, \S\ref{sec:voting}).
\item Batching shifts verdicts and does not only scatter them: on harm and grading, a random batch disagrees with single-item grading on \BvoteOneRange\% of decisions, and the majority of fifteen independent random batches still disagrees on \BvoteFifteenRange\% (both on H200 NVL GPUs), close to what the threshold model fitted to the main experiment predicted, in a check that is only partly out of sample (\S\ref{sec:voting}).
\item Both channels act: swapping in another batch's verdicts changes \OtherVerdicts\% of decisions, swapping in another batch's texts \OtherTexts\% (\S\ref{sec:how}).
\item The effect is large where the task is ambiguous. On star ratings, where models agree with the true rating at AUROC \StarsAurocRange, the written rating changes for \StarsLevelChange\% of reviews but only \StarsNewFlips\% of decisions (\S\ref{sec:when}).
\end{itemize}

\section{Related Work}

\paragraph{Batch prompting.} Placing several inputs in one prompt was proposed to save tokens and calls \citep{cheng2023batch}. Follow-up work reports aggregate accuracy as the batch grows \citep{larionov2025batchgemba,pipal2026waste,fagerberg2026batch}, uses in-batch comparison for evaluation \citep{yuan2024batcheval} or relevance \citep{korikov2025batched}, and finds that accuracy depends on an item's position; BatchPrompt votes over permutations to reduce this \citep{lin2024batchprompt}. These studies measure accuracy, not whether individual decisions change.

\paragraph{Item-level effects of company.} Closest to our work, \citet{hida2026class} report that 1--27\% of screening decisions change between single-item processing and batches of 20, and that aggregate metrics do not always reflect these changes. \citet{chen2024threshold} place relevant or irrelevant documents before fixed targets and find contrast effects in relevance judgements. \citet{huang2026contextual} show that both the order and the composition of a reranking context change relevance labels, and \citet{frohmann2026equal} show that batched scorers make different decisions under permutation at equal ranking quality and propose fine-tuning for order consistency. Related effects appear for harmful requests inside benign batches \citep{kim2026safety}, for injected instructions that corrupt answers to other items \citep{yue2025batch}, and for violent speech, which GPT-4 labels less often in larger batches \citep{matter2024incel}. We add a decomposition that separates membership from order and slot, interventions that separate two channels within one generation, measurements of what voting over permutations (as in BatchPrompt) and over independent batches leaves, and a common decision-level comparison across tasks.

\paragraph{Context effects on judgements.} LLM judges favour positions \citep{zheng2023judging,wang2024fair,shi2025judging}, anchor on scores shown to them \citep{kapetanovic2026anchoring}, drift with the polarity of earlier turns \citep{temkit2026amel} and carry earlier decisions forward in sequential tasks \citep{echterhoff2024cognitive}. Demonstrations bias predictions towards majority and recent labels \citep{zhao2021calibrate}. People rate items relative to the range of items they see \citep{parducci1965category} and avoid streaks in sequences of decisions \citep{chen2016gambler}.

\paragraph{Arbitrariness and measurement.} Decisions that change with arbitrary choices raise fairness concerns \citep{marx2020predictive,black2022model,cooper2024arbitrariness}; in content moderation, \citet{gomez2024arbitrariness} trace such arbitrariness to training randomness. Measurement frameworks for LLM judges split the uncertainty of their outputs into sources such as the judge model, the prompt wording and the sampling temperature \citep{messing2026hidden}, or fit many-facet Rasch models with rater and item facets \citep{sachdeva2026rasch}; batch composition is not among these sources. We treat the context of an item as the input of a variance decomposition \citep{sobol2001global,saltelli2010variance}.

\paragraph{Independence by construction.} Masks that let each item attend only to a shared prefix and to its own tokens were introduced for long contexts \citep{ratner2023pcw}, cached prompt modules \citep{gim2024promptcache}, order-independent inputs \citep{mcilroyyoung2024order}, position-invariant reranking \citep{bito2026onepass} and parallel decoding \citep{glavas2026ippd}; in the first three, the text generated after the items attends to all of them. Serving systems reuse a cached prefix across requests \citep{kwon2023vllm,zheng2024sglang}. We use them as the reference point: they make verdicts separable by construction if each verdict, too, is computed from the shared prefix and its own item alone, and if the key and the position ids of the item and its verdict do not depend on its slot.

\section{Formal Framework}
\label{sec:setup}

\paragraph{Batched grading.} A task gives items $1,\dots,n$, an ordinal scale and a decision threshold $\tau$. A \emph{partition} assigns the items to prompts of $k$ items. The \emph{context} of item $i$ is $c=(s,N,\omega)$: its slot $s$ in the prompt, where its text follows the key ``\texttt{c}$s$:''; its \emph{membership} $N$, the set of its $k-1$ neighbours; and the order $\omega$ in which the neighbours appear. The model writes one JSON object with a level for every item, left to right, by greedy decoding. The level $Y_i(c)$ of item $i$ and its decision $d_i(c)={\mathbf{1}[Y_i(c)\ge\tau]}$ are therefore functions of the context, up to the numerical noise of batched inference, which we treat as a further factor that every condition redraws (\S\ref{sec:design}, Appendix~\ref{app:proofs}).

\begin{definition}
The grader is \emph{separable} if $Y_i(c)=Y_i(c')$ for every item $i$ and all its contexts $c$ and $c'$. Its decisions are separable if the same holds for $d_i$.
\end{definition}

Separability contains the assumption of no interference between units in causal inference \citep{hudgens2008toward}, applied to the items of a prompt, and also excludes an effect of the item's own slot; a random partition randomises the context. A decoder-only model does not enforce it. It generates the verdicts as
\begin{equation}
p_\theta(y\mid x)=\textstyle\prod_{t=0}^{k-1}p_\theta(y_t\mid x,y_{<t}),
\label{eq:factor}
\end{equation}
where $\theta$ denotes the parameters of the model, $x_t$ is the text in slot $t$, $y_t$ its verdict, $x=(x_0,\dots,x_{k-1})$ and $y=(y_0,\dots,y_{k-1})$. The factor of slot $t$ can depend on the texts in the other slots, $x_{-t}$, the \emph{text channel}, and on the verdicts $y_{<t}$ written before it, the \emph{verdict channel}.

\paragraph{Flips.} For two conditions $A$ and $B$, the flip rate $F(A,B)$ is the share of items whose decisions differ. In a uniformly random partition, the slot, the membership and the order of an item are independent and uniform (Lemma~\ref{lem:design}; all proofs are in Appendix~\ref{app:proofs}). Let $p_i$ be the probability that $d_i=1$ in such a context.

\begin{proposition}
\label{prop:flip}
Two independent random partitions give item $i$ different decisions with probability $2p_i(1-p_i)$. Their expected flip rate is $F_\mathrm{new}=\frac{2}{n}\sum_i p_i(1-p_i)$, and $F_\mathrm{new}=0$ if and only if the decisions are separable.
\end{proposition}

The equivalence holds for a grader without numerical noise. The noise of batched inference can make $F_\mathrm{new}$ positive even for a grader that would be separable without it, which is why we compare $F_\mathrm{new}$ with baselines that measure this noise (\S\ref{sec:design}).

\paragraph{Decomposition.} Let $f$ be a function of the context of an item, for example $d_i$. For a set $R$ of factors, let $C^{(R)}$ be the random context $C$ with the factors in $R$ redrawn independently and the others kept, and let $G(R)=\frac12\,\E\,(f(C)-f(C^{(R)}))^2$. For a decision, $G(R)$ is half the probability of a flip.

\begin{proposition}
\label{prop:pf}
$G(R)$ is the expected variance of $f$ given the factors that are kept, and it does not decrease when $R$ grows. Let $V_s=\Var\E[f\mid s]$, $V_N=\Var\E[f\mid N]$, $V_{sN}=\Var\E[f\mid s,N]-V_s-V_N$ and $T_\omega=\E\Var(f\mid s,N)$. Then ${V_{sN}\ge0}$ and
\begin{align*}
G(\{\omega\})&=T_\omega,\\
G(\{s,\omega\})&=T_\omega+V_s+V_{sN},\\
G(\{N,\omega\})&=T_\omega+V_N+V_{sN},\\
G(\{s,N,\omega\})&=T_\omega+V_s+V_N+V_{sN}.
\end{align*}
\end{proposition}

\begin{table}[t]
\centering
\small
\begin{tabular}{@{}lccc@{}}
\toprule
Condition (vs.\ base partition) & $\omega$ & $s$ & $N$ \\
\midrule
Keep slot, reorder the others & \checkmark & & \\
Reorder the whole batch & \checkmark & \checkmark & \\
Keep slot, new neighbours & \checkmark & & \checkmark \\
New random partition & \checkmark & \checkmark & \checkmark \\
\bottomrule
\end{tabular}
\caption{The factors that each condition redraws for an item, relative to its base partition: the order $\omega$ of its neighbours, its slot $s$ and its membership $N$.}
\label{tab:conditions}
\end{table}

This is the variance decomposition of \citet{hoeffding1948class} and \citet{sobol2001global}, and redrawing a subset of the inputs is its usual estimator \citep{saltelli2010variance}. The four conditions of Table~\ref{tab:conditions} redraw the four sets of Proposition~\ref{prop:pf}, the two ``keep slot'' conditions up to the qualifications below. If a condition gives an item the context $C'$ where the base partition gives it $C$, its value of $G$ is $\frac12\E(f(C)-f(C'))^2$; for ``keep slot, reorder the others'' we take it only over the items in the designated slots, weighted as described below. We write $G_\mathrm{keep}$, $G_\mathrm{resh}$, $G_\mathrm{comp}$ and $G_\mathrm{new}$ for these values; $G_\mathrm{resh}=G(\{s,\omega\})$ and $G_\mathrm{new}=G(\{s,N,\omega\})$. Then the main effect of membership is $V_N=G_\mathrm{new}-G_\mathrm{resh}$, the main effect of the slot is $V_s=G_\mathrm{new}-G_\mathrm{comp}$, and $T_\omega+V_{sN}=G_\mathrm{resh}+G_\mathrm{comp}-G_\mathrm{new}$ is split into the order $T_\omega=G_\mathrm{keep}$ and the interaction $V_{sN}$ of slot and membership. No component can be negative, which the estimates must reproduce up to sampling error and the small biases described next and in Appendix~\ref{app:synthetic}.

Three facts qualify this (Proposition~\ref{prop:conditions}, Appendix~\ref{app:proofs}). First, new neighbours are drawn from outside the old batch, so that $G_\mathrm{comp}$ differs from $G(\{N,\omega\})$ by at most \DisjointUpper\% of $V_N+V_{sN}$. Second, ``keep slot, reorder the others'' keeps the slots of the items in the five designated slots 0, 6, 12, 18 and 24 (400 items per partition), and four of their 24 neighbours stay in place as well, so $G_\mathrm{keep}$ is a lower bound of the order component at these slots. Third, these slots are not representative. In the first slot, where the model writes the verdict before any other, \KeepFirst\% of decisions change under ``keep slot, reorder the others'', against \KeepOthers\% in the other designated slots (five models, harm and grading). We therefore weight the designated slots as in a random partition, $1/25$ each for the first and the last slot and $23/75$ each for the three others, which stand for all interior slots. Items that keep their slot by chance in the reordered batch, one in 25, avoid both problems: the expected number of their flips is exactly $2\overline{T}_\omega$ times their expected number, where $\overline{T}_\omega$ is the mean of $T_\omega$ over items for $f=d_i$ (Proposition~\ref{prop:conditions}). Pooled over the five models on harm and grading, \ChanceFlips\% of their \ChanceN{} decisions change (\ChancePlacements{} placements of \ChanceDistinct{} distinct items, each graded by all five models; standard error \ChanceSE, clustered by batch), against \KeepWeightedFlips\% in the weighted designated slots and \KeepRawFlips\% without weights. $V_N$ is affected by none of these facts and $V_s$ only by the first. Numerical noise of batched inference is redrawn in every condition and is therefore counted in $T_\omega$.

\paragraph{Latent scale.} Flip rates depend on how many items lie near the threshold. To compare conditions and tasks on a scale that does not, we use a threshold model.

\begin{assumption}
\label{ass:latent}
Decisions threshold latent scores, $d_i(c)=\mathbf{1}[z_i(c)>h]$, and over a population of items and contexts the scores of an item in the two contexts of each condition of Table~\ref{tab:conditions} are bivariate normal with standard margins.
\end{assumption}

This is a model: with finitely many items and contexts it can hold only approximately, and it forces the effects of the context to be normal given the item, with the same variance for every item (Appendix~\ref{app:proofs}). We test its predictions in \S\ref{sec:where} and \S\ref{sec:voting}. Let $\rho$ be the correlation of the two scores of a condition, which is the tetrachoric correlation of the two decisions \citep{olsson1979polychoric}; we estimate it by maximum likelihood with an exact bivariate normal integral and 0.5 added to every cell. The \emph{latent dissimilarity} $D=1-\rho$ is the mean over items of $G$ for $f=z_i$, in units of the variance of $z$ over items and contexts. What we said about $G_\mathrm{keep}$, $G_\mathrm{resh}$, $G_\mathrm{comp}$ and $G_\mathrm{new}$ therefore holds for $D_\mathrm{keep}$, $D_\mathrm{resh}$, $D_\mathrm{comp}$ and $D_\mathrm{new}$, with components that are shares of the latent variance. In small tables the estimator overstates $D$, more for the 400 designated items of a partition than for all 2{,}000 items; this raises the order share slightly (Appendix~\ref{app:synthetic}). Before the runs we had fixed $(D_\mathrm{comp}-D_\mathrm{keep})/D_\mathrm{new}$, with unweighted designated slots, as the \emph{membership share at a fixed slot}; because of the second and the third fact above it overstates $(V_N+V_{sN})/D_\mathrm{new}$.

\begin{proposition}
\label{prop:law}
Let $(Z,Z')$ be bivariate normal with standard margins and correlation $\rho\in(-1,1)$, and let $D=1-\rho$. The decisions $\mathbf{1}[Z>h]$ and $\mathbf{1}[Z'>h]$ differ with probability
\begin{equation}
\Psi(h,\rho)=\frac1\pi\int_\rho^1\frac{e^{-h^2/(1+r)}}{\sqrt{1-r^2}}\,dr ,
\label{eq:law}
\end{equation}
which satisfies
\begin{equation*}
\frac{\arccos\rho}{\pi}e^{-h^2/(1+\rho)}\le\Psi(h,\rho)\le\frac{\arccos\rho}{\pi}e^{-h^2/2}
\end{equation*}
and, for fixed $h$, equals ${\frac{\sqrt{2D}}{\pi}\,e^{-h^2/2}\,(1+O(D))}$ as $D\to0$.
\end{proposition}

Flips therefore grow with the square root of the variance that a change of context contributes. At the median, a share of 1\% of the latent variance flips \LawOne\% of the decisions and a share of 10\% flips \LawTen\%. A change that is small on the scale of scores is large on the scale of decisions: when $D$ is small, averaging the scores over $M$ independent contexts, which divides the variance of the context by $M$, divides the flips only by about $\sqrt{M}$, and a majority vote over the $M$ decisions divides them by less (\S\ref{sec:voting}).

\paragraph{Voting over permutations.} BatchPrompt grades a batch in several permutations and takes the majority \citep{lin2024batchprompt}. A permutation redraws the slot and the order of every item and keeps its membership. Majorities over several contexts need a joint law of the scores, which Assumption~\ref{ass:latent} implies if $D_\mathrm{resh}>0$ and $D_\mathrm{new}<1$ (Appendix~\ref{app:proofs}):

\begin{assumption}
\label{ass:components}
Decisions threshold a latent score, $d_i(c)=\mathbf{1}[z_i(c)>h]$, which in standard units is $z=\mu+a+e$: $\mu$ is the mean of the item over contexts, $a=\E[z\mid\text{item},N]-\mu$ the main effect of the membership and $e$ the rest. Let $0<D_\mathrm{resh}\le D_\mathrm{new}<1$ and $V_N=D_\mathrm{new}-D_\mathrm{resh}$. Over the population of items $\mu\sim\mathcal{N}(0,1-D_\mathrm{new})$. For every item, the effect $a$ of a random membership is $\mathcal{N}(0,V_N)$, and given the item and the membership the rest $e$ of a random arrangement $(s,\omega)$ is $\mathcal{N}(0,D_\mathrm{resh})$; both laws are the same for all items.
\end{assumption}

\begin{proposition}
\label{prop:vote}
(i) The mean of $f$ over $m$ independent permutations of a batch has variance $V_N+G(\{s,\omega\})/m$ over batches, and as $m$ grows it converges to $\E[f\mid N]$, whose variance is $V_N$. (ii) Under Assumption~\ref{ass:components}, let two batches with independent memberships, such as the batches of an item in two independent random partitions, each be graded in an odd number $m$ of independent permutations. Their majority votes give an item drawn from the population different decisions with a probability $F_m$ that equals $\Psi(h,1-D_\mathrm{new})$ for $m=1$, decreases strictly as $m$ grows, and converges to
\begin{equation}
F_\infty=\Psi\Big(\frac{h}{\sqrt{1-D_\mathrm{resh}}},\,\frac{1-D_\mathrm{new}}{1-D_\mathrm{resh}}\Big),
\label{eq:floor}
\end{equation}
which is positive if and only if $V_N>0$. Hence $F_m>F_\infty$ for every odd $m$.
\end{proposition}

Voting thus removes the interaction of slot and membership along with slot and order, but not the main effect of membership.

\paragraph{Single-item grading as the reference.} Let $r_i$ be the decision on item $i$ when it is graded alone, $d_i^\star=\mathbf{1}[p_i>\frac12]$ its \emph{main} batched decision, $B_i=\mathbf{1}[d_i^\star\ne r_i]$ its \emph{bias} and $v_i=\min(p_i,1-p_i)$ its \emph{variance}, as in the decomposition of zero-one loss by \citet{domingos2000unified}, with random contexts in the place of training sets and $r_i$ in the place of the target.

\begin{proposition}
\label{prop:bias}
(i) A random context gives item $i$ a decision that differs from $r_i$ with probability $B_i+(1-2B_i)v_i$. (ii) If $p_i\ne\frac12$, the majority over $M$ independent random contexts differs from $r_i$ with a probability that converges to $B_i$. (iii) Let $X_i$ count the contexts among three independent random ones whose decision differs from $r_i$, and $U_i$ those with decision 1. Then, with $\Prob$ over contexts and a uniformly random item,
\begin{equation}
\begin{split}
\frac1n\sum_i B_i\ \ge\ &\Prob(X=3)\\
&-\tfrac13\min\{\Prob(U=1),\Prob(U=2)\}.
\end{split}
\label{eq:bound}
\end{equation}
\end{proposition}

Voting over batches removes variance, not bias. The bound~\eqref{eq:bound} needs no model. Under the threshold model, extended so that the single-item decision thresholds a score that is jointly normal with the mean of the item over contexts (Assumption~\ref{ass:reference}, Appendix~\ref{app:proofs}), the mean bias has a closed form (Proposition~\ref{prop:biasgauss}).

\paragraph{Interventions.} When the model writes the digit for the target in slot $t$, its next-token distribution depends on the prompt and on the tokens it has generated, including its verdicts for slots $0,\dots,t-1$. We cut the generation just before the target's digit, edit tokens, and read the distribution again; all other tokens of the answer, including its layout, stay those of the generation that is edited (except for one edit that lets the target be answered first, which keeps only the opening of the answer and gives its first key the target's number). Editing the earlier verdicts while keeping the texts isolates the verdict channel of Equation~\eqref{eq:factor}; replacing the neighbours' texts while keeping the verdicts isolates the text channel, at the layout of this answer (the texts can also change the layout that the model chooses, which the edit keeps). Since edited contexts are not generated, a decision here is $\mathbf{1}[\zeta\ge0.5]$, where $\zeta=P(\text{level}\ge\tau)$ is read from the next-token distribution, renormalised over the levels, and every edit is compared with the unedited forward pass. Let $\Delta_T$, $\Delta_V$ and $\Delta_{TV}$ be the changes of $\zeta$ when the texts, the verdicts or both are taken from another batch, and $J=\Delta_{TV}-\Delta_T-\Delta_V$. If the channels add up, $J=0$; if the four contexts are unrelated, $\E J^2$ is twice $\E\Delta_T^2$; and if the channels add up on a latent scale that $\zeta$ increases with, a change of either channel cannot move $\zeta$ in opposite directions for the two values of the other channel (Proposition~\ref{prop:channels}, Appendix~\ref{app:proofs}). Unedited, the procedure should reproduce the generated verdict; \S\ref{sec:how} reports how closely it does.

\section{Experimental Design}
\label{sec:design}

\paragraph{Tasks.} \emph{Harm}: comments from the Measuring Hate Speech corpus \citep{kennedy2020constructing} with at least three annotators, graded on a five-level harm scale; the decision is level~2 (insults or harassment aimed at a person) or above, and the human label is the corpus' hate-speech score above 0.5. \emph{Grading}: student answers from SciEntsBank \citep{dzikovska2013semeval}, graded \level{incorrect}, \level{partial} or \level{correct} against a reference answer; the decision is \level{correct}. \emph{Stars}: reviews from Yelp Review Full \citep{zhang2015character}, rated one to five stars; the decision is four stars or more. For each task we sample 2{,}000 items (Appendix~\ref{app:data}).

\paragraph{Models and protocol.} We use Qwen3-4B-Instruct-2507, Qwen3-8B with thinking disabled \citep{yang2025qwen3}, Gemma-4-E4B-it \citep{gemmateam2026gemma4}, Granite-3.3-8B-Instruct \citep{granite33modelcard} and GLM-4-9B-0414 \citep{glm2024chatglm}, called Qwen3-4B, Qwen3-8B, Gemma-4, Granite and GLM-4 in tables and text, in bf16 with greedy decoding, one RTX 5090 per model. The follow-up experiments of Appendix~\ref{app:followup} add \mbox{Qwen3-14B} and \mbox{Qwen3-32B} with thinking disabled, \mbox{Gemma-4-31B-it} and \mbox{GLM-4-32B-0414}, and run on one H200 NVL per model.\footnote{\raggedright Hugging Face revisions: Qwen3-4B-Instruct-2507 \texttt{cdbee75f17c0}, Qwen3-8B \texttt{b968826d9c46}, Gemma-4-E4B-it \texttt{ee0ef6023621}, Granite-3.3-8B-Instruct \texttt{51dd4bc2ade4}, GLM-4-9B-0414 \texttt{645b8482494e}; for Appendix~\ref{app:population} also Qwen3-0.6B \texttt{c1899de289a0} and Qwen3-1.7B \texttt{70d244cc86cc}; for Appendix~\ref{app:followup} Qwen3-14B \texttt{40c069824f42}, Qwen3-32B \texttt{9216db5781bf}, Gemma-4-31B-it \texttt{842da3794eaa} and GLM-4-32B-0414 \texttt{077b5c2f5c43}.} The prompt gives the scale, one line of context, the items and a request for a JSON object mapping every item id to a digit (Appendix~\ref{app:prompts}). Decisions come from the digits the model writes; for each digit we also record its level distribution.

\paragraph{Conditions.} Every item is graded alone and in three independent random partitions with $k=25$. For each partition we build the three derived conditions of Table~\ref{tab:conditions}; the fourth, a new random partition, is another of the three partitions. ``Keep slot, reorder the others'' holds five designated slots per batch. Baselines on partition 0: renamed keys (\texttt{m0} instead of \texttt{c0}), a reworded instruction, and the same prompts generated one at a time, which measures numerical noise from batched inference. Partition 0 is also cut into batches of 5, 10 and 50. Interventions use 200 targets per task, each placed in slot 12 and, separately, in slot 24 (the last) among 24 random neighbours; a second, independent set of neighbours supplies the swapped verdicts and texts.

\paragraph{Statistics and gates.} Intervals of flip rates, of the shares fixed before the runs and of the gates are percentile bootstraps over the batches of the base partition (300 resamples); they ignore the clustering of items in the batches of the second partition. The quantities of the formal framework from the main experiment (\S\ref{sec:where}, \S\ref{sec:voting}, Appendix~\ref{app:checks}) and from the larger models of Appendix~\ref{app:followup} average over the three base partitions, which hold the same items; their intervals resample the items once for all three (1{,}000 resamples; 10{,}000 for the bound of Equation~\eqref{eq:bound}) and ignore that items share generated prompts; \S\ref{sec:where} shows how much this matters. The intervals of the voting experiments of Appendix~\ref{app:followup} instead treat the items as clustered in the batches of every partition. All intervals are pointwise. With Holm's adjustment for the ten model--task pairs (two-sided tests at 5\%, normal approximation with the bootstrap standard deviation), the main effect of membership differs significantly from zero in \HolmMembMain{} pairs instead of \MembMainSig{} (for \HolmMembNarrow{} only narrowly, so that this count depends on the random draw of the bootstrap), the bound on the bias in \HolmBias{} instead of \BiasBoundSig, and the slot in \HolmSlotMain, as without the adjustment. Before the runs we fixed three gates. G1 (generality): on each of the two tasks added to harm (grading and stars), for at least three of the five models, the flip rate between two new partitions exceeds the largest batched baseline, with a lower bound of the excess of 2 points or more; the excess is computed between partitions 0 and 1. G2 (membership): the membership share at a fixed slot on the latent scale (\S\ref{sec:setup}) has a lower bound above 0 on at least two tasks for at least three models. G3 (mechanism): the unedited intervention reproduces the generated verdict except where two levels are nearly tied, and at least one channel changes 3\% of decisions or more.\footnote{G3 first required 99.5\% reproduction. Smoke tests before the full runs (two targets per task, each in slots 12 and 24: 12 placements per model) showed that in bf16, generation with a key-value cache in a padded batch and a single forward pass can differ in probability, by up to \SmokeTieMaxDiff{} on near-ties; in the smoke outputs that were kept, those of Qwen3-4B and Gemma-4 (\SmokePlacements{} placements), \SmokeClearCount{} placement of \SmokeClearModel{} also differs by \SmokeClearDiff{} without being a near-tie. Before the full runs we therefore added the exception for near-ties, and we measure every edit against the unedited forward pass, which shares its numerical regime.}

\section{Results}

\subsection{Decisions depend on the batch}
\label{sec:exists}

\begin{table*}[t]
\centering
\small
\setlength{\tabcolsep}{4pt}
\begin{tabular}{@{}l rrl rrl rrl@{}}
\toprule
& \multicolumn{3}{c}{Harm} & \multicolumn{3}{c}{Grading} & \multicolumn{3}{c}{Stars} \\
\cmidrule(lr){2-4}\cmidrule(lr){5-7}\cmidrule(l){8-10}
Model & new & base & excess [95\% CI] & new & base & excess [95\% CI] & new & base & excess [95\% CI] \\
\midrule
Qwen3-4B & 13.3 & 4.6 & +8.7 [+6.7, +10.6] & 12.9 & 4.7 & +8.2 [+6.7, +9.7] & 3.9 & 2.2 & +1.7 [+0.6, +2.8]$^\dagger$ \\
Qwen3-8B & 15.5 & 6.2 & +9.4 [+7.5, +11.4] & 11.7 & 4.6 & +7.1 [+5.6, +8.8] & 3.6 & 1.7 & +2.0 [+1.3, +2.9]$^\dagger$ \\
Gemma-4 & 13.0 & 4.4 & +8.6 [+6.5, +10.1] & 10.7 & 4.2 & +6.5 [+5.3, +8.1] & 2.9 & 1.4 & +1.6 [+0.8, +2.3]$^\dagger$ \\
Granite & 15.7 & 5.9 & +9.8 [+7.8, +11.5] & 21.7 & 8.4 & +13.3 [+11.1, +15.5] & 3.5 & 2.1 & +1.4 [+0.7, +2.3]$^\dagger$ \\
GLM-4 & 16.2 & 6.5 & +9.7 [+7.5, +12.0] & 15.6 & 8.0 & +7.6 [+5.9, +9.2] & 2.9 & 1.3 & +1.6 [+0.9, +2.3]$^\dagger$ \\
\bottomrule
\end{tabular}
\caption{Decisions that change when the same items are graded in a new random partition (\emph{new}, \%: partition 0 against partition 1, the pair on which the excess and gate G1 are computed; Table~\ref{tab:full} gives the mean over the three pairs), the largest batched baseline (\emph{base}: renamed keys, reworded instruction or one prompt at a time, \%), and the excess of \emph{new} over \emph{base} in points with its 95\% bootstrap interval, computed before rounding, so that it can differ by 0.1 from the difference of the printed columns. $^\dagger$: lower bound below 2 points.}
\label{tab:main}
\end{table*}

Table~\ref{tab:main} gives the main result. On harm and grading, a new random partition changes \NewBatchFlips\% of decisions for each of the five main models. The largest baseline, renamed keys or a reworded instruction, changes \BatchedBaselineRange\% (numerical noise alone, from generating the prompts one at a time: \NumericalRange\%), and the lower bound of the excess is \HGExcessLower{} points across the ten model--task pairs. On star ratings a new partition changes \StarsNewFlips\% of decisions, \StarsExcessRange{} points above the baselines, with lower bounds of \StarsExcessLower{} points. G1 as fixed before the runs is therefore met on grading (\GOneGrading{} of five models) but not on stars (\GOneStars), so we claim generality only for judgements with real ambiguity; on harm, the task we started from, all five models exceed the same bar. So do the four larger models of Appendix~\ref{app:followup}, run on harm and grading with the same partitions and baselines on H200 NVL GPUs: a new partition changes \ScaleNewRange\% of their decisions, \ScaleExcessRange{} points above the largest baseline, with lower bounds of about 3 points or more.

Looking at written levels rather than decisions, a new partition changes the level of \LevelChangeRange\% of items across all tasks, \LevelRatioRange{} times as often as rewording the instruction (Table~\ref{tab:levels}, Appendix~\ref{app:full}). On stars, \StarsAdjacentRange\% of these changes are between neighbouring ratings on the same side of the threshold.

\subsection{Accuracy understates it}
\label{sec:accuracy}

\begin{table}[t]
\centering
\small
\setlength{\tabcolsep}{2.8pt}
\begin{tabular}{@{}l rrc rrc@{}}
\toprule
& \multicolumn{3}{c}{Harm} & \multicolumn{3}{c}{Grading} \\
\cmidrule(lr){2-4}\cmidrule(l){5-7}
Model & alone & batch & flagged & alone & batch & flagged \\
\midrule
Qwen3-4B & 0.781 & 0.777 & 58\,$\to$\,60 & 0.796 & 0.805 & 43\,$\to$\,34 \\
Qwen3-8B & 0.784 & 0.763 & 64\,$\to$\,62 & 0.825 & 0.800 & 37\,$\to$\,35 \\
Gemma-4 & 0.827 & 0.835 & 49\,$\to$\,31 & 0.814 & 0.796 & 46\,$\to$\,48 \\
Granite & 0.824 & 0.834 & 59\,$\to$\,43 & 0.717 & 0.686 & 61\,$\to$\,59 \\
GLM-4 & 0.805 & 0.772 & 48\,$\to$\,48 & 0.803 & 0.779 & 47\,$\to$\,41 \\
\bottomrule
\end{tabular}
\caption{AUROC of the written level against human labels when graded alone and in batches of 25 (mean of three partitions), and the share of items above the decision threshold (\%, alone $\to$ batched).}
\label{tab:accuracy}
\end{table}

AUROC of the written level against human labels changes by \AurocDeltaRange{} (Table~\ref{tab:accuracy}). For \AurocDeltaSig{} of the ten model--task pairs, the change exceeds what resampling items would produce (paired item bootstrap), so accuracy does register batching, but a change of this size would pass most validation thresholds although \NewBatchFlips\% of the decisions differ between partitions. Rank agreement with the three-level human grades falls by up to \MaxSpearmanDrop{} on grading. The operating point moves more: on harm, \mbox{Gemma-4} and Granite flag far fewer comments in batches (Table~\ref{tab:accuracy}), as do \ScaleFlagDropCount{} of the four larger models of Appendix~\ref{app:followup} (by \ScaleFlagDropRange{} points), a shift that an accuracy check with a tuned threshold would also miss. On harm, accuracy even favours batching for the larger models: AUROC of the written level rises by \ScaleAurocHarmRange{} (clearly beyond resampling noise for \ScaleAurocHarmSig{} of them and only just for \ScaleAurocHarmBorder), because when they grade the comments alone they put many of them on one level, whose ties lower AUROC (Table~\ref{tab:scale}); on grading it changes by \ScaleAurocGradingRange. A validation on labelled data would thus approve batching on harm although \ScaleHarmNewRange\% of their harm decisions change between partitions. An earlier study of a partly overlapping sample of 2{,}000 random comments (\PopOverlap{} in common with the main sample) with the same prompt (Appendix~\ref{app:population}) gives the same picture: batching changed \PopBatchedRange\% of decisions for the five main models, against \PopRenamedRange\% for renamed keys. There, models also shift in different directions on the same comments: Gemma-4 flags \GemmaRaceAlone\% of the comments that target a race when it grades them alone and \GemmaRaceBatched\% in batches, while GLM-4 flags more of them in batches (Table~\ref{tab:groups}; point estimates).

Two facts explain why accuracy is a poor test (Appendix~\ref{app:proofs}). If $e_A$ and $e_B$ are the shares of decisions in two conditions that differ from the labels and $e_{AB}$ the share on which both do, the flip rate between the conditions is $e_A+e_B-2e_{AB}$, the errors that they do not share (Proposition~\ref{prop:acc}): with error rates of \ErrorRange\% at the decision threshold, equal accuracy is compatible with any flip rate from 0 to \ErrorBoundRange\%. And under a threshold model, a context that carries a share of the variance of the scores within a class lowers AUROC, relative to the scores averaged over contexts, in proportion to this share (Proposition~\ref{prop:auc}), but flips decisions in proportion to its square root (Proposition~\ref{prop:law}).

\subsection{Where it comes from}
\label{sec:where}

\begin{figure*}[t]
\centering
\includegraphics[width=0.92\textwidth]{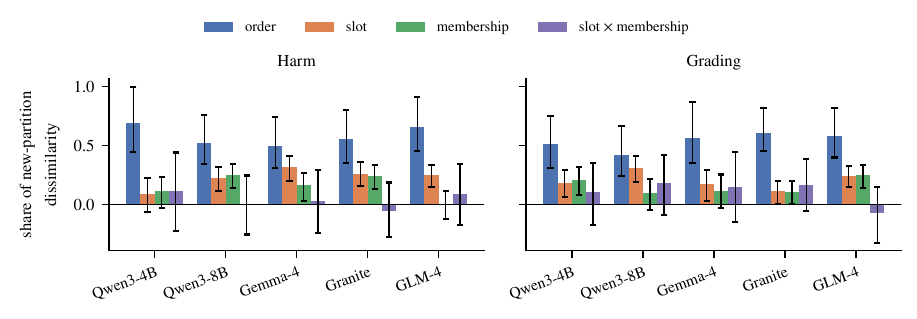}
\caption{Components of the latent dissimilarity between two random partitions, as shares of $D_\mathrm{new}$ (Proposition~\ref{prop:pf}; 95\% bootstrap intervals): the order of the neighbours $T_\omega$, the main effects of the slot $V_s$ and of the membership $V_N$, and their interaction $V_{sN}$. The four shares add up to 1. The order is estimated from the designated slots of ``keep slot, reorder the others'', weighted as in a random partition (\S\ref{sec:setup}).}
\label{fig:components}
\end{figure*}

Figure~\ref{fig:components} decomposes the latent dissimilarity between two random partitions, $D_\mathrm{new}$, which is \DNewRange{} on harm and grading. The order of the neighbours accounts for \OrderShareRange{} of it, the slot for \SlotMainRange, the main effect of membership for \MembMainRange{} (mean \MembMainMean) and the interaction of slot and membership for \SlotMembRange{} (mean \SlotMembMean). Proposition~\ref{prop:pf} requires that no component is negative; \ComponentsNegative{} of the thirty estimates of the last three are negative (the most negative is \NegativeSmallest), and their intervals include zero. The intervals exclude zero for the slot in \SlotMainSig, for the main effect of membership in \MembMainSig{} and for the interaction in \SlotMembSig{} of the ten model--task pairs\SigBorder. These intervals resample items. Treating the items that share a generated prompt as clustered as well, with a linearised two-way cluster variance \citep{cameron2011robust}, multiplies the standard deviation of the main effect of membership by \ClusterRatioMembMain; scaling the distances of the item-bootstrap limits from the estimate by this factor, its interval excludes zero in \MembMainSigCl{} pairs \MembMainClNote, and the other counts do not change. On the scale of decisions, where the decomposition needs no threshold model (the split between order and interaction still rests on the designated slots), the order accounts for \DecisionOrderRange{} of the flips and the main effect of membership for \DecisionMembRange{} (Table~\ref{tab:components}). We estimate $D_\mathrm{new}$ from the three pairs of partitions. Any two of the nine batched conditions (the partitions, their reordered batches and their batches with new neighbours) give an item independent contexts if they come from different base partitions; with $D_\mathrm{new}$ estimated from all 27 such pairs, the main effect of membership accounts for \MembMainPooledRange{} of it (mean \MembMainPooledMean) instead of \MembMainRange{} (mean \MembMainMean).

The membership share at a fixed slot as we fixed it before the runs, $(D_\mathrm{comp}-D_\mathrm{keep})/D_\mathrm{new}$ on the latent scale, is \MembershipRange, and its lower bound is above 0 in every pair, so G2 is met as specified. This measure, however, gives the first slot a fifth of the weight and so overstates the effect of membership (\S\ref{sec:setup}). With the designated slots weighted as in a random partition, it is \MembFixedRange{} (mean \MembFixedMean), and its lower bound is above 0 in \MembFixedSig{} of the ten pairs and on both tasks for \MembFixedModels{} of the five models, so G2 would \GTwoWeighted. This verdict depends on how the order component is estimated. Other items also keep their slot and their membership: those that keep their slot by chance in the reordered batch, which measure the order component exactly in expectation (Proposition~\ref{prop:conditions}(iv)), and the other items of ``keep slot, reorder the others'' that land in their own slot, which give a lower bound like the designated ones. With both added, and the first, the last and the interior slots weighted $1/25$, $1/25$ and $23/25$ as in a random partition (evenly over the items of each group), the share is \MembPoolRange, its lower bound is above 0 in \MembPoolSig{} pairs and on both tasks for \MembPoolModels{} models, and G2 would be met; adding only one of the two gives at most \MembPoolOneMax{} models, a count that also depends on the random draw of the bootstrap. Once the first slot is weighted correctly, G2 is therefore borderline. On average, more of this share is the main effect of membership (mean \MembMainMean) than the interaction of slot and membership (mean \SlotMembMean), but the interaction is at least as large in \SlotMembGeMain{} of the ten pairs, and on the scale of decisions the two means are about equal (\MembMainDecMean{} and \SlotMembDecMean). The main effect needs no designated slots, is positive in all ten pairs and excludes zero in \MembMainSig{} (in \MembMainSigCl{} with clustered prompts). The voting experiment of \S\ref{sec:voting}, which grades every batch of partitions 0 and 1 in nine orders, estimates it more precisely: \VoteVNLatentRange{} of $D_\mathrm{new}$ (pooled \VoteVNLatentPooled), and its intervals, clustered by the batches of both partitions, exclude zero in all ten pairs (Appendix~\ref{app:followup}). Which neighbours an item has therefore matters, less than the measure fixed before the runs suggested, and partly in a way that a vote over permutations does not average out. Order and slot together account for the larger part, which is consistent with earlier findings on position \citep{lin2024batchprompt,frohmann2026equal}. On stars, $D_\mathrm{new}$ is only \StarsDRange{} (flip rates of \StarsNewFlipsMean\%), too small to split into components with useful precision; we decompose only when $D_\mathrm{new}\ge0.01$. We also leave out \mbox{Gemma-4-31B} of Appendix~\ref{app:followup} ($D_\mathrm{new}$ of \GemmaLargeDNew): its estimate of the interaction on grading violates Proposition~\ref{prop:pf}.

The threshold model behind the latent scale makes predictions for patterns of decisions to which it was not fitted (Proposition~\ref{prop:patterns} in Appendix~\ref{app:proofs}, Table~\ref{tab:modelcheck}). On harm and grading, fitted to pairs of conditions, it predicts the share of items whose decisions in six batches, which we treat as independent, are not all equal (\CheckSplitObs\% of the items) with a mean absolute error of \CheckSplitErr{} points, the flips between two majority votes over three batches each with a mean absolute error of \CheckMajErr{} points, and the share of items for which two orders of one batch agree, two orders of another batch agree, and the two batches differ (\CheckRepObs\%) with a mean absolute error of \CheckRepErr{} points. The errors of the first two have a direction: the model predicts too few items whose decisions are not all equal in \CheckSplitUnder{} of the ten pairs and too many flips between majority votes in \CheckMajOver{}, in \CheckMajSig{} of them by more than two standard errors; averaged over the ten pairs, the first prediction is \CheckSplitUnderMean{} points too low and the second \CheckMajOverMean{} points too high. Without a main effect of membership, the prediction for the last pattern is below the observed share in \CheckRepLow{} of the ten pairs. Refitted to fifteen independent partitions (\S\ref{sec:voting}), the model shows the same two directions, and the first error, for the items whose decisions are not all equal, grows with the number of batches: pooled over the ten pairs, measured minus predicted is \BvoteUFive, \BvoteUNine{} and \BvoteUFifteen{} points for five, nine and fifteen batches, and \BvoteGThree{} points for the flips between two majorities over three batches. The model has limits: its dissimilarity depends on the threshold of the scale, by a factor of \ThreshHarmRatio{} between thresholds on harm and \ThreshStarsRatio{} on stars, where the primary threshold gives the smallest value for \ThreshStarsPrimaryMin{} of the five models (Table~\ref{tab:thresholds}).

\subsection{What voting leaves}
\label{sec:voting}

Voting over permutations of a batch averages over order and slot, and with them over the interaction of slot and membership; it leaves the main effect of membership (Proposition~\ref{prop:vote}). We ran it for the five models on H200 NVL GPUs: every batch of partitions 0 and 1 was graded in nine orders, the first being that of the main experiment (Appendix~\ref{app:followup}, Table~\ref{tab:votemeasured}). Majority votes over three, five and nine permutations leave \VoteMeasThreeRange\%, \VoteMeasFiveRange\% and \VoteMeasNineRange\% of the flips between the two partitions (after nine, \VoteMeasNinePooled{} pooled over the ten model--task pairs), at three, five and nine times the cost. Fitted to the same runs, the threshold model predicts the flip rates after three to nine permutations within one point in every pair, slightly too high on average (by \VoteInSampleGap{} points pooled); as shares of the flips of single orders, its predictions differ from the measured ones by up to \VoteInSampleRatioMax{} points (Table~\ref{tab:votemeasured}). The predictions of Table~\ref{tab:voting}, made from the main experiment before these runs, are too high by \VoteTwelveMean{} points on average after nine permutations, but per pair they miss by \VoteTwelveMiss{} points (predicted minus measured; as shares of the flips of single orders, by \VoteTwelveRatioMiss{} points, Table~\ref{tab:votemeasured}), mainly because the main design estimates the main effect of membership imprecisely; the measured flips lie inside their intervals in \VoteTwelveInside{} of the ten pairs. A majority vote removes flips more slowly than the mean score of the same permutations would (Appendix~\ref{app:proofs}), because it discards how far each score lies from the threshold: without a main effect of membership ($D_\mathrm{resh}$ set to $D_\mathrm{new}$, which keeps the flips of single batches), five and nine permutations would still leave \VoteFiveNull\% and \VoteNineNull\% of the flips, against \VoteFiveAvg\% and \VoteNineAvg\% for the mean score, which approaches $1/\sqrt{m}$, 45\% and 33\%, as $D\to0$. In the voting runs, averaging the level probabilities over nine orders leaves \VoteSoftNineRange\% of the flips that the same rule gives with single orders, since the main effect of membership remains. With more permutations the flips decrease towards a limit, which no odd number of permutations reaches (Proposition~\ref{prop:vote}) and which the model fitted to the voting runs puts at \VoteLimitRange\% of the items.

Voting over batches instead of permutations averages over the membership as well, but it converges to the main batched decision, not to the single-item decision (Proposition~\ref{prop:bias}). On harm and grading, a random batch gives \BiasLossRange\% of the items a decision that differs from the one they get alone. For \BiasAllThreeRange\% of the items all three partitions differ from it, so by Equation~\eqref{eq:bound} the mean bias is at least \BiasBoundRange\%, without any model; the interval of this bound excludes zero in \BiasBoundSig{} of the ten pairs. Under the threshold model the mean bias is \BiasModelRange\% (Proposition~\ref{prop:biasgauss}, Table~\ref{tab:bias}). On stars the bias under the model, \BiasStarsModel\%, is larger than the share of decisions that change between two partitions (\BiasStarsFlips\%, mean of the three pairs) for \BiasStarsAbove{} of the five models and about as large for \BiasStarsExceptions; without a model, the bound of Equation~\eqref{eq:bound} on stars is \BiasStarsBound\%, and its interval excludes zero for \BiasStarsBoundSig{} of the five models.

We measured what voting over batches leaves on harm and grading: partition 2 and twelve further random partitions, graded on the same type of GPU as the voting runs, give every item, together with partitions 0 and 1, fifteen independent batches (Appendix~\ref{app:followup}, Table~\ref{tab:bvote}). Their majority still differs from the single-item decision for \BvoteFifteenRange\% of the items (pooled \BvoteFifteenPooled), against \BvoteOneRange\% for one batch (pooled \BvoteOnePooled\%), so fifteen batches remove only about a fifth of the disagreement. The threshold model fitted to the main experiment predicted the majority of fifteen batches with a mean absolute error of \BvoteModelMAE{} points (at most \BvoteModelMax). Since three of the fifteen partitions and the single items repeat the prompts it was fitted to, this check is only partly out of sample: on the twelve new partitions alone its error reaches \BvoteNewOnlyMax{} points, and the model-free majority of the nine batched decisions of Table~\ref{tab:bias} predicts the fifteen-batch majority about as well (mean absolute error \BvoteNineMAE{} points). With the batched threshold chosen in every partition so that batches give as many positive decisions as single-item grading, the majority still differs for \BvoteCalRange\% of the items (pooled \BvoteCalPooled, \BvoteCalShare\% of the uncalibrated share); only for \mbox{Gemma-4} and Granite on harm is about half of the disagreement a shift of the operating point. Batching is therefore not only noise around the single-item decision: most of the disagreement is specific to the item, and voting over batches does not remove it. Single-item grading, however, is no privileged context: after the calibration it differs from the majority of thirteen of the fifteen batches about as often as one batch differs from the majority of thirteen of the other fourteen (pooled \CalSingleThirteen\% against \CalHeldThirteen\%; per pair from \CalLessMax{} points less often, for \CalLessPair, to \CalMoreMax{} points more often, for \CalMorePair). Nor does disagreement mean error: on the items where the majority of fifteen batches and the single item differ, the majority matches the human label for more than half of them in \BvoteLabelCount{} of the ten pairs (for up to \BvoteLabelMax\%). The measured disagreement also depends on the single-item prompt: with the reworded single-item prompt of the RTX 5090 runs as the reference, it changes by \BvoteParaShiftRange{} points per pair, although pooled only from \BvoteAlonePooledBoth\% to \BvoteParaPooled\%; under the model, the mean bias changes by up to \BiasParaShift{} points.

\subsection{Two channels: texts and earlier verdicts}
\label{sec:how}

\begin{figure*}[t]
\centering
\includegraphics[width=0.92\textwidth]{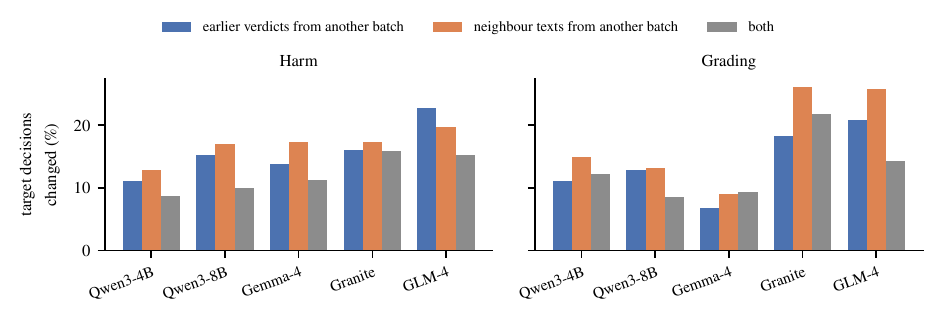}
\caption{Target decisions changed by editing one part of a generation, relative to the unedited forward pass: the earlier verdicts taken from a generation with other neighbours (texts kept), the neighbours' texts taken from that generation (verdicts kept), or both.}
\label{fig:interventions}
\end{figure*}

On harm and grading, the unedited intervention reproduces the generated verdict for \OwnReproduced\% of the placements (a target in one slot): greedy decoding of the unedited forward pass, with exact ties broken as in the model's generations, writes the same level. Most of the misses are near-ties, with the two most likely levels of the generation within 0.2 in probability, but \OwnClearMisses{} of the \OwnMisses{} misses on harm and grading are not (for \OwnClearModels, at most \OwnClearMax{} of about 400 placements per model and task), so the first condition of G3 holds only approximately. Figure~\ref{fig:interventions} shows what changes when one channel is set to its value in a batch with other neighbours. Replacing only the earlier verdicts changes \OtherVerdicts\% of target decisions, replacing only the neighbours' texts \OtherTexts\%; both channels act with similar strength, well above the 3\% that G3 requires. Replacing both, which puts the other batch's texts and verdicts into the answer of this generation, changes \BothRange\%: less than the sum of the two in every pair and less than either alone in \BothBelowEither{} of ten, so the channels overlap, and texts and verdicts that do not match can disturb the model more than a consistent context does.

The verdict channel responds to its content, not only to its presence (Table~\ref{tab:interventions}, Appendix~\ref{app:interventions}). Lowering the earlier verdicts by one level, which leaves those at the lowest level unchanged, changes \DownRange\% of decisions, while shuffling the model's own earlier verdicts changes \ShuffledRange\%. Replacing them with the verdicts the model gives to the same neighbours alone changes only \AloneVerdictRange\%. This is consistent with part of the effect travelling through the difference between the model's batched and single-item verdicts on the neighbours, although verdicts that match the texts may also simply disturb the model less. Replacing every neighbour's text by a dash, with the keys and the earlier verdicts kept, changes \FillerRange\%, and letting the target be answered first changes \FirstRange\%.

The changes of the target's probability make the same point (Proposition~\ref{prop:channels}, Table~\ref{tab:channels}). If the channels added up on the scale of $\zeta$, the interaction contrast $J$ would vanish. Instead its mean square is \ChanRatioRange{} times the mean of those of $\Delta_T$ and $\Delta_V$, between 1.8 and 2.6, close to the factor 2 of four unrelated contexts, in \ChanNearTwo{} of the ten pairs, and the two changes correlate at \ChanCorrRange, where unrelated contexts give 0.5. On the scale of the target's probability, the texts and the verdicts of a batch therefore do not contribute separate amounts, and changing either one is almost as much as drawing a new context. Nor do they add up on a latent scale that saturates near 0 and 1. If they added up on any scale that $\zeta$ increases with, a change of the texts could not move $\zeta$ in opposite directions under the two sets of verdicts, nor a change of the verdicts under the two sets of texts, so that the two consistent values $\zeta_{AA}$ and $\zeta_{BB}$ (the first index names the batch of the texts, the second that of the verdicts) could never both lie above, or both below, the two mismatched values $\zeta_{BA}$ and $\zeta_{AB}$ (Proposition~\ref{prop:channels}). Yet for \ChanCrossedAll\% of the \ChanTargets{} placements graded by the five models (a target in one slot, graded by one model; \ChanCrossedRange\% per pair) both consistent values lie on the same side of both mismatched ones by more than \ChanCrossMargin. These placements carry \ChanCrossedJ\% of the summed $J^2$, and in \ChanCrossedUndecided\% of them the mismatched values are the ones closer to 0.5. Numerical noise can explain only part of this. If the channels added up, a crossing by more than \ChanCrossMargin{} would need an error of more than half as much in one of the four values. Errors of the two consistent values explain at most \ChanSecondLost{} of the crossings: replacing these values by the probabilities read during the two generations (for $\zeta_{BB}$, the other generation in its own layout) leaves \ChanSecondCrossed{} of the \ChanCrossedTargets{} placements crossed, and replacing only $\zeta_{AA}$ leaves \ChanFirstCrossed. With the consistent values right, a crossing needs an error of more than \ChanCrossMargin{} in one of the two mismatched values. The unedited forward pass and the probability read during generation differ by that much for \ChanNoiseHugeUndecided\% of the placements that are undecided at generation ($0.1<\zeta<0.9$) and for \ChanNoiseHugeDecided{} of the \ChanNoiseDecidedN{} others. At these rates, applied to the level of each mismatched value of every placement, such errors are expected in at most \ChanSpurious{} of the \ChanTargets{} placements, so numerical noise can account for at most about \ChanNoiseTotal{} of the \ChanCrossedTargets{} crossings, or \ChanNoiseTotalUpper{} at the upper ends of the 95\% intervals of the two rates, which rest on \ChanNoiseHugeCount{} differences. Texts and verdicts that do not match leave the model less certain, although they rarely reverse, both at once, the decision that the two consistent contexts give: the four decisions cross in this way for only \ChanDecCrossed\% of the placements.

\subsection{When it matters}
\label{sec:when}

\begin{figure}[t]
\centering
\includegraphics[width=\columnwidth]{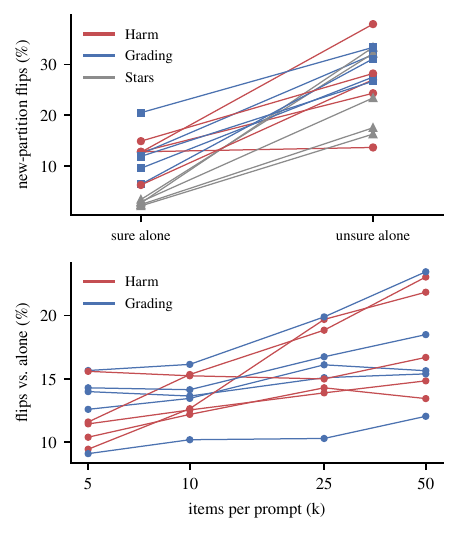}
\caption{Top: flips between partitions 0 and 1 for items the model is sure or unsure about when graded alone (unsure: $0.1<\zeta<0.9$ in the single-item prompt, where $\zeta=P(\text{level}\ge\tau)$ is read from the next-token distribution as in \S\ref{sec:setup}); one line per model and task. Bottom: flips against single-item grading as the batch grows, harm and grading, one line per model and task.}
\label{fig:when}
\end{figure}

On harm and grading, confidence on single items protects only partly against batching (Figure~\ref{fig:when}, top). Items the model is unsure about when graded alone flip \UnsureFlips\% of the time between partitions, but items it is sure about still flip \SureFlips\%, and the share of unsure items varies widely between models (\UnsureShareRange\%). Star ratings are easy for all five main models (AUROC \StarsAurocRange{} against the true rating, compared with \AurocRange{} on the other tasks): few reviews are uncertain (\StarsUnsureShare\%), and sure reviews flip only \StarsSureFlips\%. This makes star rating the boundary condition of our findings: batching changes the written rating but rarely a decision. For the five main models, stars also have the smallest dissimilarity on the latent scale: at the primary threshold $D_\mathrm{new}$ is \StarsDRange{} against \DNewRange{} on the other tasks, and at no threshold of the star scale does it exceed \StarsDMax, less than at any threshold of harm and grading with at least 5\% of the decisions on either side (\OtherDMin). How much smaller it is depends on the threshold, which an ordinal version of the threshold model would not allow (Table~\ref{tab:thresholds}, Appendix~\ref{app:checks}).

Smaller batches help less than one might hope (Figure~\ref{fig:when}, bottom). With five items per prompt, \KFiveRange\% of decisions already differ from single-item grading; with fifty, \KFiftyRange\%. Within the Qwen3 family, size helps more against trivial prompt changes than against batching: renamed keys change \PopRenamedQwenSmall\%, \PopRenamedQwenMid\% and \PopRenamedQwenLarge\% of decisions at 1.7B, 4B (the later Instruct-2507 release) and 8B, while batching changes \PopBatchedQwenSmall\%, \PopBatchedQwenMid\% and \PopBatchedQwenLarge\% (Appendix~\ref{app:population}); at 1.7B renamed keys change about as many decisions as batching, and from 4B to 8B renamed keys change a third as many decisions while batching changes about as many. On the main harm sample, \mbox{Qwen3-14B} and \mbox{Qwen3-32B} do not continue this pattern (Appendix~\ref{app:followup}): renamed keys change \ScaleQwenRenamed\% of their single-item decisions, no fewer than for \mbox{Qwen3-8B} (Table~\ref{tab:full}), while a batch differs from single-item grading on \ScaleQwenFourteenAlone\% and \ScaleQwenThirtyTwoAlone\%.

\section{Implications}

\paragraph{Validation.} Accuracy on a labelled sample understates how much batching changes. A batched pipeline should report how many decisions differ from single-item grading and between two random partitions, next to baselines such as renamed keys, which need no labels.

\paragraph{Mitigation.} Voting over permutations leaves the main effect of membership and multiplies the cost. It divides the variance that order and slot, and their interaction with membership, add to the share of votes by the number of permutations (Proposition~\ref{prop:vote}), but the flips by less than its square root: measured, five permutations leave \VoteMeasFiveRange\% of the flips between batches and nine \VoteMeasNineRange\% (\S\ref{sec:voting}). Voting over batches converges to the main batched decision (Proposition~\ref{prop:bias}): the majority of fifteen independent batches still differs from the single-item decision for \BvoteFifteenRange\% of the items (\S\ref{sec:voting}). In our earlier study, averaging six batches over three slots still left 10--24\% of borderline decisions changed (Appendix~\ref{app:replication}). Recalibrating the threshold so that batches flag as many items as single-item grading removes only the part of the disagreement that is a shift of the operating point: about half of it for \mbox{Gemma-4} and Granite on harm and less in the other pairs of the main models. The recalibrated majority of fifteen batches still differs from single-item grading on \BvoteCalRange\% of decisions (\S\ref{sec:voting}), and for the closed model of Appendix~\ref{app:replication} recalibration left the changed decisions as they were. The reliable fix is separability by construction: one item per prompt, or a block attention mask under which each item and its verdict see only the shared instructions and that item, with keys and position ids that do not depend on the slot \citep{bito2026onepass,glavas2026ippd}.

\paragraph{Cost.} With a shared prefix of $S$ tokens, $L$ tokens per item and cached tokens billed at a fraction $c_\mathrm{cache}$ of the input price, a single-item prompt with a cached prefix costs $c_\mathrm{cache}S+L$ input tokens per item, ignoring output tokens and the one-time cost of writing the cache. A batch costs $S/k+L$ per item if its prefix is billed in full, and $c_\mathrm{cache}S/k+L$ if it is read from the cache as well. Isolation therefore costs $(c_\mathrm{cache}-1/k)S$ more per item than batches without a cache, which is a saving when $c_\mathrm{cache}<1/k$, and $c_\mathrm{cache}(1-1/k)S$ more than batches with one. This requires all shared text, including the request, to precede the item; our prompts put the request after the items and would need to be reordered. Anthropic's models and OpenAI's models from GPT-5.6 on bill cache reads at $c_\mathrm{cache}\le0.1$ (earlier OpenAI models at a model-dependent rate) but cache only prefixes of at least 512--4{,}096 tokens, depending on the model \citep{anthropic2026caching,openai2026caching}; self-hosted serving reuses prefixes of any length \citep{kwon2023vllm,zheng2024sglang}, which makes $c_\mathrm{cache}$ nearly zero, so that isolation costs about as much as batches with a cached prefix and less than batches without one. When the instructions are short and the items unambiguous, as for star ratings, batching may be acceptable, but that should be measured rather than assumed.

\section{Conclusion}

Batching several items into one prompt makes an LLM's verdict on an ambiguous item depend on its company. On harm and short-answer grading, a new random partition changes about one decision in seven for five 4\nobreakdash--9B models (\NewBatchMean\% on average) and one in ten for four 14\nobreakdash--32B models (\ScaleNewMean\%), mostly through the order of the neighbours and the slot but also through which items share the batch; in the 4\nobreakdash--9B models, whose generations we edited, the effect travels through both the neighbours' texts and the model's own earlier verdicts. Because, under a threshold model, flips grow with the square root of the variance that the context contributes, accuracy understates the effect and voting removes it only slowly: for the 4\nobreakdash--9B models, nine permutations of a batch leave more than half of the flips. Because batched verdicts differ systematically from single-item verdicts, even the majority of fifteen independent batches does not restore them. Where decisions matter to the people behind the items, graders should keep items apart or at least measure what batching changes.

\section*{Limitations}

Our main models are open-weight and have 4\nobreakdash--9B parameters. Four larger ones, with 14\nobreakdash--32B parameters, repeat only the core design on harm and grading, on another type of GPU, without star ratings, other batch sizes, interventions or voting; the Qwen3 family has two of them, the Gemma-4 and GLM-4 families one each and the Granite family none. Proprietary models may be more or less sensitive. Within Qwen3 our models run from 1.7B to 32B, and the 4B model is a later release. The closed model of Appendix~\ref{app:replication} answers per item against a shared context rather than in one JSON generation, so it does not test the verdict channel. We study three English tasks with one prompt template and one reworded variant, JSON output and greedy decoding; under the reworded variant a new partition still changes \RewordedAllRange\% of decisions on harm and grading, slightly fewer than under the original wording (Appendix~\ref{app:followup}). Other formats, such as asking for one verdict at a time within a conversation, may behave differently. The decomposition of the flip rate is exact in expectation for the main effect of membership, up to the few items without a valid verdict. The split between order and the interaction of slot and membership rests on the five designated slots of ``keep slot, reorder the others'', which reorder only 20 of the 24 neighbours and which we reweight because the first slot is less sensitive than the others; the membership share that we fixed before the runs does not allow for this and overstates the effect of membership. The latent scale, the model's predictions for voting and its estimate of the bias assume a Gaussian threshold model in which the context contributes the same variance for all items, and the estimate of the bias also assumes that the single-item score is jointly normal with the mean of the item over contexts; what voting over up to nine permutations and fifteen batches leaves is measured without the model, for the five main models on harm and grading only (on stars the bias rests on the model and on three partitions). The model predicts, with a mean absolute error below one point, most patterns of the main experiment to which it was not fitted, but it over-predicts the flips between majority votes and under-predicts the items whose decisions in fifteen batches are not all equal (by \BvoteUFifteen{} points, measured minus predicted; \S\ref{sec:where}), which suggests that the context matters more for some items than for others. Fitted to the voting runs, it predicts within a point the flip rates that permutation voting leaves; its predictions from the main design, made before these runs, missed by \VoteTwelveMiss{} points (predicted minus measured; \S\ref{sec:voting}). The latent dissimilarity depends on the threshold of the scale, by a factor of \ThreshHarmRatio{} on harm and \ThreshStarsRatio{} on stars. The interventions edit tokens inside a generation: the most extreme edits (all earlier verdicts at the lowest level) produce contexts the model would not write itself, and we base our conclusions about the two channels on swaps between natural generations. Most intervals treat items as independent; treating the items that share a generated prompt as clustered reduces the pairs whose interval for the main effect of membership excludes zero from \MembMainSig{} to \MembMainSigCl, and the clustered intervals of the voting experiment exclude zero in all ten. Finally, bf16 inference makes the edited forward pass differ slightly from generation with a cache; the unedited check reproduces \OwnReproduced\% of verdicts on harm and grading, and all effects are measured against it.

\section*{Ethical Considerations}

The harm task uses comments that contain hate speech; we do not quote any of them. Measuring Hate Speech is released under CC BY 4.0 and SciEntsBank under CC BY-SA 3.0 (the Hugging Face copy that we used is labelled CC BY 4.0). Yelp Review Full is used under the Yelp Dataset Terms of Use; we submitted our findings to Yelp for review before publication. For its reviews we release only their row numbers in the test set, not their texts, star ratings or model outputs. Our findings concern decisions that affect people, such as removing a comment or grading an answer; they argue for measuring the arbitrariness that batching introduces before deploying it. The main experiment ran on five rented RTX 5090 instances, one per model, and the population study of Appendix~\ref{app:population} on two rented RTX 5090 instances, one model at a time; the open-model runs of Appendix~\ref{app:replication} used a shared RTX 5090 server and rented instances, and the closed model its API. The follow-up experiments of Appendix~\ref{app:followup} ran on a shared university cluster, one H200 NVL per model, for about 15 GPU-hours.

\section*{Data and Code Availability}

The experimental designs, the runner and analysis scripts, and the model outputs behind the tables and figures (parsed verdicts, level distributions and JSON answers), except the outputs for the Yelp reviews, will be available at \url{https://github.com/PX-CPS/Judged_in_company}, together with the designs, job scripts and outputs of the follow-up experiments of Appendix~\ref{app:followup}. Without the outputs for the Yelp reviews, the star-rating results and the few numbers that pool the three tasks (the changes of the written level in \S\ref{sec:exists}, the smoke tests of the footnote on G3 and the counts of valid verdicts in Appendix~\ref{app:proofs}) can be reproduced only by running the models on the corpus with the released designs and runner. Appendix~\ref{app:replication} and the two statements of the paragraph on mitigation that cite it rest on our earlier study of Chinese moderation data, with the closed model and open models, whose data we do not release. The texts of the items come from the original corpora and are not redistributed, except the grading item shown in Figure~\ref{fig:example}.

\bibliography{references}

\begin{thebibliography}{59}
\providecommand{\natexlab}[1]{#1}

\bibitem[{{Anthropic}(2026{\natexlab{a}})}]{anthropic2026moderation}
{Anthropic}. 2026{\natexlab{a}}.
\newblock Content moderation.
\newblock
  \url{https://platform.claude.com/docs/en/about-claude/use-case-guides/content-moderation}.
\newblock Accessed 2026-09-28.

\bibitem[{{Anthropic}(2026{\natexlab{b}})}]{anthropic2026caching}
{Anthropic}. 2026{\natexlab{b}}.
\newblock Prompt caching.
\newblock
  \url{https://platform.claude.com/docs/en/build-with-claude/prompt-caching}.
\newblock Accessed 2026-09-29.

\bibitem[{Bito et~al.(2026)Bito, Ren, and He}]{bito2026onepass}
Ethan Bito, Yongli Ren, and Estrid He. 2026.
\newblock \href {https://doi.org/10.1145/3805712.3809952} {One pass, any order:
  Position-invariant listwise reranking for {LLM}-based recommendation}.
\newblock In \emph{Proceedings of the 49th International ACM SIGIR Conference
  on Research and Development in Information Retrieval}, SIGIR '26, pages
  3625--3629, New York, NY, USA. Association for Computing Machinery.

\bibitem[{Black et~al.(2022)Black, Raghavan, and Barocas}]{black2022model}
Emily Black, Manish Raghavan, and Solon Barocas. 2022.
\newblock \href {https://doi.org/10.1145/3531146.3533149} {Model multiplicity:
  Opportunities, concerns, and solutions}.
\newblock In \emph{Proceedings of the 2022 ACM Conference on Fairness,
  Accountability, and Transparency}, FAccT '22, pages 850--863, New York, NY,
  USA. Association for Computing Machinery.

\bibitem[{Cameron et~al.(2011)Cameron, Gelbach, and Miller}]{cameron2011robust}
A.~Colin Cameron, Jonah~B. Gelbach, and Douglas~L. Miller. 2011.
\newblock \href {https://doi.org/10.1198/jbes.2010.07136} {Robust inference
  with multiway clustering}.
\newblock \emph{Journal of Business \& Economic Statistics}, 29(2):238--249.

\bibitem[{Chen et~al.(2016)Chen, Moskowitz, and Shue}]{chen2016gambler}
Daniel~L. Chen, Tobias~J. Moskowitz, and Kelly Shue. 2016.
\newblock \href {https://doi.org/10.1093/qje/qjw017} {Decision making under the
  gambler's fallacy: Evidence from asylum judges, loan officers, and baseball
  umpires}.
\newblock \emph{The Quarterly Journal of Economics}, 131(3):1181--1242.

\bibitem[{Chen et~al.(2024)Chen, Liu, Dong, Liu, Sakai, and
  Wu}]{chen2024threshold}
Nuo Chen, Jiqun Liu, Xiaoyu Dong, Qijiong Liu, Tetsuya Sakai, and Xiao-Ming Wu.
  2024.
\newblock \href {https://doi.org/10.1145/3673791.3698420} {{AI} can be
  cognitively biased: An exploratory study on threshold priming in {LLM}-based
  batch relevance assessment}.
\newblock In \emph{Proceedings of the 2024 Annual International ACM SIGIR
  Conference on Research and Development in Information Retrieval in the Asia
  Pacific Region}, SIGIR-AP 2024, pages 54--63, New York, NY, USA. Association
  for Computing Machinery.

\bibitem[{Cheng et~al.(2023)Cheng, Kasai, and Yu}]{cheng2023batch}
Zhoujun Cheng, Jungo Kasai, and Tao Yu. 2023.
\newblock \href {https://doi.org/10.18653/v1/2023.emnlp-industry.74} {Batch
  prompting: Efficient inference with large language model {API}s}.
\newblock In \emph{Proceedings of the 2023 Conference on Empirical Methods in
  Natural Language Processing: Industry Track}, pages 792--810, Singapore.
  Association for Computational Linguistics.

\bibitem[{Cooper et~al.(2024)Cooper, Lee, Choksi, Barocas, De~Sa, Grimmelmann,
  Kleinberg, Sen, and Zhang}]{cooper2024arbitrariness}
A.~Feder Cooper, Katherine Lee, Madiha~Zahrah Choksi, Solon Barocas,
  Christopher De~Sa, James Grimmelmann, Jon Kleinberg, Siddhartha Sen, and
  Baobao Zhang. 2024.
\newblock \href {https://doi.org/10.1609/aaai.v38i20.30203} {Arbitrariness and
  social prediction: The confounding role of variance in fair classification}.
\newblock \emph{Proceedings of the AAAI Conference on Artificial Intelligence},
  38(20):22004--22012.

\bibitem[{Cram{\'e}r(1936)}]{cramer1936eigenschaft}
Harald Cram{\'e}r. 1936.
\newblock \href {https://doi.org/10.1007/BF01180430} {{\"U}ber eine
  {Eigenschaft} der normalen {Verteilungsfunktion}}.
\newblock \emph{Mathematische Zeitschrift}, 41(1):405--414.

\bibitem[{Deng et~al.(2022)Deng, Zhou, Sun, Zheng, Mi, Meng, and
  Huang}]{deng2022cold}
Jiawen Deng, Jingyan Zhou, Hao Sun, Chujie Zheng, Fei Mi, Helen Meng, and
  Minlie Huang. 2022.
\newblock \href {https://doi.org/10.18653/v1/2022.emnlp-main.796} {{COLD}: A
  benchmark for {C}hinese offensive language detection}.
\newblock In \emph{Proceedings of the 2022 Conference on Empirical Methods in
  Natural Language Processing}, pages 11580--11599, Abu Dhabi, United Arab
  Emirates. Association for Computational Linguistics.

\bibitem[{Domingos(2000)}]{domingos2000unified}
Pedro Domingos. 2000.
\newblock \href
  {https://aaai.org/papers/00564-aaai00-086-a-unified-bias-variance-decomposition-for-zero-one-and-squared-loss/}
  {A unified bias-variance decomposition for zero-one and squared loss}.
\newblock In \emph{Proceedings of the Seventeenth National Conference on
  Artificial Intelligence and Twelfth Conference on Innovative Applications of
  Artificial Intelligence}, pages 564--569. AAAI Press.

\bibitem[{Dzikovska et~al.(2013)Dzikovska, Nielsen, Brew, Leacock, Giampiccolo,
  Bentivogli, Clark, Dagan, and Dang}]{dzikovska2013semeval}
Myroslava Dzikovska, Rodney Nielsen, Chris Brew, Claudia Leacock, Danilo
  Giampiccolo, Luisa Bentivogli, Peter Clark, Ido Dagan, and Hoa~Trang Dang.
  2013.
\newblock \href {https://aclanthology.org/S13-2045/} {{SemEval}-2013 {Task} 7:
  The joint student response analysis and 8th recognizing textual entailment
  challenge}.
\newblock In \emph{Second Joint Conference on Lexical and Computational
  Semantics (*SEM), Volume 2: Proceedings of the Seventh International Workshop
  on Semantic Evaluation (SemEval 2013)}, pages 263--274, Atlanta, Georgia,
  USA. Association for Computational Linguistics.

\bibitem[{Echterhoff et~al.(2024)Echterhoff, Liu, Alessa, McAuley, and
  He}]{echterhoff2024cognitive}
Jessica~Maria Echterhoff, Yao Liu, Abeer Alessa, Julian McAuley, and Zexue He.
  2024.
\newblock \href {https://doi.org/10.18653/v1/2024.findings-emnlp.739}
  {Cognitive bias in decision-making with {LLM}s}.
\newblock In \emph{Findings of the Association for Computational Linguistics:
  EMNLP 2024}, pages 12640--12653, Miami, Florida, USA. Association for
  Computational Linguistics.

\bibitem[{Fagerberg et~al.(2026)Fagerberg, Sallander, Patil, Berg, Nyman, Borg,
  and Lind{\'e}n}]{fagerberg2026batch}
Petter Fagerberg, Oscar Sallander, Kim~Vikhe Patil, Anders Berg, Anastasia
  Nyman, Natalia Borg, and Thomas Lind{\'e}n. 2026.
\newblock \href {https://doi.org/10.1002/cesm.70082} {Batch size effects on
  mid-2025 state-of-the-art large language model performance in automated title
  and abstract screening}.
\newblock \emph{Cochrane Evidence Synthesis and Methods}, 4(3):e70082.

\bibitem[{Frohmann et~al.(2026)Frohmann, Alavi, Lingg, and
  Rekabsaz}]{frohmann2026equal}
Markus Frohmann, Mahdiyar Ali~Akbar Alavi, Elizabeth Lingg, and Navid Rekabsaz.
  2026.
\newblock \href {https://arxiv.org/abs/2608.26762} {Equal ranking quality,
  different decisions: Measuring and reducing order dependence in {LLM}
  scorers}.
\newblock \emph{Preprint}, arXiv:2608.26762.

\bibitem[{{Gemma Team}(2026)}]{gemmateam2026gemma4}
{Gemma Team}. 2026.
\newblock \href {https://arxiv.org/abs/2607.02770} {{Gemma 4} technical
  report}.
\newblock \emph{Preprint}, arXiv:2607.02770.

\bibitem[{Gim et~al.(2024)Gim, Chen, Lee, Sarda, Khandelwal, and
  Zhong}]{gim2024promptcache}
In~Gim, Guojun Chen, Seung-seob Lee, Nikhil Sarda, Anurag Khandelwal, and Lin
  Zhong. 2024.
\newblock \href
  {https://proceedings.mlsys.org/paper_files/paper/2024/hash/a66caa1703fe34705a4368c3014c1966-Abstract-Conference.html}
  {{Prompt Cache}: Modular attention reuse for low-latency inference}.
\newblock In \emph{Proceedings of Machine Learning and Systems}, volume~6,
  pages 325--338.

\bibitem[{Glavas et~al.(2026)Glavas, Vedula, Dhyani, Valkanas, Zhu, and
  Malmasi}]{glavas2026ippd}
Theodore Glavas, Nikhita Vedula, Dushyanta Dhyani, Antonios Valkanas, Yilun
  Zhu, and Shervin Malmasi. 2026.
\newblock \href {https://arxiv.org/abs/2609.05707} {Intra-prompt parallel
  decoding for common-context question answering}.
\newblock \emph{Preprint}, arXiv:2609.05707.
\newblock Accepted to EMNLP 2026 main conference.

\bibitem[{Gomez et~al.(2024)Gomez, Machado, Paes, and
  Calmon}]{gomez2024arbitrariness}
Juan~Felipe Gomez, Caio Machado, Lucas~Monteiro Paes, and Flavio Calmon. 2024.
\newblock \href {https://doi.org/10.1145/3630106.3659036} {Algorithmic
  arbitrariness in content moderation}.
\newblock In \emph{Proceedings of the 2024 ACM Conference on Fairness,
  Accountability, and Transparency}, FAccT '24, pages 2234--2253, New York, NY,
  USA. Association for Computing Machinery.

\bibitem[{{Granite Team, IBM}(2025)}]{granite33modelcard}
{Granite Team, IBM}. 2025.
\newblock {Granite-3.3-8B-Instruct}.
\newblock \url{https://huggingface.co/ibm-granite/granite-3.3-8b-instruct}.
\newblock Accessed 2026-09-28.

\bibitem[{Hardy(1933)}]{hardy1933theorem}
G.~H. Hardy. 1933.
\newblock \href {https://doi.org/10.1112/jlms/s1-8.3.227} {A theorem concerning
  {Fourier} transforms}.
\newblock \emph{Journal of the London Mathematical Society}, s1-8(3):227--231.

\bibitem[{Hida et~al.(2026)Hida, Ribeiro, and Hida}]{hida2026class}
Gilberto~Sussumu Hida, Danilo~Monteiro Ribeiro, and Clayton~Suguio Hida. 2026.
\newblock \href {https://arxiv.org/abs/2608.14737} {Class imbalance and batch
  effects in {LLM}-based screening for systematic reviews}.
\newblock \emph{Preprint}, arXiv:2608.14737.
\newblock Accepted at ENIAC 2026 (National Meeting on Artificial and
  Computational Intelligence), part of BRACIS 2026.

\bibitem[{Hoeffding(1948)}]{hoeffding1948class}
Wassily Hoeffding. 1948.
\newblock \href {https://doi.org/10.1214/aoms/1177730196} {A class of
  statistics with asymptotically normal distribution}.
\newblock \emph{The Annals of Mathematical Statistics}, 19(3):293--325.

\bibitem[{Huang et~al.(2026)Huang, Madala, Niu, Hockenmaier, and
  Zhang}]{huang2026contextual}
Jerry Huang, Siddarth Madala, Cheng Niu, Julia Hockenmaier, and Tong Zhang.
  2026.
\newblock \href {https://doi.org/10.18653/v1/2026.acl-long.94} {Contextual
  relevance and adaptive sampling for {LLM}-based document reranking}.
\newblock In \emph{Proceedings of the 64th Annual Meeting of the Association
  for Computational Linguistics (Volume 1: Long Papers)}, pages 2076--2089, San
  Diego, California, United States. Association for Computational Linguistics.

\bibitem[{Hudgens and Halloran(2008)}]{hudgens2008toward}
Michael~G. Hudgens and M.~Elizabeth Halloran. 2008.
\newblock \href {https://doi.org/10.1198/016214508000000292} {Toward causal
  inference with interference}.
\newblock \emph{Journal of the American Statistical Association},
  103(482):832--842.

\bibitem[{Kapetanovic et~al.(2026)Kapetanovic, Altwlkany, Mercep, Duricic, and
  Lacic}]{kapetanovic2026anchoring}
Ante Kapetanovic, Kemal Altwlkany, Andro Mercep, Tomislav Duricic, and Emanuel
  Lacic. 2026.
\newblock \href {https://arxiv.org/abs/2608.25869} {Anchoring bias in
  {LLM}-as-a-judge systems: Prior scores compromise evaluation independence}.
\newblock \emph{Preprint}, arXiv:2608.25869.
\newblock To appear in the Proceedings of the 35th ACM International Conference
  on Information and Knowledge Management (CIKM '26).

\bibitem[{Kennedy et~al.(2020)Kennedy, Bacon, Sahn, and von
  Vacano}]{kennedy2020constructing}
Chris~J. Kennedy, Geoff Bacon, Alexander Sahn, and Claudia von Vacano. 2020.
\newblock \href {https://arxiv.org/abs/2009.10277v1} {Constructing interval
  variables via faceted {Rasch} measurement and multitask deep learning: A hate
  speech application}.
\newblock \emph{Preprint}, arXiv:2009.10277v1.

\bibitem[{Kim et~al.(2026)Kim, Kim, Lee, and Kim}]{kim2026safety}
Kihyun Kim, Hee-Seon Kim, Wonjun Lee, and Changick Kim. 2026.
\newblock \href {https://arxiv.org/abs/2608.02681} {Safety in batches?
  {U}nderstanding and mitigating safety failures in batch prompting}.
\newblock \emph{Preprint}, arXiv:2608.02681.

\bibitem[{Korikov et~al.(2025)Korikov, Du, Sanner, and
  Rekabsaz}]{korikov2025batched}
Anton Korikov, Pan Du, Scott Sanner, and Navid Rekabsaz. 2025.
\newblock \href {https://doi.org/10.18653/v1/2025.emnlp-main.1661} {Batched
  self-consistency improves {LLM} relevance assessment and ranking}.
\newblock In \emph{Proceedings of the 2025 Conference on Empirical Methods in
  Natural Language Processing}, pages 32687--32703, Suzhou, China. Association
  for Computational Linguistics.

\bibitem[{Kwon et~al.(2023)Kwon, Li, Zhuang, Sheng, Zheng, Yu, Gonzalez, Zhang,
  and Stoica}]{kwon2023vllm}
Woosuk Kwon, Zhuohan Li, Siyuan Zhuang, Ying Sheng, Lianmin Zheng, Cody~Hao Yu,
  Joseph Gonzalez, Hao Zhang, and Ion Stoica. 2023.
\newblock \href {https://doi.org/10.1145/3600006.3613165} {Efficient memory
  management for large language model serving with {PagedAttention}}.
\newblock In \emph{Proceedings of the 29th Symposium on Operating Systems
  Principles}, SOSP '23, pages 611--626, New York, NY, USA. Association for
  Computing Machinery.

\bibitem[{Larionov and Eger(2025)}]{larionov2025batchgemba}
Daniil Larionov and Steffen Eger. 2025.
\newblock \href {https://doi.org/10.1109/GACLM67198.2025.11232289}
  {{BatchGEMBA}: Token-efficient machine translation evaluation with batched
  prompting and prompt compression}.
\newblock In \emph{2025 2nd International Generative AI and Computational
  Language Modelling Conference (GACLM)}, pages 1--9, Valencia, Spain. IEEE.

\bibitem[{Lin et~al.(2024)Lin, Diesendruck, Du, and
  Abraham}]{lin2024batchprompt}
Jianzhe Lin, Maurice Diesendruck, Liang Du, and Robin Abraham. 2024.
\newblock \href
  {https://proceedings.iclr.cc/paper_files/paper/2024/hash/5d8c01de2dc698c54201c1c7d0b86974-Abstract-Conference.html}
  {{BatchPrompt}: Accomplish more with less}.
\newblock In \emph{The Twelfth International Conference on Learning
  Representations}, pages 21590--21612.

\bibitem[{Lov{\'a}sz(1979)}]{lovasz1979shannon}
L{\'a}szl{\'o} Lov{\'a}sz. 1979.
\newblock \href {https://doi.org/10.1109/TIT.1979.1055985} {On the {Shannon}
  capacity of a graph}.
\newblock \emph{IEEE Transactions on Information Theory}, 25(1):1--7.

\bibitem[{Lu et~al.(2023)Lu, Xu, Zhang, Min, Yang, and Lin}]{lu2023toxicn}
Junyu Lu, Bo~Xu, Xiaokun Zhang, Changrong Min, Liang Yang, and Hongfei Lin.
  2023.
\newblock \href {https://doi.org/10.18653/v1/2023.acl-long.898} {Facilitating
  fine-grained detection of {C}hinese toxic language: Hierarchical taxonomy,
  resources, and benchmarks}.
\newblock In \emph{Proceedings of the 61st Annual Meeting of the Association
  for Computational Linguistics (Volume 1: Long Papers)}, pages 16235--16250,
  Toronto, Canada. Association for Computational Linguistics.

\bibitem[{Marx et~al.(2020)Marx, Calmon, and Ustun}]{marx2020predictive}
Charles Marx, Flavio Calmon, and Berk Ustun. 2020.
\newblock \href {https://proceedings.mlr.press/v119/marx20a.html} {Predictive
  multiplicity in classification}.
\newblock In \emph{Proceedings of the 37th International Conference on Machine
  Learning}, volume 119 of \emph{Proceedings of Machine Learning Research},
  pages 6765--6774. PMLR.

\bibitem[{Matter et~al.(2024)Matter, Schirmer, Grinberg, and
  Pfeffer}]{matter2024incel}
Daniel Matter, Miriam Schirmer, Nir Grinberg, and J{\"u}rgen Pfeffer. 2024.
\newblock \href {https://doi.org/10.3389/frsps.2024.1383152} {Investigating the
  increase of violent speech in {Incel} communities with human-guided {GPT-4}
  prompt iteration}.
\newblock \emph{Frontiers in Social Psychology}, 2:1383152.

\bibitem[{McIlroy-Young et~al.(2024)McIlroy-Young, Brown, Olson, Zhang, and
  Dwork}]{mcilroyyoung2024order}
Reid McIlroy-Young, Katrina Brown, Conlan Olson, Linjun Zhang, and Cynthia
  Dwork. 2024.
\newblock \href {https://doi.org/10.52202/079017-2318} {Order-independence
  without fine tuning}.
\newblock In \emph{Advances in Neural Information Processing Systems},
  volume~37, pages 72818--72839. Curran Associates, Inc.

\bibitem[{Messing(2026)}]{messing2026hidden}
Solomon Messing. 2026.
\newblock \href {https://arxiv.org/abs/2604.11581} {Hidden measurement error in
  {LLM} pipelines distorts annotation, evaluation, and benchmarking}.
\newblock \emph{Preprint}, arXiv:2604.11581.

\bibitem[{Olsson(1979)}]{olsson1979polychoric}
Ulf Olsson. 1979.
\newblock \href {https://doi.org/10.1007/BF02296207} {Maximum likelihood
  estimation of the polychoric correlation coefficient}.
\newblock \emph{Psychometrika}, 44(4):443--460.

\bibitem[{{OpenAI}(2026)}]{openai2026caching}
{OpenAI}. 2026.
\newblock Prompt caching.
\newblock \url{https://developers.openai.com/api/docs/guides/prompt-caching}.
\newblock Accessed 2026-09-29.

\bibitem[{Parducci(1965)}]{parducci1965category}
Allen Parducci. 1965.
\newblock \href {https://doi.org/10.1037/h0022602} {Category judgment: A
  range-frequency model}.
\newblock \emph{Psychological Review}, 72(6):407--418.

\bibitem[{Pipal et~al.(2026)Pipal, Vogel, Wack, and Esser}]{pipal2026waste}
Christian Pipal, Eva-Maria Vogel, Morgan Wack, and Frank Esser. 2026.
\newblock \href {https://arxiv.org/abs/2604.03684} {Researchers waste 80\% of
  {LLM} annotation costs by classifying one text at a time}.
\newblock \emph{Preprint}, arXiv:2604.03684.

\bibitem[{Plackett(1954)}]{plackett1954reduction}
R.~L. Plackett. 1954.
\newblock \href {https://doi.org/10.1093/biomet/41.3-4.351} {A reduction
  formula for normal multivariate integrals}.
\newblock \emph{Biometrika}, 41(3--4):351--360.

\bibitem[{Ratner et~al.(2023)Ratner, Levine, Belinkov, Ram, Magar, Abend,
  Karpas, Shashua, Leyton-Brown, and Shoham}]{ratner2023pcw}
Nir Ratner, Yoav Levine, Yonatan Belinkov, Ori Ram, Inbal Magar, Omri Abend,
  Ehud Karpas, Amnon Shashua, Kevin Leyton-Brown, and Yoav Shoham. 2023.
\newblock \href {https://doi.org/10.18653/v1/2023.acl-long.352} {Parallel
  context windows for large language models}.
\newblock In \emph{Proceedings of the 61st Annual Meeting of the Association
  for Computational Linguistics (Volume 1: Long Papers)}, pages 6383--6402,
  Toronto, Canada. Association for Computational Linguistics.

\bibitem[{Sachdeva and Boudol(2026)}]{sachdeva2026rasch}
Pratik~S. Sachdeva and Nathan Boudol. 2026.
\newblock \href {https://arxiv.org/abs/2608.27463} {Rating the raters: {Rasch}
  measurement theory for {LLM} evaluation}.
\newblock \emph{Preprint}, arXiv:2608.27463.
\newblock Accepted to EMNLP 2026.

\bibitem[{Saltelli et~al.(2010)Saltelli, Annoni, Azzini, Campolongo, Ratto, and
  Tarantola}]{saltelli2010variance}
Andrea Saltelli, Paola Annoni, Ivano Azzini, Francesca Campolongo, Marco Ratto,
  and Stefano Tarantola. 2010.
\newblock \href {https://doi.org/10.1016/j.cpc.2009.09.018} {Variance based
  sensitivity analysis of model output. {Design} and estimator for the total
  sensitivity index}.
\newblock \emph{Computer Physics Communications}, 181(2):259--270.

\bibitem[{Shi et~al.(2025)Shi, Ma, Liang, Diao, Ma, and
  Vosoughi}]{shi2025judging}
Lin Shi, Chiyu Ma, Wenhua Liang, Xingjian Diao, Weicheng Ma, and Soroush
  Vosoughi. 2025.
\newblock \href {https://doi.org/10.18653/v1/2025.ijcnlp-long.18} {Judging the
  judges: A systematic study of position bias in {LLM}-as-a-judge}.
\newblock In \emph{Proceedings of the 14th International Joint Conference on
  Natural Language Processing and the 4th Conference of the Asia-Pacific
  Chapter of the Association for Computational Linguistics}, pages 292--314,
  Mumbai, India. The Asian Federation of Natural Language Processing and The
  Association for Computational Linguistics.

\bibitem[{Sobol'(2001)}]{sobol2001global}
I.~M. Sobol'. 2001.
\newblock \href {https://doi.org/10.1016/S0378-4754(00)00270-6} {Global
  sensitivity indices for nonlinear mathematical models and their {Monte Carlo}
  estimates}.
\newblock \emph{Mathematics and Computers in Simulation}, 55(1--3):271--280.

\bibitem[{{Team GLM} et~al.(2024){Team GLM}, Zeng, Xu, Wang, Zhang, Yin, Zhang,
  Rojas, Feng, Zhao, Lai, Yu, Wang, Sun, Zhang, Cheng, Gui, Tang, Zhang, Sun,
  Li, Zhao, Wu, Zhong, Liu, Huang, Zhang, Zheng, Lu, Duan, Zhang, Cao, Yang,
  Tam, Zhao, Liu, Xia, Zhang, Gu, Lv, Liu, Liu, Yang, Song, Zhang, An, Xu, Niu,
  Yang, Li, Bai, Dong, Qi, Wang, Yang, Du, Hou, and Wang}]{glm2024chatglm}
{Team GLM}, Aohan Zeng, Bin Xu, Bowen Wang, Chenhui Zhang, Da~Yin, Dan Zhang,
  Diego Rojas, Guanyu Feng, Hanlin Zhao, Hanyu Lai, Hao Yu, Hongning Wang,
  Jiadai Sun, Jiajie Zhang, Jiale Cheng, Jiayi Gui, Jie Tang, Jing Zhang, and
  39 others. 2024.
\newblock \href {https://arxiv.org/abs/2406.12793} {{ChatGLM}: A family of
  large language models from {GLM-130B} to {GLM-4 All Tools}}.
\newblock \emph{Preprint}, arXiv:2406.12793.

\bibitem[{Temkit(2026)}]{temkit2026amel}
Sid-Ali Temkit. 2026.
\newblock \href {https://arxiv.org/abs/2605.22714} {{AMEL}: Accumulated message
  effects on {LLM} judgments}.
\newblock \emph{Preprint}, arXiv:2605.22714.

\bibitem[{Wang et~al.(2024)Wang, Li, Chen, Cai, Zhu, Lin, Cao, Kong, Liu, Liu,
  and Sui}]{wang2024fair}
Peiyi Wang, Lei Li, Liang Chen, Zefan Cai, Dawei Zhu, Binghuai Lin, Yunbo Cao,
  Lingpeng Kong, Qi~Liu, Tianyu Liu, and Zhifang Sui. 2024.
\newblock \href {https://doi.org/10.18653/v1/2024.acl-long.511} {Large language
  models are not fair evaluators}.
\newblock In \emph{Proceedings of the 62nd Annual Meeting of the Association
  for Computational Linguistics (Volume 1: Long Papers)}, pages 9440--9450,
  Bangkok, Thailand. Association for Computational Linguistics.

\bibitem[{Yang et~al.(2025)Yang, Li, Yang, Zhang, Hui, Zheng, Yu, Gao, Huang,
  Lv, Zheng, Liu, Zhou, Huang, Hu, Ge, Wei, Lin, Tang, Yang, Tu, Zhang, Yang,
  Yang, Zhou, Zhou, Lin, Dang, Bao, Yang, Yu, Deng, Li, Xue, Li, Zhang, Wang,
  Zhu, Men, Gao, Liu, Luo, Li, Tang, Yin, Ren, Wang, Zhang, Ren, Fan, Su,
  Zhang, Zhang, Wan, Liu, Wang, Cui, Zhang, Zhou, and Qiu}]{yang2025qwen3}
An~Yang, Anfeng Li, Baosong Yang, Beichen Zhang, Binyuan Hui, Bo~Zheng, Bowen
  Yu, Chang Gao, Chengen Huang, Chenxu Lv, Chujie Zheng, Dayiheng Liu, Fan
  Zhou, Fei Huang, Feng Hu, Hao Ge, Haoran Wei, Huan Lin, Jialong Tang, and 41
  others. 2025.
\newblock \href {https://arxiv.org/abs/2505.09388} {{Qwen3} technical report}.
\newblock \emph{Preprint}, arXiv:2505.09388.

\bibitem[{Yuan et~al.(2024)Yuan, Feng, Li, Wang, Pan, Wang, Hu, and
  Li}]{yuan2024batcheval}
Peiwen Yuan, Shaoxiong Feng, Yiwei Li, Xinglin Wang, Boyuan Pan, Heda Wang, Yao
  Hu, and Kan Li. 2024.
\newblock \href {https://doi.org/10.18653/v1/2024.acl-long.846} {{BatchEval}:
  Towards human-like text evaluation}.
\newblock In \emph{Proceedings of the 62nd Annual Meeting of the Association
  for Computational Linguistics (Volume 1: Long Papers)}, pages 15940--15958,
  Bangkok, Thailand. Association for Computational Linguistics.

\bibitem[{Yue and Yao(2025)}]{yue2025batch}
Murong Yue and Ziyu Yao. 2025.
\newblock \href {https://doi.org/10.18653/v1/2025.findings-acl.245} {Efficient
  but vulnerable: Benchmarking and defending {LLM} batch prompting attack}.
\newblock In \emph{Findings of the Association for Computational Linguistics:
  ACL 2025}, pages 4746--4761, Vienna, Austria. Association for Computational
  Linguistics.

\bibitem[{Zhang et~al.(2015)Zhang, Zhao, and LeCun}]{zhang2015character}
Xiang Zhang, Junbo Zhao, and Yann LeCun. 2015.
\newblock \href
  {https://proceedings.neurips.cc/paper_files/paper/2015/hash/250cf8b51c773f3f8dc8b4be867a9a02-Abstract.html}
  {Character-level convolutional networks for text classification}.
\newblock In \emph{Advances in Neural Information Processing Systems},
  volume~28, pages 649--657. Curran Associates, Inc.

\bibitem[{Zhao et~al.(2021)Zhao, Wallace, Feng, Klein, and
  Singh}]{zhao2021calibrate}
Zihao Zhao, Eric Wallace, Shi Feng, Dan Klein, and Sameer Singh. 2021.
\newblock \href {https://proceedings.mlr.press/v139/zhao21c.html} {Calibrate
  before use: Improving few-shot performance of language models}.
\newblock In \emph{Proceedings of the 38th International Conference on Machine
  Learning}, volume 139 of \emph{Proceedings of Machine Learning Research},
  pages 12697--12706. PMLR.

\bibitem[{Zheng et~al.(2023)Zheng, Chiang, Sheng, Zhuang, Wu, Zhuang, Lin, Li,
  Li, Xing, Zhang, Gonzalez, and Stoica}]{zheng2023judging}
Lianmin Zheng, Wei-Lin Chiang, Ying Sheng, Siyuan Zhuang, Zhanghao Wu, Yonghao
  Zhuang, Zi~Lin, Zhuohan Li, Dacheng Li, Eric~P. Xing, Hao Zhang, Joseph~E.
  Gonzalez, and Ion Stoica. 2023.
\newblock \href {https://doi.org/10.52202/075280-2020} {Judging
  {LLM}-as-a-judge with {MT-Bench} and {Chatbot Arena}}.
\newblock In \emph{Advances in Neural Information Processing Systems},
  volume~36, pages 46595--46623. Curran Associates, Inc.

\bibitem[{Zheng et~al.(2024)Zheng, Yin, Xie, Sun, Huang, Yu, Cao, Kozyrakis,
  Stoica, Gonzalez, Barrett, and Sheng}]{zheng2024sglang}
Lianmin Zheng, Liangsheng Yin, Zhiqiang Xie, Chuyue Sun, Jeff Huang, Cody~Hao
  Yu, Shiyi Cao, Christos Kozyrakis, Ion Stoica, Joseph~E. Gonzalez, Clark
  Barrett, and Ying Sheng. 2024.
\newblock \href {https://doi.org/10.52202/079017-2000} {{SGLang}: Efficient
  execution of structured language model programs}.
\newblock In \emph{Advances in Neural Information Processing Systems},
  volume~37, pages 62557--62583. Curran Associates, Inc.

\end{thebibliography}

\appendix

\section{Prompts}
\label{app:prompts}

Every prompt has the form below, shown for grading; the opening line, the scale, the context line, the header and name of the items and the question differ by task.

\begin{quote}\small\ttfamily\frenchspacing
You grade students\textquotesingle{} short answers to science questions on this scale:\\
0 = Incorrect: the answer contradicts the reference answer, or is irrelevant or off-topic\\
1 = Partially correct: the answer has part of the reference answer\textquotesingle{}s content but is incomplete\\
2 = Correct: the answer conveys the content of the reference answer, in any wording\\[3pt]
Context: Each item gives a question, a reference answer and one student\textquotesingle{}s answer. Spelling and grammar do not matter. Each answer is graded on its own.\\[3pt]
Answers:\\
c0: Question: \dots{} Reference answer: \dots{} Student\textquotesingle{}s answer: \dots\\
c1: \dots\\[3pt]
Question: How correct is each student\textquotesingle{}s answer? Answer with a JSON object that maps every answer id to a single digit from 0 to 2, in the form \{"c0": <level>, "c1": <level>, ...\}, and nothing else.
\end{quote}

The harm prompt opens with ``You grade comments for harm on this scale:'', lists the comments under ``Comments:'' and asks ``How harmful is each comment?'' for digits from 0 to 4; its scale has five levels from \emph{safe} (ordinary chat, reactions, criticism that attacks no one) to \emph{dangerous} (credible threats, encouragement of self-harm). The star prompt opens with ``You predict how many stars a customer gave, from the text of the review, on this scale:'', lists the reviews under ``Reviews:'' and asks ``How many stars did each customer give?'' for digits from 1 to 5, one per star. The reworded variant changes the opening (for grading ``Rate each student's answer against the reference answer, using these levels:'', for harm ``Rate how harmful each comment is, using these levels:'', for stars ``Estimate the star rating that goes with each review, using these levels:'') and the request (for grading ``Which level fits each student's answer? Reply with a JSON object whose keys are the answer ids\dots'', and in the same way ``Which level fits each comment?'' and ``Which rating fits each review?''). The context lines of grading (above) and stars (``Reviews of local businesses written by customers. Each review is rated on its own.'') tell the model that each item is graded on its own; the harm context line (``Comments posted on social media (YouTube, Reddit and Twitter).'') only says where the comments come from.

\section{Data}
\label{app:data}

\emph{Harm}: Measuring Hate Speech \citep{kennedy2020constructing}, 2{,}000 comments sampled from the 15{,}046 comments of the run that included the population study (Appendix~\ref{app:population}): its 2{,}000 random comments and the targets and neighbours of a controlled design that we do not report, whose neighbours were drawn from all comments and from comments that every annotator rated as not hate speech, with a hate-speech score below $-1$, or as hate speech, with a score above 1. The 15{,}046 comments are 97\% of the 15{,}509 comments of the corpus with at least three annotators, 20--300 characters, no ``http'' and distinct texts. \emph{Grading}: the three SciEntsBank test sets (unseen answers, questions and domains) of SemEval-2013 Task 7 \citep{dzikovska2013semeval}; the five-way labels are mapped to three levels (\emph{correct}; \emph{partially correct, incomplete}; \emph{contradictory}, \emph{irrelevant} and \emph{non-domain} as incorrect); an item is the text ``Question: \dots{} Reference answer: \dots{} Student's answer: \dots'' of Appendix~\ref{app:prompts}, 15--420 characters with these labels. \emph{Stars}: the Yelp Review Full test set \citep{zhang2015character}, reviews of 15--400 characters. Of items with identical texts only the first is kept; this drops 338 items on grading, and the harm and star pools have no duplicates. For each task, 2{,}000 items are sampled with a fixed seed; the neighbours in the intervention study come from the same sample, so that each has a single-item verdict.

\section{Full results}
\label{app:full}

\begin{table*}[t]
\centering
\small
\setlength{\tabcolsep}{3pt}
\begin{tabular}{@{}ll rr rrr rrrrr@{}}
\toprule
& & \multicolumn{2}{c}{single item} & \multicolumn{3}{c}{batched baselines} & \multicolumn{5}{c}{batch conditions} \\
\cmidrule(lr){3-4}\cmidrule(lr){5-7}\cmidrule(l){8-12}
Model & Task & keys & reworded & keys & reworded & numeric & vs.\ alone & reord.\ others & reord.\ all & new neighb. & new part. \\
\midrule
Qwen3-4B & Harm & 3.0 & 4.0 & 3.5 & 4.6 & 0.9 & 13.7 & 9.3 & 12.3 & 12.4 & 13.1 \\
 & Grading & 1.0 & 4.6 & 1.6 & 4.7 & 0.8 & 15.7 & 8.0 & 12.0 & 12.2 & 13.5 \\
 & Stars & 0.6 & 1.7 & 1.0 & 2.2 & 0.2 & 5.7 & 2.5 & 3.3 & 3.4 & 3.9 \\
Qwen3-8B & Harm & 1.5 & 5.0 & 4.7 & 6.2 & 1.1 & 14.0 & 9.3 & 12.8 & 13.0 & 14.9 \\
 & Grading & 1.5 & 8.3 & 2.0 & 4.6 & 0.5 & 15.0 & 6.4 & 11.5 & 10.1 & 12.1 \\
 & Stars & 0.4 & 1.3 & 0.8 & 1.7 & 0.4 & 4.7 & 2.3 & 4.0 & 3.5 & 4.0 \\
Gemma-4 & Harm & 1.1 & 4.8 & 3.6 & 4.4 & 1.2 & 19.7 & 7.8 & 12.1 & 10.8 & 13.0 \\
 & Grading & 0.8 & 5.8 & 2.4 & 4.2 & 0.6 & 10.7 & 7.4 & 10.2 & 9.8 & 10.8 \\
 & Stars & 0.3 & 2.1 & 0.4 & 1.4 & 0.1 & 4.8 & 1.8 & 2.8 & 2.8 & 3.1 \\
Granite & Harm & 1.8 & 3.1 & 5.9 & 5.1 & 1.1 & 19.5 & 9.8 & 13.4 & 13.3 & 15.5 \\
 & Grading & 1.6 & 5.2 & 6.0 & 8.4 & 1.0 & 19.7 & 14.9 & 20.6 & 20.4 & 21.9 \\
 & Stars & 0.7 & 1.2 & 0.7 & 2.1 & 0.3 & 5.6 & 3.1 & 3.8 & 3.6 & 4.0 \\
GLM-4 & Harm & 1.7 & 5.3 & 4.8 & 6.5 & 2.1 & 16.1 & 12.5 & 17.5 & 15.1 & 17.4 \\
 & Grading & 1.1 & 5.7 & 4.4 & 8.0 & 1.5 & 16.7 & 10.6 & 13.9 & 13.9 & 16.1 \\
 & Stars & 0.3 & 1.3 & 0.4 & 1.3 & 0.2 & 4.9 & 2.3 & 2.7 & 2.7 & 3.2 \\
\bottomrule
\end{tabular}
\caption{Decisions changed (\%) in every condition. Single-item baselines are against the item alone; batched baselines, which are on partition 0, and batch conditions are against the base partition, except ``vs.\ alone''. Batch conditions are means over the three base partitions: \emph{reord.\ others} is ``keep slot, reorder the others'', measured on the items in the designated slots only and without the weights of \S\ref{sec:setup}, \emph{reord.\ all} is ``reorder the whole batch'', \emph{new neighb.}\ is ``keep slot, new neighbours'', and \emph{new part.}\ is the mean over the three pairs of partitions. \emph{Numeric}: the same prompts generated one at a time instead of in a GPU batch.}
\label{tab:full}
\end{table*}

\begin{table*}[t]
\centering
\small
\setlength{\tabcolsep}{4pt}
\begin{tabular}{@{}ll rrr rrr r@{}}
\toprule
& & \multicolumn{3}{c}{batched baselines} & \multicolumn{3}{c}{batch conditions} & \\
\cmidrule(lr){3-5}\cmidrule(lr){6-8}
Model & Task & keys & reworded & numeric & reord.\ all & new neighb. & new part. & ratio \\
\midrule
Qwen3-4B & Harm & 9.5 & 13.1 & 2.5 & 32.6 & 31.5 & 34.0 & 2.6 \\
 & Grading & 3.2 & 9.2 & 1.8 & 23.5 & 22.5 & 25.1 & 2.7 \\
 & Stars & 3.9 & 9.9 & 0.7 & 16.0 & 15.2 & 18.0 & 1.8 \\
Qwen3-8B & Harm & 10.3 & 16.3 & 2.3 & 30.8 & 28.7 & 34.4 & 2.1 \\
 & Grading & 4.8 & 9.1 & 1.4 & 20.5 & 19.7 & 22.3 & 2.4 \\
 & Stars & 3.7 & 8.1 & 2.0 & 20.9 & 17.3 & 20.9 & 2.6 \\
Gemma-4 & Harm & 6.5 & 10.1 & 2.2 & 24.9 & 22.5 & 27.5 & 2.7 \\
 & Grading & 5.2 & 8.0 & 1.4 & 17.9 & 16.8 & 19.2 & 2.4 \\
 & Stars & 1.8 & 5.6 & 0.4 & 15.2 & 13.3 & 16.3 & 2.9 \\
Granite & Harm & 15.6 & 16.9 & 3.5 & 37.8 & 32.4 & 39.2 & 2.3 \\
 & Grading & 9.3 & 12.7 & 1.9 & 29.4 & 28.6 & 30.1 & 2.4 \\
 & Stars & 5.0 & 12.1 & 1.4 & 18.4 & 17.4 & 20.8 & 1.7 \\
GLM-4 & Harm & 10.6 & 15.5 & 4.8 & 34.6 & 29.9 & 33.9 & 2.2 \\
 & Grading & 9.1 & 16.7 & 3.4 & 25.4 & 26.5 & 27.5 & 1.6 \\
 & Stars & 3.2 & 10.0 & 1.0 & 16.7 & 15.6 & 17.4 & 1.7 \\
\bottomrule
\end{tabular}
\caption{Written levels changed (\%, any change of the digit) against partition 0: the batched baselines, the reordered and new-neighbour batches derived from partition 0, and partition 1. \emph{Ratio}: new partition over reworded instruction.}
\label{tab:levels}
\end{table*}

\begin{table*}[t]
\centering
\small
\setlength{\tabcolsep}{5pt}
\begin{tabular}{@{}ll rrrr rrrr@{}}
\toprule
& & \multicolumn{4}{c}{all items} & \multicolumn{4}{c}{designated items} \\
\cmidrule(lr){3-6}\cmidrule(l){7-10}
Model & Task & ord. & slot & memb. & rem. & ord. & slot & memb. & rem. \\
\midrule
Qwen3-4B & Harm & 0.56 & 0.33 & 0.35 & $-$0.24 & 0.53 & 0.21 & 0.48 & $-$0.23 \\
 & Grading & 0.38 & 0.42 & 0.44 & $-$0.24 & 0.34 & 0.63 & 0.30 & $-$0.27 \\
Qwen3-8B & Harm & 0.44 & 0.31 & 0.34 & $-$0.09 & 0.51 & 0.26 & 0.29 & $-$0.06 \\
 & Grading & 0.28 & 0.63 & 0.41 & $-$0.32 & 0.21 & 0.85 & 0.30 & $-$0.35 \\
Gemma-4 & Harm & 0.35 & 0.48 & 0.33 & $-$0.17 & 0.33 & 0.59 & 0.33 & $-$0.26 \\
 & Grading & 0.51 & 0.38 & 0.32 & $-$0.20 & 0.59 & 0.46 & 0.16 & $-$0.22 \\
Granite & Harm & 0.41 & 0.35 & 0.33 & $-$0.09 & 0.48 & 0.26 & 0.20 & 0.06 \\
 & Grading & 0.51 & 0.38 & 0.37 & $-$0.26 & 0.46 & 0.48 & 0.30 & $-$0.24 \\
GLM-4 & Harm & 0.53 & 0.46 & 0.22 & $-$0.21 & 0.43 & 0.47 & 0.23 & $-$0.13 \\
 & Grading & 0.45 & 0.30 & 0.31 & $-$0.06 & 0.41 & 0.31 & 0.28 & 0.00 \\
\bottomrule
\end{tabular}
\caption{Shares of the new-partition dissimilarity as fixed before the runs: order $D_\mathrm{keep}/D_\mathrm{new}$, slot $(D_\mathrm{resh}-D_\mathrm{keep})/D_\mathrm{new}$, membership at a fixed slot $(D_\mathrm{comp}-D_\mathrm{keep})/D_\mathrm{new}$ and the remainder (rem.) ${(D_\mathrm{new}-D_\mathrm{resh}-D_\mathrm{comp}+D_\mathrm{keep})/D_\mathrm{new}}$, which the analysis fixed before the runs called the interaction, with $D_\mathrm{keep}$ on the designated items and the other dissimilarities on all items, and with all four dissimilarities on the designated items (400 per partition). If $D_\mathrm{keep}$ were the order component at a random slot, the shares on all items would estimate $V_s+V_{sN}$, $V_N+V_{sN}$ and $-V_{sN}$ (Proposition~\ref{prop:pf}). The designated slots give the first slot, which is less sensitive than the others, a fifth of the weight, so $D_\mathrm{keep}$ is smaller (\S\ref{sec:setup}): in the columns for all items the slot and membership shares are overstated and the remainder is understated. In the columns for the designated items all four dissimilarities give the first slot a fifth of the weight, and the direction of the error is not fixed.}
\label{tab:sharesKept}
\end{table*}

Table~\ref{tab:full} lists every condition for the five main models and all three tasks, and Table~\ref{tab:levels} the changes of the written level. Table~\ref{tab:sharesKept} gives the shares that we fixed before the runs and repeats them on the designated items only. There the membership share at a fixed slot is \KeptMembershipRange, and the remainder, which Proposition~\ref{prop:pf} requires to be negative or zero, is positive in one pair and zero in another; on the designated items the slots of the base partition are the five designated ones and the redrawn slots are not, so the proposition holds only approximately.

\section{Interventions}
\label{app:interventions}

\begin{table*}[t]
\centering
\small
\setlength{\tabcolsep}{3.2pt}
\begin{tabular}{@{}l rrrrr rrrrr@{}}
\toprule
& \multicolumn{5}{c}{Harm} & \multicolumn{5}{c}{Grading} \\
\cmidrule(lr){2-6}\cmidrule(l){7-11}
Edit & Q-4B & Q-8B & Gemma & Granite & GLM & Q-4B & Q-8B & Gemma & Granite & GLM \\
\midrule
Earlier verdicts from another batch & 11 & 15 & 14 & 16 & 23 & 11 & 13 & 7 & 18 & 21 \\
Neighbour texts from another batch & 13 & 17 & 17 & 17 & 20 & 15 & 13 & 9 & 26 & 26 \\
Both from another batch & 9 & 10 & 11 & 16 & 15 & 12 & 9 & 9 & 22 & 14 \\
Neighbour texts removed & 15 & 12 & 41 & 27 & 26 & 12 & 16 & 13 & 26 & 17 \\
Earlier verdicts from single-item grading & 6 & 5 & 8 & 12 & 10 & 6 & 8 & 2 & 8 & 8 \\
Own earlier verdicts, shuffled & 8 & 13 & 13 & 13 & 18 & 9 & 11 & 5 & 17 & 19 \\
Earlier verdicts one level lower & 25 & 38 & 13 & 8 & 21 & 29 & 34 & 18 & 54 & 40 \\
Earlier verdicts one level higher & 12 & 5 & 17 & 15 & 18 & 12 & 13 & 5 & 12 & 33 \\
Earlier verdicts all at the lowest level & 62 & 69 & 25 & 32 & 53 & 31 & 35 & 24 & 60 & 44 \\
Earlier verdicts all at the highest level & 28 & 19 & 54 & 50 & 40 & 38 & 45 & 14 & 20 & 53 \\
Texts removed, verdicts from another batch & 16 & 12 & 42 & 27 & 26 & 12 & 18 & 14 & 26 & 17 \\
Texts removed, verdicts at the lowest level & 13 & 10 & 36 & 24 & 16 & 14 & 19 & 13 & 26 & 15 \\
Other texts, verdicts at the lowest level & 60 & 67 & 24 & 34 & 51 & 31 & 34 & 25 & 61 & 42 \\
Target answered first & 10 & 6 & 14 & 18 & 18 & 10 & 11 & 10 & 21 & 22 \\
\bottomrule
\end{tabular}
\caption{Target decisions changed (\%) by each edit, relative to the unedited forward pass; 200 targets in two slots per task. Q-4B, Q-8B, Gemma and GLM are Qwen3-4B, Qwen3-8B, Gemma-4 and GLM-4. A decision is $P(\text{level} \ge \tau) \ge 0.5$ in the next-token distribution, renormalised over the levels (\S\ref{sec:setup}). \emph{Texts removed}: every neighbour's text is replaced by a dash and its key is kept. \emph{One level lower} leaves the verdicts at the lowest level unchanged, \emph{one level higher} those at the highest level. Counts in the text are made before rounding.}
\label{tab:interventions}
\end{table*}

Table~\ref{tab:interventions} lists all edits. Setting every earlier verdict to the lowest or the highest level produces contexts that the model would not write itself; the text does not rely on these edits.

\section{Estimator check on synthetic data}
\label{app:synthetic}

We simulated decisions for the grading design of the main experiment (the same items, partitions and derived batches) with latent scores $z=\mu+u_\mathrm{o}+u_\mathrm{s}+u_\mathrm{m}+u_\mathrm{sm}$ with variances 1, 0.05, 0.2, 0.1 and 0.1. All components are drawn separately for every item: the order component $u_\mathrm{o}$ is redrawn whenever the sequence of the item's neighbours changes, the slot component $u_\mathrm{s}$ has one value per slot of the item, the membership component $u_\mathrm{m}$ one per set of neighbours, and $u_\mathrm{sm}$ one per slot and set of neighbours. Scores are cut into three levels (30\%, 30\%, 40\%) and the decision is the top level, as for grading. The single-item score is $\mu$ plus an independent shift with variance 0.15, cut at the same value. With 25 slots, the components of Proposition~\ref{prop:pf} are shares \SynTrueOrder{} (order), \SynTrueSlot{} (slot), \SynTrueMembership{} (membership) and \SynTrueSlotMembership{} (interaction) of the dissimilarity between partitions, which is \SynTrueD.

The order component is estimated from the weighted designated slots, as for the data; since all slots are alike here, the weights change only the precision of the estimate. Over \SynRuns{} simulations, the estimators returned \SynOrder, \SynSlot, \SynMembership{} and \SynSlotMembership{} (mean and standard deviation over simulations), and the bootstrap intervals covered the true shares in \SynCoverOrder, \SynCoverSlot, \SynCoverMembership{} and \SynCoverSlotMembership{} of the simulations. The order share is slightly too high because the estimator of $D$ overstates it in the small tables of the designated slots (\S\ref{sec:setup}). We also simulated voting over 3, 5 and 9 permutations of the batches of two partitions. The votes of the two batches differed for \SynVoteThreeSim\%, \SynVoteFiveSim\% and \SynVoteNineSim\% of the items, against \SynVoteThreePred\%, \SynVoteFivePred\% and \SynVoteNinePred\% predicted from the estimated dissimilarities by Proposition~\ref{prop:vote} and \SynVoteThreeExact\%, \SynVoteFiveExact\% and \SynVoteNineExact\% from the true ones. With 25 slots the model satisfies Assumption~\ref{ass:components} only approximately: two permutations can put the item in the same slot, and the rest of its score given the item and the membership is then a mixture over its 25 slot values, not a normal variable. Simulated directly with \SynModelItems{} items, its votes differ for \SynVoteThreeModel\%, \SynVoteFiveModel\% and \SynVoteNineModel\% of the items, \SynVoteGap{} points above the formula at the true dissimilarities (Monte Carlo standard error at most \SynModelSE{} points); the simulations on the design lie within \SynVoteResidZ{} standard errors of their mean from these values (standard error up to \SynVoteSE{} points over the \SynRuns{} simulations). The true mean bias was \SynBiasTrue\%. The model gave \SynBiasModel\% and the bound of Equation~\eqref{eq:bound} \SynBiasBound\%, which is valid but not informative here. The majority of the nine batched decisions differed from the single-item decision for \SynBiasNine\% of the items: it overstates the bias when the context carries as much variance as in this simulation (\texttt{synthetic\_check.py}).

\section{Population study with seven models}
\label{app:population}

\begin{table}[t]
\centering
\footnotesize
\setlength{\tabcolsep}{2pt}
\begin{tabular}{@{}l rrr cc@{}}
\toprule
Model & vs.\ alone & keys & between & flagged & AUROC \\
\midrule
Qwen3-1.7B & 24.0 & 24.2 & 25.4 & 59\,$\to$\,55 & 0.77\,$\to$\,0.72 \\
Qwen3-4B & 12.9 & 3.2 & 13.1 & 57\,$\to$\,59 & 0.79\,$\to$\,0.79 \\
Qwen3-8B & 12.8 & 1.1 & 12.9 & 65\,$\to$\,64 & 0.80\,$\to$\,0.77 \\
Gemma-4 & 19.8 & 0.9 & 12.8 & 51\,$\to$\,32 & 0.81\,$\to$\,0.83 \\
Granite & 20.6 & 2.1 & 13.6 & 61\,$\to$\,43 & 0.83\,$\to$\,0.84 \\
GLM-4 & 17.0 & 2.0 & 18.3 & 50\,$\to$\,50 & 0.80\,$\to$\,0.79 \\
\bottomrule
\end{tabular}
\caption{Harm on 2{,}000 random comments from Measuring Hate Speech, $k=25$: decisions changed (\%) between batched and single-item grading, by renamed keys on single items, and between three random partitions; comments flagged (\%, alone $\to$ batched) and AUROC of the most likely level (alone $\to$ batched).}
\label{tab:population}
\end{table}

\begin{table}[t]
\centering
\footnotesize
\setlength{\tabcolsep}{2.5pt}
\begin{tabular}{@{}l ccccc@{}}
\toprule
Model & race & religion & origin & gender & sexuality \\
\midrule
Qwen3-4B & 62\,$\to$\,64 & 55\,$\to$\,58 & 61\,$\to$\,62 & 58\,$\to$\,59 & 53\,$\to$\,55 \\
Qwen3-8B & 69\,$\to$\,70 & 57\,$\to$\,60 & 67\,$\to$\,68 & 66\,$\to$\,63 & 58\,$\to$\,58 \\
Gemma-4 & 58\,$\to$\,35 & 44\,$\to$\,30 & 49\,$\to$\,28 & 50\,$\to$\,33 & 50\,$\to$\,36 \\
Granite & 66\,$\to$\,48 & 54\,$\to$\,41 & 60\,$\to$\,44 & 64\,$\to$\,43 & 54\,$\to$\,40 \\
GLM-4 & 49\,$\to$\,55 & 50\,$\to$\,49 & 48\,$\to$\,49 & 55\,$\to$\,51 & 47\,$\to$\,47 \\
\midrule
Comments & 654 & 367 & 341 & 609 & 404 \\
\bottomrule
\end{tabular}
\caption{Comments flagged (\%, level~2 or above), alone $\to$ batched, by the group that annotators say the comment targets; population study. The last row gives the number of comments per group (\PopGroupFewer); the differences between groups are point estimates.}
\label{tab:groups}
\end{table}

Before the main experiment we graded 2{,}000 random comments from the same corpus with the harm prompt and protocol of the main experiment and seven models (Table~\ref{tab:population}); \PopOverlap{} of them are also in the main experiment's harm sample. Qwen3-0.6B returned only \PopTinyParsed{} valid verdicts of the \PopTinyBatched{} it had to write in batches of 25 (2{,}000 comments in each of three partitions) and is omitted. The results agree with the main experiment and show a trend with size within Qwen3 up to 8B, whose 4B model is a later release (Instruct-2507) than the 1.7B and 8B models; on the main harm sample, a new partition changes about as many decisions for Qwen3-32B as for Qwen3-8B (Tables~\ref{tab:main} and~\ref{tab:scale}).

\section{Replication on Chinese moderation data}
\label{app:replication}

We first observed the effect in a moderation system for Chinese video comments that uses a closed, probability-returning model (Jev 1.13) with 16 comments per request. On the COLD \citep{deng2022cold} and ToxiCN \citep{lu2023toxicn} test sets, grading in batches of 16 instead of one comment per request changed 9.2\% and 10.7\% of decisions at level~2, against 1.7\% and 1.5\% between two runs of the same batches, while AUROC fell by 0.018 and 0.004. Regrouping the comments by topic instead of file order, at the same batch size, changed 6.7\% and 8.7\% of decisions, more than rewording the rubric (5.8\% and 7.1\%). Recalibrating the batched threshold to reproduce the single-comment flag rate left the changed decisions as they were (COLD 6.5\% before and 6.6\% after, ToxiCN 8.9\% and 8.6\%; here the decision is on the expected level, so the rates differ from those above), which shows that batching reorders comments rather than shifting them all. A controlled design with fixed targets and chosen neighbours, run on the closed model and the five open models of the main experiment, showed the same effect for all of them except GLM-4, whose verdicts rarely reached the decision threshold although its probabilities moved. It also showed that removing the phrase ``read in context'' from the question, quoting the target inside the question, or checking that the model can copy the target verbatim does not remove the effect. In the same design (144 comments that are borderline for the closed model, batches of 16 with benign or harmful neighbours), averaging the verdict distributions of six batches, three slots with two draws of neighbours each, still left 10--24\% of decisions different from single-comment grading for the closed model, Qwen3-8B, Qwen3-4B and Gemma-4.

\section{Proofs and further results}
\label{app:proofs}

Every lemma and proposition of Section~\ref{sec:setup} and of this appendix is also checked by computation: by exact enumeration of small designs, by numerical integration, in part to 30 digits, or by simulation (\texttt{verify\_framework.py}).

\subsection{Setting}

A context of item $i$ is $c=(s,N,\omega)$ with $s\in\{0,\dots,k-1\}$, $N$ a set of $k-1$ other items and $\omega$ a permutation of $\{1,\dots,k-1\}$: the $t$-th neighbour shown in the prompt is the one with the $\omega(t)$-th smallest index in $N$. Let $\mathcal{D}_i$ be the distribution under which $s$, $N$ and $\omega$ are independent and uniform. We assume that $k\ge2$ divides $n$, and we write $\Phi$ and $\varphi$ for the distribution function and the density of the standard normal distribution.

\begin{lemma}
\label{lem:design}
If the items are arranged in a uniformly random sequence that is cut into consecutive blocks of $k$, the context of every item $i$ has distribution $\mathcal{D}_i$.
\end{lemma}

\begin{proof}
The position of $i$ is uniform on the $n$ positions and every slot corresponds to $n/k$ positions, so $s$ is uniform. Given the position of $i$, the other items are arranged uniformly at random over the other positions. Hence the other positions of the block of $i$ hold a uniformly random sequence of $k-1$ distinct items, whose law does not depend on the position of $i$; so the sequence is independent of $s$. A sequence of $k-1$ distinct items is a pair $(N,\omega)$, and every set $N$ has $(k-1)!$ sequences, so $N$ and $\omega$ are independent and uniform.
\end{proof}

\begin{proof}[Proof of Proposition~\ref{prop:flip}]
The decisions of item $i$ in two independent contexts are independent Bernoulli variables with parameter $p_i$, so they differ with probability $2p_i(1-p_i)$. By Lemma~\ref{lem:design} the two contexts that independent partitions give to item $i$ have distribution $\mathcal{D}_i$ and are independent, and the expected share follows by linearity. $F_\mathrm{new}=0$ if and only if $p_i\in\{0,1\}$ for every $i$; since every context has positive probability, this means that every $d_i$ is constant, that is, the decisions are separable.
\end{proof}

\begin{proof}[Proof of Proposition~\ref{prop:pf}]
Let $C_{-R}$ be the factors that are kept and $g=\E[f(C)\mid C_{-R}]$. Given $C_{-R}$, $f(C)$ and $f(C^{(R)})$ are independent with the same conditional law, because the factors are independent and those in $R$ are replaced by independent draws from the same distributions. Hence $\E[f(C)f(C^{(R)})]=\E g^2$, and $G(R)=\E f^2-\E g^2=\Var f-\Var g=\E\Var(f\mid C_{-R})$ by the law of total variance. If $R\subseteq R'$, then $\E[f\mid C_{-R'}]=\E[g\mid C_{-R'}]$, whose variance is at most $\Var g$; so $G(R)\le G(R')$.

Let $\eta=\E[f\mid s,N]$, $\eta_s=\E[f\mid s]-\E f$, $\eta_N=\E[f\mid N]-\E f$ and $\eta_{sN}=\eta-\E f-\eta_s-\eta_N$. By the tower property $\E[\eta\mid s]=\E[f\mid s]$, and $\E[\eta_N\mid s]=0$ because $s$ and $N$ are independent; so $\E[\eta_{sN}\mid s]=0$ and, in the same way, $\E[\eta_{sN}\mid N]=0$. Therefore $\eta_s$, $\eta_N$ and $\eta_{sN}$ are uncorrelated, $\Var\eta=V_s+V_N+\Var\eta_{sN}$, and $V_{sN}=\Var\eta_{sN}\ge0$. For the four sets, the kept factors are $(s,N)$, $N$, $s$ and none, so $G$ equals $\Var f-\Var\eta$, $\Var f-V_N$, $\Var f-V_s$ and $\Var f$, where $\Var f=T_\omega+\Var\eta$. For values in $\{0,1\}$, $(f(C)-f(C^{(R)}))^2$ is the indicator of a flip.
\end{proof}

All sets $R$ that we use contain $\omega$. For them $G(R)$ depends on $f$ through conditional expectations given $s$ and $N$ only, so it does not depend on how the order is encoded as a permutation.

\subsection{The conditions of the experiment}

The derived conditions are built from a base partition as follows. The \emph{reordered batch} puts the members of every batch in a new uniformly random order. For \emph{keep slot, new neighbours}, the batch with index $b$ takes, for every slot $j$, the item that the base partition has in slot $j$ of the batch with index $b-j$ modulo $n/k$. For \emph{keep slot, reorder the others}, the items in the designated slots $K=\{0,6,12,18,24\}$ keep their slots and the others are put in a new uniformly random order among the other slots. For a condition let $\hat F$ be the observed share of items whose decision differs from the base partition, measured on all items, on the items in a designated slot $s$ for the last condition ($\hat F_{\text{keep},s}$), or on the items that keep their slot in the reordered batch ($\hat F_\mathrm{chance}$). Items without a valid verdict in one of the two conditions are left out, and so are items without a valid single-item verdict; in every condition they are at most \MissingMax{} of the 2{,}000 items of a task. The patterns of decisions, the comparisons with single-item grading, the error rates against the labels in \S\ref{sec:accuracy} and the thresholds (Tables~\ref{tab:modelcheck}, \ref{tab:bias} and~\ref{tab:thresholds} and the numbers taken from them) use only the items with a valid verdict in the three single-item conditions and in the nine batched conditions of the three partitions (the partitions, their reordered batches and their batches with new neighbours), at least \ValidMin{} per model and task, so their rates can differ slightly from those of Table~\ref{tab:full}; Table~\ref{tab:accuracy} uses the items with a valid verdict alone and in the three partitions. Numerical noise of batched inference acts like a further factor that every condition redraws, so the statements below hold with $\omega$ standing for the order and this noise together. Bars denote means over items, for $f=d_i$, of the quantities of Proposition~\ref{prop:pf} and of the values of $G$ of the conditions.

\begin{lemma}
\label{lem:kneser}
Let $n\ge2k$, let $N$ be uniform on the memberships of item $i$ and, given $N$, let $N'$ be uniform on the memberships that are disjoint from $N$. For every real function $\beta$ of the membership,
the covariance $\Cov(\beta(N),\beta(N'))$ lies between $-\varepsilon_1\Var\beta(N)$ and $\varepsilon_2\Var\beta(N)$, where
\begin{equation*}
\varepsilon_1=\frac{k-1}{n-k},\qquad\varepsilon_2=\frac{(k-1)(k-2)}{(n-k)(n-k-1)}.
\end{equation*}
\end{lemma}

\begin{proof}
$N'$ is uniform as well, because every membership is disjoint from the same number $\binom{n-k}{k-1}$ of memberships. The pair $(N,N')$ is a step of the random walk on the Kneser graph of the $(k-1)$-subsets of $n-1$ elements, in which two subsets are adjacent if they are disjoint. Its adjacency matrix is symmetric with eigenvalues $(-1)^t\binom{n-k-t}{k-1-t}$, $t=0,\dots,k-1$, and the eigenvectors of $t=0$ are the constant functions \citep[proof of Theorem~13]{lovasz1979shannon}. Dividing by the degree gives the eigenvalues $\lambda_t$ of the transition matrix $P$. Since $|\lambda_{t+1}/\lambda_t|={(k-1-t)/(n-k-t)<1}$, the smallest $\lambda_t$ with $t\ge1$ is $\lambda_1=-\varepsilon_1$ and the largest is $\lambda_2=\varepsilon_2$ (for $k=2$ there is only $\lambda_1$, and $\varepsilon_2=0$). With $\beta_0=\beta-\E\beta(N)$, which is orthogonal to the constants, $\Cov(\beta(N),\beta(N'))=\E[\beta_0(N)(P\beta_0)(N)]$ lies between $\lambda_1\E\beta_0^2$ and $\lambda_2\E\beta_0^2$.
\end{proof}

\begin{proposition}
\label{prop:conditions}
Let $n/k\ge k$ and let every item have a valid verdict in the conditions that are compared. Over the randomness of the design,
\begin{enumerate}\itemsep 0pt
\item[(i)] $\E\hat F_\mathrm{new}=2\overline{G}(\{s,N,\omega\})$ and $\E\hat F_\mathrm{resh}=2\overline{G}(\{s,\omega\})$, so that $\E[(\hat F_\mathrm{new}-\hat F_\mathrm{resh})/2]=\overline{V}_N$;
\item[(ii)] $\E\hat F_\mathrm{comp}=2\overline{G}_\mathrm{comp}$, where $\overline{G}_\mathrm{comp}-\overline{G}(\{N,\omega\})$ lies between $-\varepsilon_2(\overline{V}_N+\overline{V}_{sN})$ and $\varepsilon_1(\overline{V}_N+\overline{V}_{sN})$;
\item[(iii)] for every $s\in K$, $\E\hat F_{\text{keep},s}$ is twice the mean over items of $\E[\Var(d_i\mid s,N,\omega_K)\mid s]$, where $\omega_K$ records which neighbours are in the other designated slots; this is at most twice the mean of $\E[\Var(d_i\mid s,N)\mid s]$, the order component at slot $s$;
\item[(iv)] an item keeps its slot in the reordered batch with probability $1/k$, whatever its context, and its decision then differs from the base with probability $2T_\omega$ for $f=d_i$; so the numerator of $\hat F_\mathrm{chance}$, the number of flips among these items, has expectation $2\overline{T}_\omega\,n/k$, and its denominator, their number, has expectation $n/k$.
\end{enumerate}
\end{proposition}

\begin{proof}
The expectation of a share is the mean over items of the probability that the decision of the item differs, by linearity; it does not matter that the items of a batch share their neighbours.

(i) The context of $i$ in a new partition is independent of its context in the base, and reordering a batch uniformly keeps $N$ and redraws the sequence of the batch independently. The two pairs of contexts are therefore $(C,C^{(R)})$ with $R=\{s,N,\omega\}$ and $R=\{s,\omega\}$.

(ii) The item keeps its slot $s$. Its new neighbours are the items that the base partition has in the slots $j\ne s$ of $k-1$ other batches, which are distinct batches because $n/k\ge k$. Given the base context of $i$, the base partition arranges the items outside the batch of $i$ uniformly at random, so the new neighbours form a uniformly random sequence of distinct items outside the old batch: $N'$ is uniform on the memberships disjoint from $N$, and the new order $\omega'$ is uniform and independent of $(s,N,N',\omega)$. With $\eta=\E[d_i\mid s,N]$, this gives $\E[d_i(C)d_i(C')]=\E[\eta(s,N)\eta(s,N')]$. For fixed $s$ let $\beta_s(N)=\eta(s,N)-\E[d_i\mid s]$. Since $N$ and $N'$ are both uniform, $\Cov(d_i(C),d_i(C'))=V_s+\E_s\Cov(\beta_s(N),\beta_s(N'))$, and therefore $G_\mathrm{comp}=G(\{N,\omega\})-\E_s\Cov(\beta_s(N),\beta_s(N'))$. Lemma~\ref{lem:kneser} and $\E_s\Var\beta_s(N)=\Var\eta-V_s=V_N+V_{sN}$ give the bounds.

(iii) An item in a designated slot keeps its slot and its membership, and the neighbours in the other designated slots stay in place; the other neighbours are arranged uniformly at random over the other slots, independently of their arrangement in the base. Under $\mathcal{D}_i$ given $s$, the arrangement of the neighbours consists of $\omega_K$ and of the arrangement of the remaining neighbours, which is uniform given $\omega_K$. So the pair of contexts keeps $(s,N,\omega_K)$ and redraws the rest, and as in Proposition~\ref{prop:pf} the probability of a flip is $2\,\E[\Var(d_i\mid s,N,\omega_K)\mid s]$. Every item is in slot $s$ with probability $1/k$, and the share is measured on the $n/k$ items in slot $s$. Conditioning on less cannot decrease the expected conditional variance.

(iv) The reordered batch is a uniformly random order of the members of the batch, drawn independently of the base partition. Item $i$ therefore keeps its slot with probability $1/k$, whatever its context, and given this the other $k-1$ members are arranged uniformly at random over the other slots, independently of their arrangement in the base. The pair of contexts keeps $(s,N)$ and redraws $\omega$, so the probability of a flip is $2\,\E\Var(d_i\mid s,N)$, and summing over items gives the last claim.
\end{proof}

For $n=2000$ and $k=25$, $\varepsilon_1=24/1975\approx0.0122$ and $\varepsilon_2\approx0.00014$. An item that is left out of a comparison changes its flip rate by at most its weight over the total weight of the items of the comparison, left-out items included, whatever its decisions would have been; on harm and grading at most \LeftOutMax{} items are left out of any comparison of the decomposition, which can move a flip rate by at most \LeftOutShift{} points. By (iii), a weighted mean of the $\hat F_{\text{keep},s}/2$ is in expectation at most the same weighted mean of the order components at the designated slots; with the weights of Section~\ref{sec:setup}, $1/25$ for slots 0 and 24 and $23/75$ for slots 6, 12 and 18, this is $\overline{T}_\omega$ if the mean order component of slots 6, 12 and 18 equals that of all interior slots. (iv) needs no such condition, but it uses only one item in $k$. The argument of (iii) applies unchanged to an item outside the designated slots that lands in its own slot, which happens with probability $1/(k-|K|)$ whatever its context: its flips give a lower bound of the order component at its slot as well. The proposition holds for every function of the context, in particular for the latent scores of the next subsection.

\subsection{Latent scale}

Assumption~\ref{ass:latent} describes a large population of items and contexts. With finitely many of them it can hold only approximately: the scores take finitely many values, and a redrawn factor equals the old one with positive probability ($1/k$ for the slot). The assumption also forces the effects of the context to be normal given the item, with the same variance for all items. Let $Z$ and $Z'$ be the scores of a random item in its contexts in two independent partitions, and $\delta$ and $\delta'$ their deviations from the mean of the item, which are independent with the same law given the item (Lemma~\ref{lem:design}). By the assumption $Z-Z'=\delta-\delta'$ is $\mathcal{N}(0,2D_\mathrm{new})$, and its law is a mixture over items of the laws of $\delta-\delta'$, whose characteristic functions are non-negative. For any set of items, the part of the mixture that comes from these items therefore has a density at most that of $\mathcal{N}(0,2D_\mathrm{new})$ and a non-negative characteristic function at most that of $\mathcal{N}(0,2D_\mathrm{new})$. By Hardy's theorem \citep{hardy1933theorem} this part is a multiple of $\mathcal{N}(0,2D_\mathrm{new})$, so $\delta-\delta'$ is $\mathcal{N}(0,2D_\mathrm{new})$ given almost every item, and by Cram\'er's theorem \citep{cramer1936eigenschaft} $\delta$ is then $\mathcal{N}(0,D_\mathrm{new})$ given the item. We therefore treat it as a model and compare its predictions with patterns of decisions to which it was not fitted (Appendix~\ref{app:checks}).

In each condition the two contexts of an item have the same distribution, so the two scores have the same law and can be standardised together. Then $D=1-\rho=\frac12\E(z_i(C)-z_i(C'))^2$, which needs no normality, and the statements of Propositions~\ref{prop:pf} and~\ref{prop:conditions} hold for $D$ with the components of $z$, in units of the variance of $z$ over items and contexts.

\begin{proof}[Proof of Proposition~\ref{prop:law}]
Let $(Z,Z')$ be bivariate normal with standard margins, correlation $\rho\in(-1,1)$ and density $\varphi_2(x,y;\rho)$, and let $\Lambda(h,h';\rho)=\Prob(Z>h,Z'>h')$. The density satisfies $\partial\varphi_2/\partial\rho=\partial^2\varphi_2/\partial x\,\partial y$ \citep{plackett1954reduction}, hence $\partial\Lambda/\partial\rho=\varphi_2(h,h';\rho)$, and $\varphi_2(h,h;r)=e^{-h^2/(1+r)}/(2\pi\sqrt{1-r^2})$. Writing $Z'=\rho Z+\sqrt{1-\rho^2}\,W$ with $W$ independent of $Z$ shows that $\Lambda(h,h;\rho)\to{\Prob(Z>h)}$ as $\rho\to1$, and $\varphi_2(h,h;r)$ is integrable on $(\rho,1)$. So $\Psi={2(\Prob(Z>h)-\Lambda(h,h;\rho))}=2\int_\rho^1\varphi_2(h,h;r)\,dr$, which is Equation~\eqref{eq:law}. On $[\rho,1]$ the function $r\mapsto e^{-h^2/(1+r)}$ does not decrease, and $\int_\rho^1(1-r^2)^{-1/2}dr=\arccos\rho$; this gives the two bounds. Let $\vartheta={\arccos(1-D)}$, so that $\sqrt{2D}=2\sin(\vartheta/2)$. Then ${\sqrt{2D}\le\vartheta}$, and $\vartheta=2\arcsin\sqrt{D/2}=\sqrt{2D}\,(1+D/12+O(D^2))$. Together with $e^{-h^2/(1+\rho)}=e^{-h^2/2}e^{-h^2D/(2(2-D))}$, the bounds give the last claim. We set ${\Psi(h,1)=0}$, the limit of $\Psi(h,\rho)$ as $\rho\to1$.
\end{proof}

\subsection{Voting}

Independent memberships of an item have independent effects, and independent arrangements have independent rests given the item and their memberships, because the contexts are independent. Under Assumption~\ref{ass:components}, the scores of an item in two contexts are bivariate normal with correlation $1-D_\mathrm{new}$ if the contexts are independent and $1-D_\mathrm{resh}$ if they share their membership and their arrangements are independent, as for a batch and its reordering (Assumption~\ref{ass:latent}); the assumption does not specify pairs that also share the slot, as in the two ``keep slot'' conditions. Like Assumption~\ref{ass:latent}, it can hold only approximately with finitely many slots. It follows from Assumption~\ref{ass:latent} for the new partition and the reordered batch if $D_\mathrm{resh}>0$ and $D_\mathrm{new}<1$ ($D_\mathrm{resh}\le D_\mathrm{new}$ holds by Proposition~\ref{prop:pf}). The argument of the previous subsection, with the item and its membership in place of the item, makes $e$ normal with variance $D_\mathrm{resh}$ given the item and the membership, and so independent of the membership given the item. Since $a+e$ is normal with variance $D_\mathrm{new}$ given the item, Cram\'er's theorem makes $a$ normal with variance $D_\mathrm{new}-D_\mathrm{resh}=V_N$ given the item. Finally $\mu$ is normal with variance $1-D_\mathrm{new}$, because $a+e$ has the same law given every item and $\mu+a+e$ is standard normal. When an estimate of $V_N$ is negative, which happens only in bootstrap resamples, we set it to 0 by setting $D_\mathrm{resh}$ to $D_\mathrm{new}$.

\begin{proof}[Proof of Proposition~\ref{prop:vote}]
(i) Let $\bar f_m$ be the mean of $f$ over $m$ independent uniform arrangements $(s_j,\omega_j)$ of a batch with membership $N$. Given $N$, the $m$ terms are independent with mean $\E[f\mid N]$ and variance $\Var(f\mid N)$. The law of total variance gives $\Var\bar f_m=V_N+\E\Var(f\mid N)/m$, and $\E\Var(f\mid N)=G(\{s,\omega\})$ by Proposition~\ref{prop:pf}. The mean converges to $\E[f\mid N]$ by the law of large numbers. For $f=d_i$ this limit is $p_i(N)=\Prob(d_i=1\mid N)$, and the majority converges to $\mathbf{1}[p_i(N)>\frac12]$ if $p_i(N)\ne\frac12$. If $V_N=0$, $p_i(N)=p_i$ for every $N$; if moreover $p_i\ne\frac12$, the majorities of any two batches that contain item $i$ have the same limit.

(ii) Let $g_m(x)={\Prob(\mathrm{Bin}(m,x)>m/2)}$ for odd $m$. Given $(\mu,a)$, the decisions in the $m$ arrangements are independent Bernoulli variables with parameter $\Phi((\mu+a-h)/\sqrt{D_\mathrm{resh}})$. Since the memberships of the two batches are independent, so are $a$ and $a'$ and the rests of their arrangements; given $\mu$, the votes of the two batches are therefore independent with
\begin{equation*}
\Gamma_m(\mu)=\E\Big[g_m\Big(\Phi\Big(\frac{\mu+a-h}{\sqrt{D_\mathrm{resh}}}\Big)\Big)\,\Big|\,\mu\Big],
\end{equation*}
so $F_m=\E[2\Gamma_m(\mu)(1-\Gamma_m(\mu))]$. For $m=1$ the two scores are standard normal with correlation $1-D_\mathrm{new}$. For odd $m$, $g_m(1-x)=1-g_m(x)$, and $g_{m+2}(x)-g_m(x)=\binom{m}{(m-1)/2}[x(1-x)]^{(m+1)/2}(2x-1)$, because two further decisions change the majority only if the first $m$ are split $(m+1)/2$ to $(m-1)/2$ and both go against it. Let $u(y)$ be $g_m(\Phi(y/\sqrt{D_\mathrm{resh}}))-\frac12$ or $g_{m+2}(\Phi(y/\sqrt{D_\mathrm{resh}}))-g_m(\Phi(y/\sqrt{D_\mathrm{resh}}))$; in both cases $u$ is odd and positive for $y>0$. Given $\mu$, $y=\mu+a-h$ is normal with mean $\mu-h$, and for $\mu>h$ its density $q$ satisfies $q(y)>q(-y)$ for $y>0$ (if $V_N=0$, then $y=\mu-h>0$), so $\E[u(y)\mid\mu]=\int_0^\infty u(y)(q(y)-q(-y))\,dy>0$. Hence $\frac12<\Gamma_m(\mu)<\Gamma_{m+2}(\mu)$ for $\mu>h$, and in the same way $\Gamma_{m+2}(\mu)<\Gamma_m(\mu)<\frac12$ for $\mu<h$. Since $2\Gamma(1-\Gamma)=\frac12-2(\Gamma-\frac12)^2$, $F_m$ decreases strictly with $m$. The argument uses the normality of $a$: with a two-point membership effect, $F_m$ can lie below its limit. As $m\to\infty$, $g_m(x)\to\mathbf{1}[x>\frac12]$ for $x\ne\frac12$, and the event $\mu+a=h$ has probability 0 because $\mu+a$ has variance $1-D_\mathrm{resh}>0$; by dominated convergence $F_m$ tends to the probability that $\mathbf{1}[\mu+a>h]$ and $\mathbf{1}[\mu+a'>h]$ differ. If $V_N=0$ this is 0. Otherwise $\mu+a$ and $\mu+a'$ are normal with variance $1-D_\mathrm{resh}$ and covariance $1-D_\mathrm{new}$, which gives Equation~\eqref{eq:floor} after division by $\sqrt{1-D_\mathrm{resh}}$; the correlation is below 1, so $F_\infty>0$. Since $F_m$ decreases strictly to $F_\infty$, $F_m>F_\infty$ for every odd $m$.
\end{proof}

For odd $m\ge3$ the majority also leaves more flips than the mean score of the same $m$ arrangements would (\S\ref{sec:voting}; with $D_\mathrm{resh}=D_\mathrm{new}$ these are $m$ independent contexts, as in the remark after Proposition~\ref{prop:law}). For independent $X_1,\dots,X_m\sim\mathcal{N}(\xi,1)$ the tests $\mathbf{1}[\sum_jX_j>0]$ and ${\mathbf{1}[\#\{j:X_j>0\}>m/2]}$ both have level $\frac12$ at $\xi=0$, and by the Neyman--Pearson lemma the first is the most powerful test of this level against every $\xi>0$. For $m\ge3$ the two tests differ with positive probability, so by the uniqueness part of the lemma the majority test is strictly less powerful: $g_m(\Phi(\xi))<\Phi(\sqrt m\,\xi)$ for $\xi>0$. Given $\mu$, the mean score of a batch exceeds $h$ with probability $\bar\Gamma_m(\mu)=\E[\Phi(\sqrt m\,(\mu+a-h)/\sqrt{D_\mathrm{resh}})\mid\mu]$. Since $u(y)=\Phi(\sqrt m\,y/\sqrt{D_\mathrm{resh}})-g_m(\Phi(y/\sqrt{D_\mathrm{resh}}))$ is odd and positive for $y>0$, the argument of the proof above gives $\frac12<\Gamma_m(\mu)<\bar\Gamma_m(\mu)$ for $\mu>h$ and the reverse for $\mu<h$, so that $F_m>\E[2\bar\Gamma_m(\mu)(1-\bar\Gamma_m(\mu))]$, the probability that the mean scores of the two batches give different decisions.

\begin{proposition}
\label{prop:patterns}
Under Assumption~\ref{ass:components} let $p(\mu)=\Phi((\mu-h)/\sqrt{D_\mathrm{new}})$, $p_N(u)=\Phi((u-h)/\sqrt{D_\mathrm{resh}})$ and, for odd $M$, $g_M(x)=\Prob(\mathrm{Bin}(M,x)>M/2)$.
\begin{enumerate}\itemsep 0pt
\item[(i)] The decisions of an item in $M$ independent contexts are not all equal with probability $1-\E[p(\mu)^M+(1-p(\mu))^M]$.
\item[(ii)] The majorities over two disjoint sets of $M$ independent contexts each ($M$ odd) differ with probability $\E[2g_M(p(\mu))(1-g_M(p(\mu)))]$.
\item[(iii)] Let two batches with independent memberships each be graded in two independent arrangements. The two arrangements of each batch agree and the two batches differ with probability $\E[2R_1(\mu)R_0(\mu)]$, where $R_1(\mu)=\E[p_N(\mu+a)^2\mid\mu]$ and $R_0(\mu)=\E[(1-p_N(\mu+a))^2\mid\mu]$.
\end{enumerate}
\end{proposition}

\begin{proof}
For independent contexts, the effects $a+e$ are independent $\mathcal{N}(0,D_\mathrm{new})$ given $\mu$, so the decisions are independent Bernoulli variables with parameter $p(\mu)$; this gives (i) and (ii). For (iii), given $\mu$ and the effects $a$ and $a'$ of the two memberships, the four decisions are independent with parameters $p_N(\mu+a)$ and $p_N(\mu+a')$. The event has conditional probability ${p_N(\mu+a)^2(1-p_N(\mu+a'))^2}+{(1-p_N(\mu+a))^2p_N(\mu+a')^2}$, and $a$ and $a'$ are independent with the same law.
\end{proof}

\subsection{Single-item grading as the reference}

The bias is relative to single-item grading, not to a label. For $p_i=\frac12$ the main decision is 0 by our convention; the statements hold for either convention.

\begin{proof}[Proof of Proposition~\ref{prop:bias}]
(i) If $B_i=0$, then $r_i=d_i^\star$ and the probability is $v_i$. If $B_i=1$, the decision differs from $r_i$ if and only if it equals $d_i^\star$, which has probability $1-v_i$. (ii) follows from the law of large numbers. (iii) Given the item, each of the three decisions differs from $r_i$ with probability $v_i$ if $B_i=0$ and $1-v_i$ if $B_i=1$, independently. Hence $\Prob(X_i=3)\le v_i^3+B_i$. Since $v_i\le1-v_i$, $v_i^3\le v_i^2(1-v_i)=\min\{p_i^2(1-p_i),\,p_i(1-p_i)^2\}$. The mean over items of a minimum is at most the minimum of the means, and $\Prob(U=2)$ and $\Prob(U=1)$ are three times the means of $p_i^2(1-p_i)$ and of $p_i(1-p_i)^2$.
\end{proof}

Since $\min\{x,y\}\le(x+y)/2$ and ${\Prob(U=1)}+{\Prob(U=2)}=\frac32F_\mathrm{new}$, the right side of~\eqref{eq:bound} is at least $\Prob(X=3)-\frac14F_\mathrm{new}$. The contexts that three independent partitions give to an item are independent, so the bound applies to them.

\begin{assumption}
\label{ass:reference}
In addition to Assumption~\ref{ass:components}, the single-item decision is $r_i=\mathbf{1}[z^0_i>h_0]$ for a latent score $z^0_i$ of the item, standardised over items, such that $(z^0,\mu)$ is bivariate normal.
\end{assumption}

Since $z^0$ and $\mu$ belong to the item, the deviation $a+e$ of the score in a random context is independent of $(z^0,\mu)$ by Assumption~\ref{ass:components}.

Let $\rho_0$ be the correlation of $z^0$ with the score in a random context; under the assumption it is the tetrachoric correlation of the single-item and the batched decisions.

\begin{proposition}
\label{prop:biasgauss}
Let Assumption~\ref{ass:reference} hold. Then $\Cov(z^0,\mu)=\rho_0$ and $\rho_0^2\le1-D_\mathrm{new}$; let the inequality be strict. The mean bias is the probability that exactly one of two standard normal variables with correlation $\rho_0/\sqrt{1-D_\mathrm{new}}$ exceeds its threshold, which is $h_0$ for the first and $h/\sqrt{1-D_\mathrm{new}}$ for the second. With $p$ and $g_M$ as in Proposition~\ref{prop:patterns} and $q_0(\mu)=\Prob(z^0>h_0\mid\mu)$, the decisions in $M$ independent contexts all differ from $r$ with probability ${\E[q_0(\mu)(1-p(\mu))^M+(1-q_0(\mu))p(\mu)^M]}$, and their majority ($M$ odd) differs from $r$ with probability
\begin{equation*}
\E[q_0(\mu)(1-g_M(p(\mu)))+(1-q_0(\mu))g_M(p(\mu))].
\end{equation*}
\end{proposition}

\begin{proof}
$\Cov(z^0,z)=\Cov(z^0,\mu)$ by independence, and it equals $\rho_0$ because both have variance 1; the Cauchy--Schwarz inequality gives $\rho_0^2\le\Var\mu$. The probability $p_i=p(\mu_i)$ exceeds $\frac12$ if and only if $\mu_i>h$, so ${d^\star=\mathbf{1}[\mu>h]}$ and $B=\mathbf{1}[d^\star\ne\mathbf{1}[z^0>h_0]]$; the pair $(z^0,\mu/\sqrt{1-D_\mathrm{new}})$ is standard bivariate normal with correlation $\rho_0/\sqrt{1-D_\mathrm{new}}$. Given $(z^0,\mu)$, the decisions in independent contexts are independent Bernoulli variables with parameter $p(\mu)$, because $a+e$ is independent of $(z^0,\mu)$. If $r=1$ they all differ from $r$ with probability $(1-p)^M$ and if $r=0$ with probability $p^M$; the expectation given $\mu$ and then over $\mu$ gives the formula, and the one for the majority follows in the same way. Given $\mu$, $z^0$ is normal with mean $\rho_0\mu/(1-D_\mathrm{new})$ and variance $1-\rho_0^2/(1-D_\mathrm{new})$, which gives $q_0$.
\end{proof}

\subsection{Accuracy}

\begin{proposition}
\label{prop:acc}
Let $e_A$ and $e_B$ be the shares of items whose decisions in the conditions $A$ and $B$ differ from a binary label, and $e_{AB}$ the share for which both differ. Then $F(A,B)=e_A+e_B-2e_{AB}$, and $|e_A-e_B|\le F(A,B)\le{\min\{e_A+e_B,\,2-e_A-e_B\}}$.
\end{proposition}

\begin{proof}
If the two decisions on an item differ, exactly one of them differs from the label; if they agree, both or none do. So the items with a flip are those with exactly one error, whose share is $(e_A-e_{AB})+(e_B-e_{AB})$. The bounds follow from $\max\{0,e_A+e_B-1\}\le e_{AB}\le\min\{e_A,e_B\}$.
\end{proof}

\begin{proposition}
\label{prop:auc}
Let the score of an item be $z=\mu+\delta$, where, given the label $y\in\{0,1\}$, $\mu$ is $\mathcal{N}(\nu_y,\sigma_w^2)$ with $\nu_1\ge\nu_0$ and $\sigma_w>0$, and $\delta$ is $\mathcal{N}(0,\sigma_c^2)$, independent of $(\mu,y)$ and of the $\delta$ of other items. Let $\kappa=(\nu_1-\nu_0)/(\sqrt2\,\sigma_w)$ and let $D_w=\sigma_c^2/(\sigma_w^2+\sigma_c^2)$ be the share of the context in the variance within a class. The AUROC of $\mu$ is $\mathrm{AUC}_\mu=\Phi(\kappa)$, the AUROC of $z$ is $\mathrm{AUC}_z=\Phi(\kappa\sqrt{1-D_w})$, and
\begin{equation*}
0\le \mathrm{AUC}_\mu-\mathrm{AUC}_z\le\kappa\,D_w\,\varphi\big(\kappa\sqrt{1-D_w}\big),
\end{equation*}
with $\mathrm{AUC}_\mu-\mathrm{AUC}_z=\frac12\kappa\,\varphi(\kappa)D_w+O(D_w^2)$.
\end{proposition}

\begin{proof}
For two items with labels 1 and 0, the difference of the scores is normal with mean $\nu_1-\nu_0$ and variance $2(\sigma_w^2+\sigma_c^2)$, so $\mathrm{AUC}_z=\Phi(\kappa\sqrt{1-D_w})$, and with $\sigma_c=0$, $\mathrm{AUC}_\mu=\Phi(\kappa)$. On the interval from $\kappa\sqrt{1-D_w}$ to $\kappa$ the density $\varphi$ is at most its value at the left end, and the length of the interval is $\kappa(1-\sqrt{1-D_w})\le\kappa D_w$. The expansion follows from $1-\sqrt{1-D_w}=D_w/2+O(D_w^2)$.
\end{proof}

Here $\mathrm{AUC}_\mu$ is the AUROC of the scores averaged over contexts, not that of single-item grading, so the proposition explains only the part of a change of AUROC that the variance of the context causes. Within a class, the scores of an item in two independent contexts are bivariate normal with correlation $1-D_w$, so the loss of AUROC is proportional to $D_w$ and the flip rate within a class to $\sqrt{D_w}$ (Proposition~\ref{prop:law}).

\subsection{Two channels}

For a target in a fixed slot let $x$ be the texts of its neighbours in their order and $\gamma(x)$ the verdicts that the grader writes for the neighbours before the target. Take two independent sets of neighbours with texts $x^A$ and $x^B$, and let $\zeta(x,y)$ be the probability, renormalised over the levels, that the grader assigns to a level of at least $\tau$ for the target when, in its generation with the texts $x^A$, cut before the target's digit, the texts are set to $x$ and the earlier verdicts to $y$, all other tokens of that answer being kept. Let $\zeta_{AA}=\zeta(x^A,\gamma(x^A))$, $\zeta_{BA}=\zeta(x^B,\gamma(x^A))$, $\zeta_{AB}=\zeta(x^A,\gamma(x^B))$ and $\zeta_{BB}=\zeta(x^B,\gamma(x^B))$, so that $\Delta_T=\zeta_{BA}-\zeta_{AA}$, $\Delta_V=\zeta_{AB}-\zeta_{AA}$, $\Delta_{TV}=\zeta_{BB}-\zeta_{AA}$ and $J=\Delta_{TV}-\Delta_T-\Delta_V$.

\begin{proposition}
\label{prop:channels}
(i) If $\zeta(x,y)=\alpha(x)+\beta(y)$ for some functions $\alpha$ and $\beta$, then $J=0$ for every target. (ii) If, for every target, the four values are pairwise uncorrelated with a common mean and a common variance $\sigma^2>0$ over the draws of the neighbours, then $\E\Delta_T^2=\E\Delta_V^2=\E\Delta_{TV}^2=2\sigma^2$, $\E J^2=4\sigma^2$, and the correlation of $\Delta_T$ and $\Delta_V$ is $\frac12$. (iii) Let $\Delta'_T=\zeta_{BB}-\zeta_{AB}$ and $\Delta'_V=\zeta_{BB}-\zeta_{BA}$. If $\zeta(x,y)=g(\alpha(x)+\beta(y))$ for a non-decreasing function $g$, then for every target $\Delta_T$ and $\Delta'_T$ do not have opposite signs, and neither do $\Delta_V$ and $\Delta'_V$. Hence $\zeta_{AA}$ and $\zeta_{BB}$ are neither both larger nor both smaller than $\zeta_{AB}$ and $\zeta_{BA}$, $|J|\le\min\{\max(|\Delta_T|,|\Delta'_T|),\max(|\Delta_V|,|\Delta'_V|)\}$, and $\E J^2\le\E\Delta_V^2+\E{\Delta'_V}^2$.
\end{proposition}

\begin{proof}
(i) $J=\zeta_{BB}-\zeta_{BA}-\zeta_{AB}+\zeta_{AA}$, in which $\alpha$ and $\beta$ cancel. (ii) Each $\Delta$ is a difference of two uncorrelated variables with the same mean and variance, and $J$ is a signed sum of four; $\Cov(\Delta_T,\Delta_V)=\Var\zeta_{AA}=\sigma^2$. (iii) Let $u=\alpha(x^A)+\beta(\gamma(x^A))$, $a=\alpha(x^B)-\alpha(x^A)$ and $b=\beta(\gamma(x^B))-\beta(\gamma(x^A))$. Then $\Delta_T=g(u+a)-g(u)$ and $\Delta'_T=g(u+a+b)-g(u+b)$ are both non-negative if $a\ge0$ and both non-positive if $a\le0$, because $g$ does not decrease; in the same way $\Delta_V=g(u+b)-g(u)$ and $\Delta'_V=g(u+a+b)-g(u+a)$ have the sign of $b$ or are zero. If $\zeta_{AA}$ and $\zeta_{BB}$ were both larger than $\zeta_{AB}$ and $\zeta_{BA}$, then $\Delta_T=\zeta_{BA}-\zeta_{AA}<0<\zeta_{BB}-\zeta_{AB}=\Delta'_T$, and if both were smaller, the signs would be reversed. Finally $J=\Delta'_T-\Delta_T=\Delta'_V-\Delta_V$, the absolute difference of two numbers that do not have opposite signs is at most the larger of their absolute values, and $\max(x^2,y^2)\le x^2+y^2$.
\end{proof}

\section{Components, model checks and predictions}
\label{app:checks}

\begin{table*}[t]
\centering
\small
\setlength{\tabcolsep}{2pt}
\begin{tabular}{@{}ll llll rrrr@{}}
\toprule
& & \multicolumn{4}{c}{latent scale, shares of $D_\mathrm{new}$ [95\% CI]} & \multicolumn{4}{c}{decisions, shares of $F_\mathrm{new}$} \\
\cmidrule(lr){3-6}\cmidrule(l){7-10}
Model & Task & $T_\omega$ & $V_s$ & $V_N$ & $V_{sN}$ & $T_\omega$ & $V_s$ & $V_N$ & $V_{sN}$ \\
\midrule
Qwen3-4B & Harm & 0.69 [0.44, 0.99] & 0.09 [$-$0.06, 0.22] & 0.11 [$-$0.03, 0.24] & 0.11 [$-$0.22, 0.44] & 0.80 & 0.05 & 0.06 & 0.09 \\
 & Grading & 0.51 [0.31, 0.75] & 0.18 [0.06, 0.29] & 0.20 [0.08, 0.32] & 0.10 [$-$0.17, 0.35] & 0.69 & 0.09 & 0.11 & 0.11 \\
Qwen3-8B & Harm & 0.52 [0.34, 0.76] & 0.23 [0.11, 0.32] & 0.25 [0.14, 0.35] & 0.00 [$-$0.25, 0.25] & 0.69 & 0.12 & 0.14 & 0.04 \\
 & Grading & 0.42 [0.24, 0.67] & 0.31 [0.19, 0.41] & 0.09 [$-$0.04, 0.21] & 0.18 [$-$0.09, 0.42] & 0.62 & 0.17 & 0.05 & 0.16 \\
Gemma-4 & Harm & 0.49 [0.31, 0.74] & 0.32 [0.20, 0.41] & 0.16 [0.03, 0.27] & 0.03 [$-$0.24, 0.30] & 0.69 & 0.17 & 0.08 & 0.06 \\
 & Grading & 0.56 [0.35, 0.87] & 0.17 [0.03, 0.29] & 0.12 [$-$0.03, 0.25] & 0.15 [$-$0.15, 0.44] & 0.72 & 0.09 & 0.06 & 0.13 \\
Granite & Harm & 0.56 [0.35, 0.80] & 0.26 [0.15, 0.36] & 0.24 [0.13, 0.33] & $-$0.05 [$-$0.27, 0.19] & 0.73 & 0.14 & 0.14 & $-$0.01 \\
 & Grading & 0.61 [0.45, 0.81] & 0.12 [0.01, 0.20] & 0.11 [0.00, 0.20] & 0.16 [$-$0.06, 0.39] & 0.77 & 0.07 & 0.06 & 0.10 \\
GLM-4 & Harm & 0.65 [0.45, 0.91] & 0.25 [0.15, 0.33] & 0.01 [$-$0.12, 0.12] & 0.09 [$-$0.17, 0.34] & 0.80 & 0.14 & 0.00 & 0.07 \\
 & Grading & 0.58 [0.40, 0.81] & 0.24 [0.15, 0.33] & 0.25 [0.14, 0.34] & $-$0.07 [$-$0.32, 0.15] & 0.75 & 0.13 & 0.14 & $-$0.02 \\
\bottomrule
\end{tabular}
\caption{Components of the dissimilarity between two random partitions (Proposition~\ref{prop:pf}): order $T_\omega$, slot $V_s$, membership $V_N$ and the interaction of slot and membership $V_{sN}$. On the latent scale they are shares of $D_\mathrm{new}$; on the scale of decisions they are shares of the flip rate and need no threshold model. The order is estimated from the designated slots of ``keep slot, reorder the others'', weighted as in a random partition (\S\ref{sec:setup}). Intervals are bootstraps over items, drawn once for the three base partitions; they ignore that items share generated prompts (\S\ref{sec:where}). Harm and grading.}
\label{tab:components}
\end{table*}

\begin{table*}[t]
\centering
\small
\setlength{\tabcolsep}{4pt}
\begin{tabular}{@{}ll rr rr rrr@{}}
\toprule
& & \multicolumn{2}{c}{not all equal} & \multicolumn{2}{c}{two majorities} & \multicolumn{3}{c}{replicated flips} \\
\cmidrule(lr){3-4}\cmidrule(lr){5-6}\cmidrule(l){7-9}
Model & Task & obs. & pred. & obs. & pred. & obs. & pred. & pred., $V_N=0$ \\
\midrule
Qwen3-4B & Harm & 30.8 & 30.4 & 8.9 & 8.9 & 3.0 & 2.8 & 2.2 \\
 & Grading & 31.1 & 30.7 & 8.3 & 9.2 & 3.4 & 3.4 & 2.1 \\
 & Stars & 8.6 & 8.6 & 2.4 & 2.6 & 0.9 & 1.1 & 0.6 \\
Qwen3-8B & Harm & 30.8 & 30.4 & 8.5 & 10.2 & 3.5 & 3.9 & 2.3 \\
 & Grading & 28.0 & 28.3 & 8.6 & 8.3 & 2.5 & 2.5 & 2.1 \\
 & Stars & 8.6 & 8.6 & 2.9 & 2.7 & 0.8 & 0.7 & 0.7 \\
Gemma-4 & Harm & 29.2 & 28.9 & 8.9 & 9.0 & 2.9 & 3.1 & 2.1 \\
 & Grading & 24.5 & 24.4 & 7.6 & 7.3 & 2.5 & 2.3 & 1.8 \\
 & Stars & 7.6 & 7.4 & 1.8 & 2.1 & 1.0 & 0.7 & 0.5 \\
Granite & Harm & 34.7 & 34.2 & 10.6 & 10.7 & 3.8 & 4.1 & 2.4 \\
 & Grading & 49.9 & 48.8 & 13.7 & 15.5 & 5.2 & 4.8 & 3.8 \\
 & Stars & 9.6 & 9.4 & 2.4 & 2.7 & 0.9 & 0.8 & 0.7 \\
GLM-4 & Harm & 40.1 & 39.9 & 12.2 & 12.1 & 3.6 & 3.2 & 3.2 \\
 & Grading & 33.1 & 32.5 & 9.3 & 11.1 & 3.4 & 4.3 & 2.5 \\
 & Stars & 7.2 & 7.1 & 1.7 & 2.2 & 0.7 & 0.9 & 0.5 \\
\bottomrule
\end{tabular}
\caption{Patterns of decisions (\% of items), observed and predicted by the threshold model from the dissimilarities and the share of positive decisions (Proposition~\ref{prop:patterns}). \emph{Not all equal}: the decisions in six batches, three reordered and three with new neighbours, which we treat as independent; it is predicted from the mean dissimilarity of the 15 pairs of these batches and their share of positive decisions, the other patterns from $D_\mathrm{new}$, $D_\mathrm{resh}$ and the share of positive decisions in the nine batched conditions. \emph{Two majorities}: the majority of the three reordered batches differs from that of the three batches with new neighbours. \emph{Replicated flips}: two orders of one batch agree, two orders of another batch agree, and the batches differ; the last column sets the main effect of membership to zero by setting $D_\mathrm{new}$ to $D_\mathrm{resh}$; setting $D_\mathrm{resh}$ to $D_\mathrm{new}$ instead changes this prediction by at most \CheckRepAltMax{} points, and it stays below the observed share in \CheckRepAltLow{} of the ten pairs of harm and grading. The standard errors of observed minus predicted, from the item bootstrap with the prediction re-estimated in every resample, are \CheckSplitSE, \CheckMajSE{} and \CheckRepSE{} points for the three patterns on harm and grading. Counts in the text are made before rounding.}
\label{tab:modelcheck}
\end{table*}

\begin{table*}[t]
\centering
\small
\setlength{\tabcolsep}{4pt}
\begin{tabular}{@{}ll r llll@{}}
\toprule
Model & Task & $m=1$ & $m=3$ & $m=5$ & $m=9$ & $m\to\infty$ \\
\midrule
Qwen3-4B & Harm & 13.1 & 9.5 [8.3, 10.9] & 8.1 [6.7, 9.7] & 6.8 [5.1, 8.7] & 4.5 [0.0, 7.0] \\
 & Grading & 13.5 & 10.3 [9.0, 11.6] & 9.1 [7.6, 10.5] & 8.0 [6.4, 9.6] & 6.3 [3.8, 8.2] \\
Qwen3-8B & Harm & 14.8 & 11.5 [10.2, 12.8] & 10.3 [8.8, 11.7] & 9.3 [7.6, 10.8] & 7.6 [5.4, 9.4] \\
 & Grading & 12.2 & 8.8 [7.7, 10.0] & 7.4 [6.2, 8.8] & 6.2 [4.7, 7.7] & 3.8 [0.0, 6.1] \\
Gemma-4 & Harm & 13.2 & 9.8 [8.5, 11.1] & 8.5 [7.0, 9.9] & 7.4 [5.5, 8.9] & 5.5 [2.3, 7.4] \\
 & Grading & 10.8 & 7.9 [6.7, 9.1] & 6.7 [5.4, 8.1] & 5.7 [4.1, 7.2] & 3.8 [0.0, 5.8] \\
Granite & Harm & 15.5 & 12.0 [10.7, 13.5] & 10.8 [9.2, 12.3] & 9.7 [7.9, 11.3] & 7.8 [5.5, 9.7] \\
 & Grading & 21.9 & 16.4 [14.7, 17.9] & 14.1 [12.0, 16.0] & 12.0 [9.4, 14.2] & 8.0 [1.4, 11.2] \\
GLM-4 & Harm & 17.5 & 12.2 [11.2, 13.6] & 9.9 [9.0, 11.6] & 7.7 [6.9, 9.8] & 1.7 [0.0, 6.6] \\
 & Grading & 16.1 & 12.6 [11.2, 14.0] & 11.3 [9.7, 12.7] & 10.2 [8.5, 11.7] & 8.4 [6.1, 10.2] \\
\bottomrule
\end{tabular}
\caption{Decisions on which two batches with independent memberships (from two independent random partitions) would differ after majority voting over $m$ permutations each (\%), predicted by Proposition~\ref{prop:vote} from $D_\mathrm{new}$, $D_\mathrm{resh}$ and the share of positive decisions, with 95\% bootstrap intervals over items, drawn as for Table~\ref{tab:components}; they ignore that items share generated prompts. These predictions were made from the main experiment before the voting runs; Table~\ref{tab:votemeasured} gives the measured flips, which lie inside these intervals at $m=9$ in \VoteTwelveInside{} of the ten pairs (for Qwen3-8B on harm the measured value lies just below the lower end, and both round to the same value). Harm and grading.}
\label{tab:voting}
\end{table*}

\begin{table*}[t]
\centering
\small
\setlength{\tabcolsep}{4pt}
\begin{tabular}{@{}ll rr rr rr rr l@{}}
\toprule
& & \multicolumn{2}{c}{one batch} & \multicolumn{2}{c}{majority of 3} & \multicolumn{2}{c}{all three} & \multicolumn{2}{c}{mean bias} & \\
\cmidrule(lr){3-4}\cmidrule(lr){5-6}\cmidrule(lr){7-8}\cmidrule(lr){9-10}
Model & Task & obs. & pred. & obs. & pred. & obs. & pred. & maj.\ of 9 & model & bound [95\% CI] \\
\midrule
Qwen3-4B & Harm & 13.7 & 13.7 & 11.6 & 12.0 & 5.0 & 4.8 & 10.7 & 10.2 & 2.15 [1.12, 3.18] \\
 & Grading & 15.6 & 15.7 & 14.2 & 14.1 & 6.2 & 6.3 & 13.2 & 12.7 & 3.34 [2.20, 4.51] \\
 & Stars & 5.7 & 5.8 & 5.3 & 5.4 & 3.0 & 3.0 & 5.3 & 5.2 & 2.04 [1.34, 2.93] \\
Qwen3-8B & Harm & 14.0 & 14.0 & 11.4 & 11.8 & 4.2 & 4.0 & 10.0 & 9.4 & 1.03 [0.03, 2.05] \\
 & Grading & 15.0 & 15.0 & 13.3 & 13.7 & 6.7 & 6.6 & 13.4 & 12.4 & 3.92 [2.77, 5.12] \\
 & Stars & 4.7 & 4.7 & 4.2 & 4.2 & 2.0 & 1.9 & 4.0 & 3.8 & 0.98 [0.42, 1.69] \\
Gemma-4 & Harm & 19.7 & 19.8 & 19.3 & 19.3 & 10.2 & 10.2 & 18.7 & 19.3 & 7.06 [5.70, 8.57] \\
 & Grading & 10.7 & 10.8 & 9.0 & 9.2 & 3.5 & 3.4 & 8.9 & 7.6 & 1.08 [0.18, 2.03] \\
 & Stars & 4.8 & 4.8 & 4.5 & 4.5 & 2.6 & 2.6 & 4.6 & 4.3 & 1.78 [1.13, 2.58] \\
Granite & Harm & 19.5 & 19.5 & 18.3 & 18.2 & 8.5 & 8.6 & 18.1 & 17.1 & 4.73 [3.42, 6.09] \\
 & Grading & 19.7 & 19.7 & 15.8 & 16.4 & 5.3 & 5.0 & 12.9 & 12.3 & 0.36 [$-$0.77, 1.54] \\
 & Stars & 5.6 & 5.7 & 5.2 & 5.3 & 2.8 & 2.8 & 4.7 & 4.9 & 1.89 [1.15, 2.69] \\
GLM-4 & Harm & 16.1 & 16.2 & 13.1 & 13.5 & 4.6 & 4.4 & 11.8 & 10.5 & 0.35 [$-$0.62, 1.47] \\
 & Grading & 16.7 & 16.7 & 14.3 & 14.6 & 5.9 & 5.7 & 12.2 & 12.4 & 2.20 [1.05, 3.37] \\
 & Stars & 4.9 & 5.0 & 4.6 & 4.7 & 2.7 & 2.7 & 4.2 & 4.6 & 1.95 [1.25, 2.74] \\
\bottomrule
\end{tabular}
\caption{Disagreement with single-item grading (\% of items). \emph{One batch}: the decision in a random partition differs from the single-item decision. \emph{Majority of 3}, \emph{all three}: the majority of the three partitions differs from it, or all three do; observed and under the threshold model (Proposition~\ref{prop:biasgauss}), which is fitted to the single-item and batched decisions in pairs. \emph{Mean bias}: the share of items whose main batched decision differs from the single-item decision, as the majority of the nine batched decisions that we have per item, which are not independent, and under the model. \emph{Bound}: the lower bound of Equation~\eqref{eq:bound}, rounded down to two decimals, with a percentile interval from 10{,}000 resamples of the items. Counts in the text are made before rounding. Table~\ref{tab:bvote} measures the majority over fifteen independent batches.}
\label{tab:bias}
\end{table*}

\begin{table*}[t]
\centering
\small
\setlength{\tabcolsep}{5pt}
\begin{tabular}{@{}ll rrrr rr@{}}
\toprule
Model & Task & $\Delta_T^2$ & $\Delta_V^2$ & $\Delta_{TV}^2$ & $J^2$ & ratio & corr. \\
\midrule
Qwen3-4B & Harm & 7.7 & 6.6 & 7.3 & 15.1 & 2.1 & 0.57 \\
 & Grading & 10.3 & 7.7 & 11.3 & 17.2 & 1.9 & 0.54 \\
Qwen3-8B & Harm & 10.0 & 9.2 & 7.8 & 21.1 & 2.2 & 0.47 \\
 & Grading & 8.5 & 8.3 & 6.2 & 20.9 & 2.5 & 0.62 \\
Gemma-4 & Harm & 9.1 & 6.8 & 6.0 & 17.2 & 2.2 & 0.48 \\
 & Grading & 4.5 & 2.2 & 4.8 & 2.3 & 0.7 & 0.26 \\
Granite & Harm & 9.9 & 8.5 & 10.0 & 19.5 & 2.1 & 0.57 \\
 & Grading & 14.6 & 9.9 & 14.8 & 22.6 & 1.8 & 0.54 \\
GLM-4 & Harm & 6.8 & 6.8 & 6.5 & 16.4 & 2.4 & 0.61 \\
 & Grading & 10.6 & 8.9 & 7.3 & 24.6 & 2.5 & 0.63 \\
\bottomrule
\end{tabular}
\caption{Mean squares ($\times100$) of the change of the target's probability $\zeta$ when the neighbours' texts, the earlier verdicts or both are taken from another batch, and of the interaction contrast $J$. \emph{Ratio}: the mean square of $J$ over the mean of those of $\Delta_T$ and $\Delta_V$, which is 0 if the channels add up and 2 for unrelated contexts. \emph{Corr.}: correlation of $\Delta_T$ and $\Delta_V$, which is 0.5 for unrelated contexts. Harm and grading.}
\label{tab:channels}
\end{table*}

\begin{table*}[t]
\centering
\small
\setlength{\tabcolsep}{5pt}
\begin{tabular}{@{}ll llll@{}}
\toprule
Model & Task & level $\ge1$ & level $\ge2$ & level $\ge3$ & level $\ge4$ \\
\midrule
Qwen3-4B & Harm & 0.058 (8.9) & 0.089 (13.1) & 0.145 (14.7) & 0.232$^\dagger$ (2.7) \\
 & Grading & 0.113 (12.9) & 0.103 (13.4) & -- & -- \\
 & Stars & 0.024 (4.8) & 0.013 (4.4) & 0.008 (3.9) & 0.019 (5.7) \\
Qwen3-8B & Harm & 0.111 (13.7) & 0.116 (14.9) & 0.258 (16.2) & 0.203$^\dagger$ (1.5) \\
 & Grading & 0.077 (12.1) & 0.084 (12.1) & -- & -- \\
 & Stars & 0.043 (6.1) & 0.012 (4.7) & 0.008 (4.0) & 0.028 (6.8) \\
Gemma-4 & Harm & 0.081 (12.8) & 0.106 (13.0) & 0.128 (8.3) & 0.249$^\dagger$ (0.2) \\
 & Grading & 0.065 (10.3) & 0.057 (10.8) & -- & -- \\
 & Stars & 0.025 (3.9) & 0.008 (3.6) & 0.005 (3.1) & 0.023 (5.5) \\
Granite & Harm & 0.165 (18.4) & 0.120 (15.5) & 0.135 (14.9) & 0.163 (8.1) \\
 & Grading & 0.228 (16.1) & 0.238 (21.8) & -- & -- \\
 & Stars & 0.029 (5.5) & 0.012 (4.6) & 0.008 (4.0) & 0.027 (6.8) \\
GLM-4 & Harm & 0.071 (11.0) & 0.147 (17.4) & 0.197 (16.6) & 0.252$^\dagger$ (0.3) \\
 & Grading & 0.292 (14.2) & 0.132 (16.1) & -- & -- \\
 & Stars & 0.023 (4.9) & 0.008 (3.8) & 0.005 (3.2) & 0.026 (6.3) \\
\bottomrule
\end{tabular}
\caption{Latent dissimilarity between two random partitions, with the flip rate in \% in brackets, when the decision is that the index of the level (counted from 0) is at least 1, 2, 3 or 4. The primary decisions are $\ge2$ for harm and grading and $\ge3$ for stars. $^\dagger$Fewer than 5\% of the decisions on one side; left out wherever thresholds are compared.}
\label{tab:thresholds}
\end{table*}

Tables~\ref{tab:components} to~\ref{tab:thresholds} give the results of the main experiment behind Sections~\ref{sec:where} to~\ref{sec:how} for its five models on harm and grading (Appendix~\ref{app:followup} gives the follow-up experiments); Tables~\ref{tab:modelcheck}, \ref{tab:bias} and~\ref{tab:thresholds} also give stars, whose $D_\mathrm{new}$ is too small to decompose (\S\ref{sec:where}). If all levels threshold one latent score, as in an ordinal version of the threshold model, the latent dissimilarity between two partitions would be the same at every threshold of the scale. Between the thresholds that have at least 5\% of the decisions on either side, the largest dissimilarity is \ThreshGradingRatio{} times the smallest on grading, where \ThreshGradingFlat{} of the five models are below 1.2, \ThreshHarmRatio{} times on harm, where it grows with the threshold for \ThreshHarmGrows{} of the five models, and \ThreshStarsRatio{} times on stars, where all four thresholds qualify and the primary threshold gives the smallest dissimilarity for \ThreshStarsPrimaryMin{} of the five models. A context that contributes more variance to some items than to others, or items whose scores are not normal, such as the five rating classes of stars, make the dissimilarity depend on the threshold; the dissimilarities that we report describe the primary decision.

\section{Follow-up experiments}
\label{app:followup}

After the main experiment, and without gates fixed in advance, we ran three follow-ups on harm and grading, on H200 NVL GPUs with torch 2.11 and transformers 5.17, in bf16 with greedy decoding and the prompt templates of the main experiment (model revisions in the footnote of \S\ref{sec:design}). Before its run, every model passed a smoke test of 44 prompts whose answers all had to parse, and recomputations from the raw records, mostly without our analysis code, reproduced every point estimate below.

\paragraph{Larger models.} \mbox{Qwen3-14B} and \mbox{Qwen3-32B} with thinking disabled, \mbox{Gemma-4-31B-it} and \mbox{GLM-4-32B-0414} grade the single items with their two baselines, the three partitions with their three derived conditions (Table~\ref{tab:conditions}), the baselines of partition 0, and partition 1 with the reworded instruction (Table~\ref{tab:scale}). A new partition changes \ScaleNewRange\% of their decisions, \ScaleExcessRange{} points above the largest baseline, which is the reworded instruction in all eight pairs; every lower bound is about 3 points or more, so every pair clears the bar of G1 (stars, on which G1 failed for the main models, was not run). By the measure fixed before the runs, the bar of G2 is cleared in \ScaleGTwoCount{} of the eight pairs, on both tasks only for \ScaleGTwoBoth; with the designated slots weighted as in a random partition (\S\ref{sec:where}), only in \ScaleGTwoWeighted, for \ScaleGTwoWeightedPairs. The excess over the largest baseline is smaller for \mbox{Gemma-4-31B} than for \mbox{Gemma-4} and for \mbox{GLM-4-32B} than for \mbox{GLM-4} on both tasks, and in Qwen3 it falls with size on grading but not on harm; with one larger model in each of the other two families, another chat template and output format for \mbox{Gemma-4-31B} (an empty thought channel, fenced JSON), and another type of GPU, this does not establish a trend with size. A batch disagrees with single-item grading on \ScaleAloneRange\% of decisions, and the bound of Equation~\eqref{eq:bound} excludes zero in \ScaleBoundSig{} of the eight pairs. Once the batched threshold is chosen in every partition so that batches give as many positive decisions as single-item grading, the bound falls by more than half in \ScaleBoundCalHalf{} of these \ScaleBoundSig{} pairs and excludes zero clearly only for \ScaleBoundCalPairs{} (and only just for \ScaleBoundCalBorder). The bound reacts strongly to a shift of the operating point: for the main models, the same recalibration of their runs in the main experiment removes more than half of it in \MainBoundCalHalf{} of the \BiasBoundSig{} pairs in which it excludes zero, although, pooled, the majority of fifteen batches keeps \BvoteCalShare\% of its disagreement after recalibration (\S\ref{sec:voting}). How much of the bias of the larger models is such a shift therefore remains open. Under the threshold model, which we did not check with more partitions for them, the mean bias is \ScaleBiasModelRange\%. With the reworded instruction a new partition changes \ScaleRewordedRange\% of decisions, \ScaleWordingRange{} points against the original wording. Order and slot carry most of $D_\mathrm{new}$ for \mbox{Qwen3-14B}, \mbox{Qwen3-32B} and \mbox{GLM-4-32B}, as for the main models, although \ScaleVsNNeg{} of their six estimates of the interaction of slot and membership are negative (\ScaleVsNRange, all with intervals that include zero). For \mbox{Gemma-4-31B}, $D_\mathrm{new}$ is \GemmaLargeDNew, and the estimated share of the interaction on grading, \GemmaLargeVsN, has an interval below zero, which Proposition~\ref{prop:pf} excludes; we do not decompose it (\S\ref{sec:where}). On harm, the four models put \ScaleLevelThreeAlone\% of the comments on level~3 when they grade them alone and \ScaleLevelThreeBatched\% in batches, so that their single-item levels tie more often; the AUROC of the expected level, which has no ties, changes by \ScaleExpAurocRange{} on harm and grading (\S\ref{sec:accuracy}).

\begin{table*}[t]
\centering
\small
\setlength{\tabcolsep}{3.5pt}
\begin{tabular}{@{}ll rrl r rrr rr@{}}
\toprule
& & & & & & \multicolumn{3}{c}{AUROC} & \multicolumn{2}{c}{flagged} \\
\cmidrule(lr){7-9}\cmidrule(l){10-11}
Model & Task & new & base & excess [95\% CI] & vs.\ alone & alone & batched & $\Delta$ expected & alone & batched \\
\midrule
Qwen3-14B & Harm & 11.0 & 3.8 & +7.3 [+5.9, +8.6] & 12.0 & 0.793 & 0.804 & $-$0.011 & 62 & 61 \\
 & Grading & 9.1 & 2.8 & +6.3 [+5.0, +7.6] & 11.9 & 0.832 & 0.811 & $-$0.023 & 36 & 44 \\
Qwen3-32B & Harm & 14.6 & 5.5 & +9.1 [+7.0, +10.9] & 17.0 & 0.760 & 0.815 & +0.012 & 64 & 50 \\
 & Grading & 10.3 & 4.7 & +5.7 [+4.3, +6.8] & 9.1 & 0.842 & 0.831 & $-$0.022 & 34 & 35 \\
Gemma-4-31B & Harm & 8.4 & 2.6 & +5.8 [+4.6, +6.9] & 10.8 & 0.837 & 0.869 & +0.012 & 54 & 45 \\
 & Grading & 5.9 & 1.8 & +4.2 [+3.1, +5.3] & 5.8 & 0.853 & 0.832 & $-$0.005 & 48 & 48 \\
GLM-4-32B & Harm & 11.1 & 3.2 & +7.9 [+6.4, +9.5] & 17.9 & 0.800 & 0.857 & $-$0.001 & 66 & 49 \\
 & Grading & 10.6 & 4.3 & +6.3 [+4.8, +7.8] & 14.2 & 0.799 & 0.834 & +0.003 & 54 & 42 \\
\bottomrule
\end{tabular}
\caption{Four larger models on harm and grading, on H200 NVL GPUs. \emph{New}, \emph{base} and \emph{excess} as in Table~\ref{tab:main}; \emph{vs.\ alone}: decisions that differ from single-item grading (\%, mean of the three partitions). AUROC of the written level alone and in batches (mean of the three partitions), and the change of the AUROC of the expected level (the mean of the level distribution), batched minus alone; share of items above the decision threshold (\%).}
\label{tab:scale}
\end{table*}

\paragraph{Voting over permutations.} Every batch of partitions 0 and 1 was graded in its own order, p0, and in eight further uniformly random orders of the same members, together with the single items and, with the reworded instruction, partitions 0 and 1 (7{,}200 prompts per model). Table~\ref{tab:votemeasured} gives the flips between the majority votes of the two partitions over $m$ orders, averaged over all sets of $m$ of the nine orders of each partition, and the share of the flips of single orders that is left after nine permutations. The statistics use the items with a verdict in all eighteen orders (at least \VoteNMin{} of 2{,}000); their intervals are normal, with a variance clustered by the batches of both partitions \citep{cameron2011robust}. Fitted to the same runs, the threshold model (Proposition~\ref{prop:vote}) over-predicts the flips after $m$ permutations by \VoteInSampleGap{} points when pooled, significantly after Holm's adjustment in \VoteHolm{} of the ten pairs at $m=3$, 5, 7 and~9. On the scale of decisions, the main effect of membership, half the difference between the flips across the two partitions and those between orders of the same batch, is \VoteVNDecRange{} points, and on the latent scale it is \VoteVNLatentRange{} of $D_\mathrm{new}$; on both scales its interval excludes zero in all ten pairs. Against single items, the majority of the nine orders of one batch still differs for \VoteAloneNineRange\% of the items (one order: \VoteAloneOneRange\%), because voting over permutations converges to the decision given the membership. With the reworded instruction a new partition changes \VoteRewordedRange\% of decisions, \VoteWordingRange{} points against order p0 in the original wording; over this experiment and the previous one, it changes fewer decisions than under the original wording in \WordingNegative{} of the \WordingPairs{} model--task pairs.

\begin{table*}[t]
\centering
\small
\setlength{\tabcolsep}{4.5pt}
\begin{tabular}{@{}ll r rrrr lrr@{}}
\toprule
& & & \multicolumn{4}{c}{flips after $m$ permutations} & \multicolumn{3}{c}{left after nine (\% of $m=1$)} \\
\cmidrule(lr){4-7}\cmidrule(l){8-10}
Model & Task & $n$ & $m=1$ & $m=3$ & $m=5$ & $m=9$ & measured [95\% CI] & in sample & main exp. \\
\midrule
Qwen3-4B & Harm & 2{,}000 & 13.3 & 9.4 & 8.0 & 6.9 & 52 [44, 60] & 52 & 52 \\
 & Grading & 1{,}977 & 12.9 & 9.7 & 8.5 & 7.4 & 57 [50, 64] & 60 & 59 \\
Qwen3-8B & Harm & 2{,}000 & 14.1 & 10.3 & 8.8 & 7.6 & 54 [48, 59] & 57 & 63 \\
 & Grading & 1{,}984 & 12.6 & 9.2 & 8.0 & 7.0 & 55 [48, 63] & 58 & 51 \\
Gemma-4 & Harm & 2{,}000 & 12.4 & 9.4 & 8.2 & 7.1 & 57 [50, 65] & 58 & 56 \\
 & Grading & 2{,}000 & 11.0 & 8.4 & 7.6 & 6.6 & 60 [51, 69] & 60 & 52 \\
Granite & Harm & 1{,}998 & 14.4 & 10.6 & 9.3 & 8.4 & 58 [51, 65] & 56 & 62 \\
 & Grading & 2{,}000 & 21.9 & 16.1 & 13.7 & 12.4 & 56 [51, 62] & 57 & 55 \\
GLM-4 & Harm & 2{,}000 & 18.0 & 13.2 & 11.3 & 9.7 & 54 [47, 60] & 54 & 44 \\
 & Grading & 2{,}000 & 15.4 & 11.1 & 9.5 & 8.0 & 52 [46, 58] & 58 & 63 \\
\bottomrule
\end{tabular}
\caption{Voting over permutations, measured: decisions on which the majority votes of partitions 0 and 1 over $m$ of the nine orders of each batch differ (\%), averaged over all sets of $m$ orders of each partition, so that $m=1$ averages over the 81 pairs of orders and differs from \emph{new} in Table~\ref{tab:main}; and the share of the flips at $m=1$ that is left at $m=9$: measured (95\% normal intervals, with a variance clustered by the batches of both partitions), predicted by Proposition~\ref{prop:vote} from $D_\mathrm{new}$, $D_\mathrm{resh}$ and the share of positive decisions of the same runs (in sample), and predicted from the main experiment before these runs (main exp., Table~\ref{tab:voting}). $n$: items with a verdict in all eighteen orders.}
\label{tab:votemeasured}
\end{table*}

\paragraph{Hardware.} Order p0 repeats the prompts of partitions 0 and 1 of the main experiment. On the H200, \HardwareDifferRange\% of these decisions per partition differ from the RTX 5090 run. On partition 0, where the main experiment measured numerical noise by generating the same prompts one at a time on the RTX 5090 (\S\ref{sec:design}), this share minus that noise is \HardwareVsNumericRange{} points; partition 1 has no such baseline and reaches \HardwarePartOneMax\%. The flips between the partitions change by \HardwareFlipRange{} points, and the interval of this change excludes zero in \HardwareFlipSig{} of the ten pairs. Partition 2, which the voting over batches below grades again, differs from the RTX 5090 run for \BvoteHardwareRange\% of the decisions.

\paragraph{Voting over batches.} Partition 2 of the main experiment and twelve new uniformly random partitions were graded once each (2{,}080 prompts per model); with partitions 0 and 1 (order p0) and the single items of the voting runs, every item has fifteen independent batches and a single-item decision from the same type of GPU, driver and software. Table~\ref{tab:bvote} gives the share of items whose majority over $M$ random batches, averaged over all sets of $M$ of the fifteen partitions, differs from the single-item decision; it changes by at most \BvoteNineToFifteen{} points between nine and fifteen batches. The statistics use the items with a verdict alone and in all fifteen partitions (\BvoteNRange). Their intervals are normal, with a variance computed from the influence of every item (delete-one jackknife), clustered by the batches of all fifteen partitions and not below the variance for independent items; on synthetic decisions from the threshold model with effects shared within batches, such intervals covered the true values in 95--99\% of the simulations. The calibrated statistics, whose cuts are quantiles, take their variance from resamples of the items and of the batches of every partition instead; on the synthetic decisions, these standard deviations were about 40\% too large. Fitted in sample, the model predicts the majority over $M$ batches within \BvoteInSampleMax{} points for every $M$. Measured minus predicted, pooled over the ten pairs, is \BvoteDFivePooled{} points for this majority at $M=5$, \BvoteGThree{} points for the flips between two majorities over three batches and \BvoteUFifteen{} points for the items whose fifteen decisions are not all equal; after Holm's adjustment these differences are significant in \BvoteHolmD, \BvoteHolmG{} and \BvoteHolmU{} of the ten pairs. The calibrated column chooses the cut on $\zeta$ in every partition so that the batches give the single-item positive rate of the same items: it measures how much of the disagreement is a shift of the operating point, not what a recalibration estimated on other data would achieve.

\begin{table*}[t]
\centering
\small
\setlength{\tabcolsep}{4.5pt}
\begin{tabular}{@{}ll r r l rrr r@{}}
\toprule
& & & & & \multicolumn{3}{c}{threshold model} & \\
\cmidrule(lr){6-8}
Model & Task & $n$ & $M=1$ & $M=15$ [95\% CI] & in sample & main exp. & $M\to\infty$ & calibrated \\
\midrule
Qwen3-4B & Harm & 1{,}999 & 13.6 & 10.2 [8.7, 11.6] & 10.1 & 10.7 & 10.2 & 10.1 \\
 & Grading & 1{,}975 & 15.7 & 13.3 [11.8, 14.8] & 13.2 & 13.0 & 12.7 & 9.6 \\
Qwen3-8B & Harm & 1{,}999 & 13.6 & 9.6 [8.3, 10.8] & 10.1 & 10.0 & 9.4 & 9.9 \\
 & Grading & 1{,}988 & 15.7 & 13.7 [12.2, 15.2] & 13.3 & 12.7 & 12.4 & 13.5 \\
Gemma-4 & Harm & 2{,}000 & 19.0 & 17.8 [15.4, 20.2] & 18.6 & 19.3 & 19.3 & 8.9 \\
 & Grading & 2{,}000 & 10.9 & 8.4 [7.1, 9.6] & 8.2 & 8.0 & 7.6 & 7.9 \\
Granite & Harm & 1{,}999 & 19.2 & 17.6 [15.9, 19.2] & 17.3 & 17.3 & 17.1 & 9.8 \\
 & Grading & 2{,}000 & 19.8 & 13.1 [11.5, 14.6] & 13.4 & 13.4 & 12.3 & 12.9 \\
GLM-4 & Harm & 2{,}000 & 16.3 & 10.7 [9.3, 12.1] & 11.2 & 11.3 & 10.5 & 10.7 \\
 & Grading & 2{,}000 & 15.9 & 12.2 [10.7, 13.6] & 12.4 & 13.0 & 12.4 & 10.7 \\
\bottomrule
\end{tabular}
\caption{Voting over batches: items whose majority over $M$ independent random batches differs from the single-item decision of the same runs (\%), measured, averaged over all sets of $M$ of the fifteen partitions, with 95\% intervals clustered by the batches of all fifteen partitions; under the threshold model at $M=15$, fitted to these runs (in sample) and to the main experiment (main exp.), and the limit of the latter as $M\to\infty$, which is the mean bias under the model in Table~\ref{tab:bias}; and measured at $M=15$ with the batched threshold chosen in every partition to match the single-item positive rate (calibrated). $n$: items with a verdict alone and in all fifteen partitions.}
\label{tab:bvote}
\end{table*}

\end{document}